\documentclass{article}
\usepackage{cpg_arxiv}
\usepackage{microtype}
\usepackage{graphicx}
\usepackage{subcaption}
\usepackage{booktabs}
\usepackage{hyperref}
\usepackage{multirow}

\usepackage{amsmath}
\usepackage{amssymb}
\usepackage{mathtools}
\usepackage{amsthm}
\usepackage[capitalize,noabbrev]{cleveref}
\usepackage{algorithm}
\usepackage{algorithmic}
\usepackage{array} 
\newcolumntype{C}[1]{>{\centering\arraybackslash}p{#1}}
\usepackage{pifont}
\usepackage{fancyvrb}

\usepackage{tcolorbox}
\tcbuselibrary{breakable}
\usepackage{xurl}
\usepackage{natbib}
\usepackage{wrapfig}

\begin{document}

\title{RAT-Bench: A Comprehensive Benchmark for Text Anonymization}
% \vspace{100.2cm} % tune if needed
% \begin{flushright}
%   \includegraphics[width=2cm]{rat.png}
% \end{flushright}
% \vspace{-2.6cm} % tune if needed

\authors{
  Nata\v sa Kr\v co\equalcontrib,
  Zexi Yao\equalcontrib,
  Matthieu Meeus\equalcontrib,
  Yves-Alexandre de Montjoye
}

\affiliations{
  \textit{Imperial College London} \\
  \equalcontrib Equal contribution
}

\maketitle
\makeatletter\def\Hy@Warning#1{}\makeatother
%%%%IF EQUAL CONTRIBUTION 
\def\thefootnote{*}\footnotetext{Equal contribution}\def\thefootnote{\arabic{footnote}}

\begin{abstract}
  Data containing personal information is increasingly used to train, fine-tune, or query Large Language Models (LLMs). Text is typically scrubbed of identifying information prior to use, often with tools such as Microsoft's Presidio or Anthropic's PII purifier. These tools have traditionally been evaluated on their ability to remove specific identifiers (e.g., names), yet their effectiveness at preventing re-identification remains unclear. We introduce RAT-Bench, a comprehensive benchmark for text anonymization tools based on re-identification risk. Using U.S. demographic statistics, we generate synthetic text containing various direct and indirect identifiers across domains, languages, and difficulty levels. We evaluate a range of NER- and LLM-based text anonymization tools and, based on the attributes an LLM-based attacker is able to correctly infer from the anonymized text, we report the risk of re-identification in the U.S. population, while properly accounting for the disparate impact of identifiers. We find that, while capabilities vary widely, even the best tools are far from perfect in particular when direct identifiers are not written in standard ways and when indirect identifiers enable re-identification. Overall we find LLM-based anonymizers, including new iterative anonymizers, to provide a better privacy-utility trade-off albeit at a higher computational cost. Importantly, we also find them to work well across languages. We conclude with recommendations for future anonymization tools and will release the benchmark and encourage community efforts to expand it, in particular to other geographies.
\end{abstract}

\section{Introduction}
Domain-specific text is essential for advancing today’s AI models and systems. Modern Large Language Models (LLMs)~\citep{grattafiori2024llama,openai2025gpt5} are not only trained on vast amounts of publicly available data~\citep{brown2020language,touvron2023llama} but are also frequently fine-tuned on specialized datasets to perform specific tasks. For example, AI companies, including OpenAI~\citep{openai2025userdata} and Anthropic~\citep{Loizos2025AnthropicOptOut} are leveraging user–chatbot interactions to improve their models~\citep{king2025user}, a practice recognized as key to generating a sustainable competitive advantage~\citep{huang2023generative} while specialized models have been developed for high-stakes domains such as medicine~\citep{kraljevic2022foresight} and law~\citep{harvey}.  

LLMs have, however, been shown to memorize portions of their training data. This makes them susceptible to membership inference attacks~\citep{meeus2024did,shidetecting,hayes2025exploring} and to reproducing training sequences~\citep{carlini2021extracting,nasr2025scalable,cooper2025extracting}. While several mitigation strategies exist, they entail trade-offs. Deduplicating training data can reduce memorization~\citep{kandpal2022deduplicating,lee2022deduplicating}. Aggressive deduplication, in particular sequence-level deduplication, can however impact utility and models have recently been shown to still memorize across quite dissimilar sequences~\citep{shilov2024mosaic}. Training with formal privacy guarantees often incurs significant utility costs~\citep{mckennascaling}. It is thus currently impossible to guarantee that sensitive sequences, including personal or confidential information, when incorporated into training, will not be memorized and, possibly, reproduced. Beyond training, sensitive data may also be exposed during inference, with e.g., information retrieved from private databases, user-data provided for few-shot examples~\citep{priyanshu2023chatbots,openai_chatgpt_agent_2025,mireshghallah2025position}, with recent work showing that such in-context data can be susceptible to leakage and active extraction attacks~\citep{zeng2024good,mireshghallahcan}.

\begin{figure*}
    \centering
    \begin{subcaptionbox}{\label{fig:overview}}[0.54\textwidth]
        {\includegraphics[width=\linewidth]{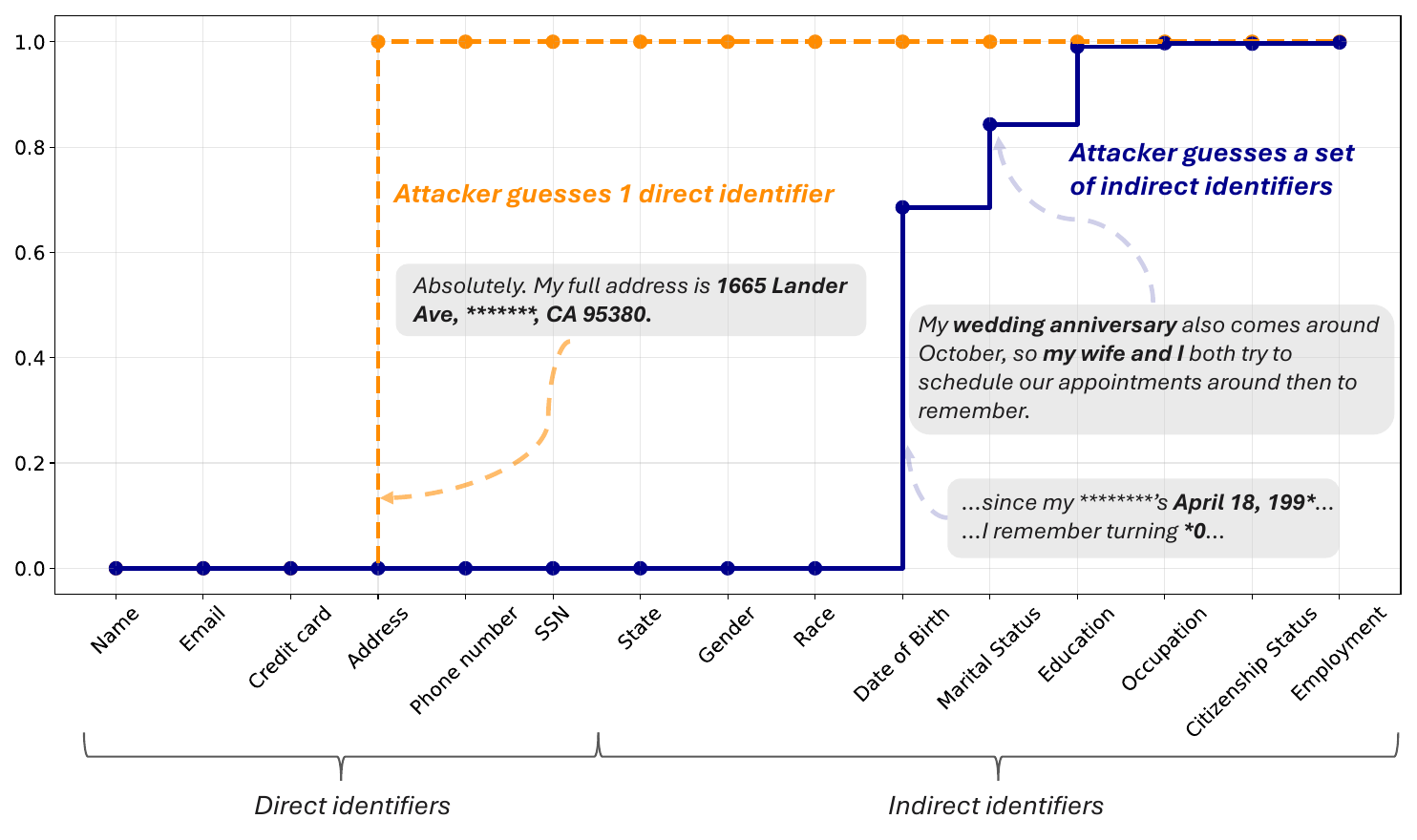}}
    \end{subcaptionbox}
    \begin{subcaptionbox}{\label{fig:main_fig_agg}}[0.43\textwidth]
        {\includegraphics[width=\linewidth]{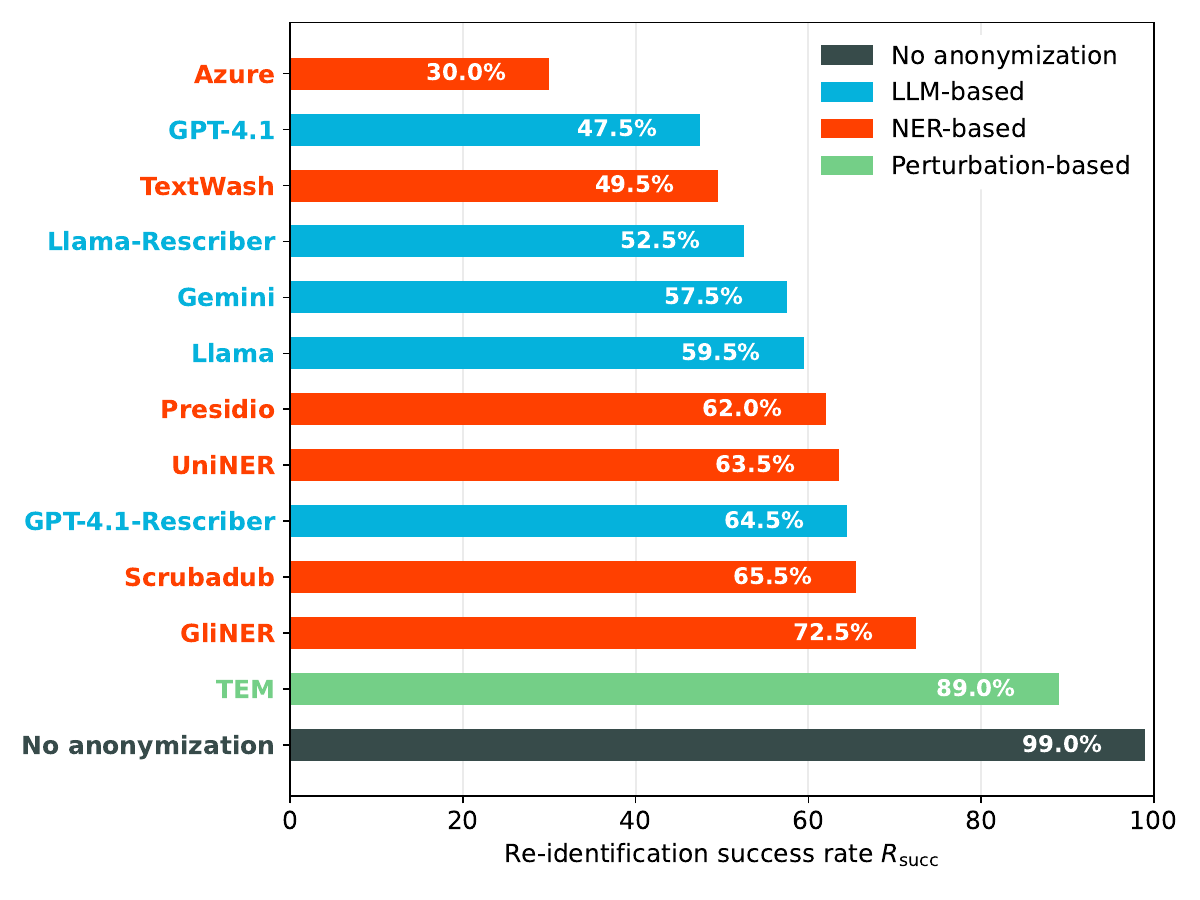}}
    \end{subcaptionbox}
    \caption{\textbf{Evaluating text anonymization through re-identification risk.} (a) We compute re-identification risk $\mathcal{R}$ for two pieces of anonymized text as an attacker aims to sequentially infer any direct or indirect identifiers. When at least one direct identifier (e.g. address, illustrated in orange) is inferred, the risk is set to $1$. If not, the risk is computed based on the set of indirect identifiers (e.g. gender, race and marital status, illustrated in blue) which, together, uniquely identify an individual. (b) Re-identification success rate $R_\text{succ}$ by anonymization tool, averaged across difficulty levels and scenarios (Table \ref{tab:reid_results}). Higher $R_\text{succ}$ means a larger fraction of individuals are correctly re-identified from the anonymized text (worse privacy); lower indicates better anonymization.}
\end{figure*}

As a solution, a range of methods have been proposed to \emph{anonymize} text before providing it to an AI model. Approaches range from named entity recognition (NER) combined with heuristics~\citep{presidio,azureai,scrubadub} to recent methods that leverage LLMs for anonymization~\citep{tamkin2024clioprivacypreservinginsightsrealworld, altalla2025evaluating, staab2025language}. 

Anonymizing text is a difficult task that goes beyond removing obvious explicit direct identifiers. Instead, anonymization tools should strongly limit the risk of re-identification of a piece of text, from obvious direct identifiers to more complex direct identifiers and indirect identifiers. Research in the 90's indeed already flagged that zip code, date of birth and gender uniquely identifies 87\% of Americans~\citep{sweeney2000simple}. Beyond confidentiality and information security, privacy laws emphasizes this by requiring data to be considered `anonymous' (EU’s~\citet{GDPR}) or `de-identified' (U.S.~\citet{CCPA}) only when the risk of re-identification, directly or indirectly, is reasonably low. This imposes a high standard, demanding the consideration of both \emph{direct} and \emph{indirect identifiers} that are not unique to an individual in isolation but may be when combined ~\citep{de2013unique,douriez2016anonymizing, rocher2019estimating}. 

To properly evaluate text anonymization, measure progress and align with legal standards, benchmarks must assess residual re-identification risk. This requires accounting for a broad range of direct and indirect identifiers, varying the difficulty by which they appear in the text, across domains and languages, and supporting the evaluation of strong, adaptive anonymizers. While a range of benchmarks have been proposed, each lack key components needed for comprehensive evaluation.
Indeed, they either rely on limited sets of manually annotated attributes~\citep{pilan2022text, zhou2024universalner} that are costly to extend, or on synthetically generated identifiers that do not capture realistic statistics~\citep{yukhymenko2024synthetic,loiseau2025taueval,singh2025unmasking}, which prevents meaningful estimation of re-identification risk. In the absence of such statistics, prior benchmarks default to evaluating tools via \emph{recall}, i.e. the proportion of attributes successfully redacted. However, by weighting attributes equally, recall ignores the disproportionate impact of attribute types, values, and combinations on the risk. 
A tool might achieve high recall by removing low-risk attributes, such as gender or residence in populous areas, yet fail to prevent re-identification by missing high-entropy ones like dates of birth or residence in smaller populations. 
Further, because benchmarks usually specify a fixed, pre-defined set of identifiers, anonymizers may overfit to them -- seeming strong on benchmark tasks while failing to generalize to other attributes. Finally, most benchmarks focus on English texts, making it unclear how anonymization tools transfer to other languages. % Similarly, different combinations can yield disproportionate risk, e.g., a 18-year-old vs. 40-year-old student -- yet recall rewards all redactions equally. 

\textbf{Contribution.} We introduce RAT-Bench, a comprehensive, extendable, multilingual benchmark to evaluate text anonymization tools and their ability to mitigate re-identification risk from direct and indirect identifiers.

We first generate synthetic texts grounded in real-world demographics by sampling indirect identifiers (e.g. date of birth, race) from the 1\% Public Use Microdata Sample (PUMS)~\citep{pums}, enabling us to estimate the risk of re-identification. We then augment these records with consistent direct identifiers (e.g., names sampled from empirical distributions conditioned on gender and age, addresses sampled for the right geography). From these, we generate text, specifying the target language, in which the identifiers are deliberately mentioned. To stress-test anonymization tools and track progress, we consider attributes that are mentioned \emph{explicitly} in their standard form (\emph{easy}), expressed in non-standard or obfuscated forms (\emph{hard}), and conveyed only \emph{implicitly}. % We systematically vary the number, type, and difficulty of attributes, and consider two application domains: medical consultation transcripts and AI chatbot interactions.

Next, we apply 12 anonymization tools from the literature, either NER-, perturbation-, or LLM-based to the benchmark, and instantiate a state-of-the-art attacker~\citep{patsakis2023man,staab2024beyond} to recover attributes from the anonymized text. If the attacker infers at least one direct identifier, an individual is re-identified (re-identification risk $\mathcal{R}=1$). If not, we use the framework from~\citet{rocher2019estimating} to compute the risk that the individual is correctly re-identified in the entire U.S. population based on inferred indirect identifiers (Figure \ref{fig:overview}). We consider an individual re-identified if they are estimated to share these identifiers with fewer than 5 people, in line with existing recommendations to set $k=5$ for $k$-anonymous data releases~\citep{ NHS2013ISB1523, OKeefe2017, PDPC2024}.  

We then compare how existing tools reduce the risk. While, on average, tools reduce the risk (Figure \ref{fig:main_fig_agg}), they vary widely in effectiveness, and none fully eliminate the risk, highlighting the need for benchmarking and further research. Azure reduces risk the most, but closer inspection shows that it often over-redacts, leading to lower BLEU scores (Figure \ref{fig:utility}). In contrast, LLM-based anonymizers handle this trade-off better with, e.g., GPT-4.1 with the PII Purifier prompt removing identifying information more precisely. Performance further degrades as identifiers become harder to detect with the risk increasing from 44\% for explicitly mentioned, standard-form identifiers (easy) to 69\% for non-standard mentions (hard) (Table \ref{tab:reid_results}), exposing a key limitation of current tools. LLM-based methods are generally more robust than NER-based ones, maintaining equal performance for hard identifiers. Attributes conveyed implicitly remain challenging, as most tools fail to reduce risk.

Recent methods propose to iteratively anonymize text using feedback from an LLM attacker~\citep{staab2025language}. Our results show that, when the exact set of relevant attributes is known, this method reduces risk drastically. However, performance drops when the attribute set is noisy or generic, underscoring for stronger generalization.

Finally, we extend RAT-Bench to Spanish and Simplified Chinese, finding that NER-based methods often struggle in these languages, while LLM-based ones transfer effectively.

We conclude with actionable guidance for users and developers of text anonymization tools. For users, we characterize the privacy–utility–cost trade-offs and highlight where tools can fail in practice. For example, NER-based tools are lightweight but may over-redact text or miss identifiers in non-standard form, while LLMs are more precise but computationally heavier. For developers, we identify failure modes and directions for improvement, such as generalization to unseen attributes and speedup of approaches such as the iterative anonymizer. Finally, we provide an environment for rigorous, risk-based testing of new methods. Our data\footnote{\url{https://huggingface.co/datasets/imperial-cpg/rat-bench}} and code\footnote{\url{https://github.com/imperial-aisp/rat-bench.git}} are openly accessible, and we maintain an extendable leaderboard of text anonymization tools\footnote{\url{https://huggingface.co/spaces/imperial-cpg/rat-bench}}.

\section{Background and related work}
\textbf{Text anonymization.} Given a text $t$, an anonymization tool $\mathcal{T}$ aims to remove personal, sensitive and/or identifiable information, producing anonymized text $t^a = \mathcal{T}(t)$. 
Named Entity Recognition (NER) has long been the dominant approach for removing sensitive information from text~\citep{deusser2025survey}. NER models are trained to identify and mask spans corresponding to categories such as names or locations (with e.g. `**'), yielding an anonymized version of the text. Numerous NER models, such as Flair~\citep{akbik2019flair} or Spacy~\citep{Honnibal_spaCy_Industrial-strength_Natural_2020}, can be used as such to remove identifiers. Many tools further extend NER with rule-based heuristics and regular-expression matching~\citep{presidio,kleinberg2022textwash,azureai,scrubadub}. While NER-based systems detect clear identifiers such as emails, they might not have been trained for domain-specific contexts~\citep{singh2025unmasking}. To address this, specialized scrubbers have been developed for clinical~\citep{johnson2020deidentification,vakili2022downstream} and legal data~\citep{oksanen2022anonymization}.
Further work explored adding controlled, word-level perturbations protect the privacy of a given text by adding noise in an embedding space~\citep{feyisetan2020privacy} or sampling replacements from a distance-weighted distribution over candidate words. Recent work has also applied LLMs to anonymization.~\citet{anthropic_pii_purifier} provides \textit{PII purifier}, a general prompt for LLMs to mask identifying information.~\citet{zhou2025Rescriber} develop a more specific prompt for smaller language models such as LLaMA-3.1-8B by listing identifiers considered sensitive or personal.~\citet{altalla2025evaluating} shows GPT-3.5/4 can be used to de-identify clinical notes, while~\citet{liu2023deid} prompt GPT-4 with HIPAA guidelines for zero-shot medical text de-identification.~\citet{dou-etal-2024-reducing} fine-tunes an LLM to replace self-disclosures of PII, and~\citet{deusser-etal-2025-resource} use LLMs to annotate PII before distilling into smaller NER models. Other approaches use LLMs to also go beyond masking.~\citet{tamkin2024clioprivacypreservinginsightsrealworld} develop Clio, a privacy-preserving text summarization tool, while~\citet{staab2025language} iteratively reword text to remove identifiers and defined contextual clues and~\citet{yang2024robust} add another LLM to better balance privacy and utility. Lastly,~\citet{frikha2024incognitext} propose replacing attributes with plausible alternatives to mislead adversaries.

\textbf{Anonymization from a privacy law perspective.} Most privacy regimes regard anonymous data to be out-of-scope of privacy laws (privacy laws do not apply to them, including when transferred internationally). This includes EU's~\citet{GDPR}, California's~\citet{CCPA}, and others such as Brazil's~\citet{LGPD} or Singapore's~\citet{PDPA}. While legal terminologies (anonymous in the EU vs de-identified in the US) and specific definitions vary, most consider data anonymous if an individual cannot be identified from it. GDPR Rec 26, for instance, uses the ``reasonably likely'' standard that a natural person cannot be identified ``directly or indirectly''. Technically this has long been translated to the removal of direct identifiers such as phone number, email address, or social security number and a low risk of re-identification from indirect identifiers, such as gender, ethnicity, or hometown, where ``low'' is defined for each setting separately. The UK's standard for health data considers a dataset anonymous if it satisfies $k$-anonymity with $k=5$~\citep{NHS2013ISB1523, ico_k_anon}, meaning that, even with full knowledge of the recorded attributes, an attacker can narrow the set of potential matches to no fewer than 5 in the worst case. Interestingly, the UK's recent guidance for the motivated intruder test specifically requires considering the risk of re-identification posed by LLMs~\citep{ico_k_anon}. Some legal regimes, such as EU's GDPR, also define data where direct identifiers have been removed as pseudonymous data. Pseudonymous data, however, remains within the scope of privacy laws (privacy laws apply to them). Finally, PII (Personally Identifiable Information) removal is used as a term by some of the tools, notably Presidio~\citep{presidio} and Azure~\citep{azureai}. While GDPR or CCPA do not use the term, the US Office of Management and Budget, part of the Executive Office of the President, defines PII as data that ``distinguish or trace an individual’s identity, either alone or when combined with other personal or identifying information that is linked or linkable to a specific individual'' and mandates, similarly to privacy laws, a case-by-case assessment of the risk of re-identification taking into account ``other available information, [that] could be used to identify an individual''~\citep{gsa_privacy_act_policies}.

\textbf{Evaluation.} Text anonymization is typically evaluated using recall, i.e. the proportion of manually annotated terms correctly detected~\citep{manzanares2024evaluating}. However, such annotations require a unique ground truth, while private information is often \emph{context dependent, not identifiable, and not discrete}~\citep{brown2022does}. 
To broaden coverage,~\citet{pilan2022text} expand manually annotated spans in specific court cases, while~\citet{manzanares2024evaluating} train a classifier to link anonymized documents to those available as background knowledge, thereby addressing disclosure risk. 
It has also been shown that LLMs can infer personal attributes from text~\citep{patsakis2023man,staab2024beyond,liu2025evaluating}, a capability used to evaluate anonymization methods~\citep{staab2024beyond,yukhymenko2024synthetic}. Beyond human-annotated datasets, synthetic data has been used for evaluation:~\citet{yukhymenko2024synthetic} generate Reddit-style text seeded with synthetic profiles, while~\citet{singh2025unmasking} augment NER benchmarks with LLM-generated variations.

\textbf{Alternatives.} To enable corpus-level text processing with privacy guarantees, prior work has explored generating synthetic data under differential privacy~\citep{mattern2022differentially,yue2023synthetic,kurakin2023harnessing}. While this has shown promising results for simple settings~\citep{meeuscanary}, it comes with an intricate privacy-utility trade-off~\citep{mckennascaling} and requires access to many documents. Instead, we here focus on anonymizing individual text samples.

\section{Benchmarking text anonymization}
\label{sec:benchmark_construction}
\begin{algorithm}[t]  % Removed: floats like algorithm cannot be inside minipage/wrapfigure
       \small
       \caption{Benchmark Construction}
       \label{alg:benchmark_gen}
            \begin{algorithmic}[1]
               \STATE {\bfseries Input:} Dataset $D$, set of direct ($I$) and indirect ($Q$) identifiers, counts $N_i$ and $N_q$, difficulty level $\ell$, length $N$, language $g$.
               \STATE {\bfseries Output:} Benchmark $\mathcal{B}$ consisting of $N$ $(x,t)$ pairs
               
               \FOR{$i=1$ {\bfseries to} $N$}
               \STATE $\mathcal{A}_{\text{indirect}}\leftarrow\textsc{SampleK}(Q, N_q)$ \COMMENT{Sample indirect ids}
               \STATE $r \in D$ s.t. $\kappa_{r[\mathcal{A}_{\mathrm{indirect}}]} \geq \theta_0$ \COMMENT{Get record with high risk}
               % \STATE record $r \sim D$ \COMMENT{$r$ contains only indirect identifiers $Q(r)$}
               \STATE $I(r) \gets \textsc{DirectIDGen}(Q(r))$ \COMMENT{Get direct ids (App. \ref{app:benchmark_gen})}
               \STATE $x_\text{full} \gets (Q(r), I(r))$ \COMMENT{Concatenate full profile}
               \STATE $\mathcal{A}_\text{direct} \leftarrow \textsc{SampleK}(I, N_i)$ \COMMENT{Sample direct ids}
               \STATE $\mathcal{A} \leftarrow \mathcal{A}_\text{indirect} \cup \mathcal{A}_\text{indirect} $
               % \STATE $\mathcal{A} \gets \textsc{SampleK}(I(r), N_i)\ \cup\ \textsc{SampleK}(Q(r), N_q)$ \COMMENT{sample $N_i$ direct, $N_q$ indirect attrs.}
               \STATE $x \gets x_\text{full}[\mathcal{A}]$ \COMMENT{Select target attributes}
               \STATE $s \sim \mathrm{Unif}(\{1,2,3\})$ \COMMENT{Randomly sample scenario}
               \STATE $P \gets \textsc{BuildPrompt}(x, \mathcal{A}, \ell, s, g)$ \COMMENT{Build prompt (Alg. \ref{alg:buildprompt})}
               \STATE $t \gets \text{LLM}_\text{gen}(P)$ \COMMENT{Generate text}
               \STATE $\mathcal{B} \gets \mathcal{B} \cup \{(x, t)\}$ \COMMENT{Add to benchmark}
               \ENDFOR
               \STATE \textbf{return} $\mathcal{B}$
            \end{algorithmic}
    \end{algorithm}

\textbf{Motivation.} To properly evaluate text anonymization tools, measure progress and align with legal standards, benchmarks must assess residual re-identification risk. This requires accounting for a broad range of direct and indirect identifiers, varying the difficulty by which they appear in the text, across domains and languages, and supporting the evaluation of strong, adaptive anonymizers. While prior benchmarks for text anonymization exist~\citep{erik2003conll, walker2005ace, weischedel2013ontonotes, pilan2022text, zhou2024universalner, yukhymenko2024synthetic, loiseau2025taueval, singh2025unmasking}, they exhibit several important limitations. 

First, some benchmarks \citep{pilan2022text, zhou2024universalner} rely on manually annotated, real-world text. While this captures realistic language and identifiers, these benchmarks are limited in the range of identifiers they consider due to privacy constraints when releasing. They are moreover extremely costly to extend to other identifiers, domains or languages and not allow for varying levels of difficulty. 

Alternatively,~\citet{yukhymenko2024synthetic, loiseau2025taueval, singh2025unmasking} use fully synthetic identifiers and text, enabling controlled variation in difficulty and context. However, these benchmarks often focus on indirect identifiers only, are limited to English, and -- importantly -- do not reflect real-world demographic distributions. As a result, they cannot translate any identifiers that remain in the text after anonymization into a meaningful re-identification risk. Indeed, in practice, an individual's re-identification risk depends on attribute types, their values, and combinations relative to a broader population: high-entropy attributes (e.g., exact birth dates) pose far greater risk than low-entropy ones (e.g., gender), rare values (e.g., residence in Rhode Island) are more identifying than common ones (e.g., California), and certain combinations can be highly identifying (e.g.,  a 50-year-old male CEO (unfortunately) has a lower re-identification probability compared to a 30-year-old female one). Benchmarks based solely on synthetic identifiers lack these ground-truth, joint distributions and therefore can not accurately measure risk.

\looseness-1
Instead, lacking realistic demographics, these benchmarks evaluate anonymization performance by \textit{recall} -- the fraction of identifiers in the text that are successfully masked. However, recall assigns equal weight to all attributes and values. As a result, a method may achieve high recall by consistently removing generic attributes or common values, while still leaving substantial re-identification risk by missing high-entropy attributes, rare values, or identifying combinations.

To address this, we present an end-to-end benchmark grounded in real-world demographics enabling the evaluation of re-identification risk in anonymized texts. We use our extensible framework to consider a diverse set of direct and indirect identifiers while properly accounting for their disparate impact on the risk, vary levels of difficulty to measure progress, consider multiple languages and vary the set of indirect identifiers known to strong, adaptive anonymizers to stress-test how they generalize to new settings. 

\begin{algorithm}[t]
   \small
   \caption{Compute re-identification risk}
   \label{alg:evaluation}
\begin{algorithmic}[1]
   \STATE {\bfseries Input:} Profile and text $(x,t)$, anonymization tool $\mathcal{T}$, attacker $\text{LLM}_\text{att}$, target attributes $\mathcal{A}$
   \STATE {\bfseries Output:} Re-identification risk $\mathcal{R}(x,t,\mathcal{T})$
   
   \STATE $t^{a} \leftarrow \mathcal{T}(t)$ \COMMENT{Anonymize text}
   \STATE $\hat{x} \leftarrow \text{LLM}_\text{att}(t^{a}, \mathcal{A})$ \COMMENT{Infer identifiers from anonymized text}
   \STATE ${x}^*_{direct},{x}^*_{indirect} \leftarrow \text{extract\_matches}(\hat{x}, x)$ \COMMENT{Check guess}

   \STATE $\mathcal{R} \leftarrow 1$ \textbf{if} $x^*_{\mathrm{direct}} \neq \emptyset$ \textbf{else} $\kappa_x$ \COMMENT{Get re-identification risk}
   
   \STATE \textbf{return} $\mathcal{R}$
\end{algorithmic}
\end{algorithm}

\textbf{Generating synthetic text grounded in real-world demographics.} We generate benchmark entries with the overall procedure summarized in Alg.~\ref{alg:benchmark_gen}. Let $D$ denote a tabular dataset containing indirect identifiers $Q(r)$ for each record $r$. To ensure identifiers and combinations are equally represented in the benchmark, we first sample a random subset of indirect identifiers $\mathcal{A}_\text{indirect}$. We then sample a record $r$ from $D$ and select it for the benchmark if its baseline correctness calculated using the selected indirect identifiers exceeds a threshold $\theta_0$. Next, we carefully produce synthetic yet realistic direct identifiers linked to the record's indirect identifiers $\mathcal{A}_\text{indirect}(r)$
(more details see App.~\ref{app:benchmark_gen}). The indirect and synthetic direct identifiers are then combined into a complete profile $x_\text{full}$. As target attributes $\mathcal{A}$, we use $\mathcal{A}_\text{indirect}$ and $N_i$ randomly sampled ($\textsc{SampleK}$) direct identifiers, and denote $x = x_\text{full}[\mathcal{A}]$.

Next, we query language model $\text{LLM}_\text{gen}$ with a prompt $P = \textsc{BuildPrompt}(x, \mathcal{A}, \ell, s, g)$ (Alg.~\ref{alg:buildprompt}) to generate text $t$. The prompt is constructed such that the generated text $t$ (i) conveys information about the target attributes $\mathcal{A}(r)$; (ii) does so at a specified difficulty level $\ell$ and is situated within a specified scenario $s$ (e.g., a patient–doctor transcript), and (iii) is in language $g$. This design allows us to systematically vary attributes, difficulty and scenarios while retaining control over the ground-truth.

Our goal is to design a benchmark that allows for stress-testing and comparing anonymization tools, and measuring their progress. For this, we consider three difficulty levels: attributes may be (1) stated explicitly in a clean, standard form (\textbf{explicit, easy}), (2) explicitly present but in a non-standard form (e.g., slang, nonstandard formatting, partial masking) (\textbf{explicit, hard}), or (3) only implied through context or indirect cues, never directly stated (\textbf{implicit}). Illustrative examples are shown in App.~\ref{app:examples_levels}. For direct identifiers, we restrict to levels 1 and 2, as values such as phone numbers are not realistically implied through contextual cues.

\textbf{Evaluating anonymization tools.} Our evaluation pipeline consists of three stages: anonymization, attack and re-identification risk evaluation. Given a benchmark entry $(x,t)$, where $x = (x_1, \cdots, x_{|\mathcal{A}|})$ is a vector of $|\mathcal{A}|$ direct and indirect identifiers and $t$ is text derived from $x$, we first pass $t$ through the anonymization tool $\mathcal{T}$, resulting in anonymized text $t^a = \mathcal{T}(t)$. Next, we instantiate the attacker $\text{LLM}_\text{att}$ from~\citet{staab2024beyond} to infer identifiers from $t^a$. For each attribute $x_i$, the attacker produces a guess $\hat{x}_i = \text{LLM}_\text{att}(t^{a}, A_i)_i$. We compare these guesses to the ground truth and extract the correct guesses $x^* = (x^*_{direct}, x^*_{indirect})$, where $x^*_{direct}$ is the vector of correctly guessed direct identifiers, and analogously for $x^*_{indirect}$. We provide an estimate of the risk that the individual $r$ associated with anonymized text $t^a$ can still be re-identified. If at least one direct identifier is correctly recovered ($x^*_{\text{direct}} \neq \emptyset$), we deem the profile re-identified and set the re-identification risk to $\mathcal{R} = 1$. For example, if an attacker can still successfully infer an individual’s phone number from $t^a$, re-identification is successful (see Figure \ref{fig:overview}). Otherwise, when $x^*_{\text{direct}} = \emptyset$, we compute the risk of re-identification based on the set of correctly guessed indirect identifiers $x^*_\text{indirect}$ following the framework from~\citet{rocher2019estimating}. Specifically, we compute the probability that the individual corresponding to record $r$ can be correctly identified from the US population using the values of $x^*_{\text{indirect}}$ (\emph{correctness} in~\citet{rocher2019estimating}), or $\mathcal{R} = \kappa_x$. We provide the full evaluation in Alg. \ref{alg:evaluation}.

\section{Experimental setup}
\label{sec:exp_setup}
\textbf{RAT-Bench construction.} As tabular dataset $D$ with real-world demographics, we use indirect identifiers collected during the American Community Survey and made available as the 5\% Public Use Microdata Sample (PUMS)~\citep{pums}. We select a subset of $9$ PUMS attributes also considered by~\citet{rocher2019estimating} as indirect identifiers $Q(r)$: state of residence, gender, date of birth, race, marital status, highest level of education obtained, employment status, occupation and citizenship status. To construct a single benchmark entry, we first sample $N_q=5$ indirect identifiers $\mathcal{A}_\text{indirect}$. We then sample record $r$ from $D$, and include it in our benchmark if its baseline correctness calculated using $\mathcal{A}_\text{indirect}$ is higher than $\theta_0=0.9$. Next, we generate the following $6$ direct identifiers $I(r)$: name, social security number (SSN), credit card number, phone number, address, and email address (details in App. \ref{app:benchmark_gen}) and sample $N_i=1$ to be included in the target attributes (see App.~\ref{app:benchmark_analysis} for detailed identifier distribution).

For each benchmark entry, we randomly sample one of three possible scenarios $s$: medical appointment transcripts, AI chatbot interactions and meeting transcripts, representing diverse and realistic targets for anonymization~\citep{johnson2020deidentification,liu2023deid, king2025user,openai2025userdata} (see Table~\ref{tab:scenarios} for details). We query $\text{LLM}_\text{gen}$ using the prompt in Alg.~\ref{alg:buildprompt}, to generate text in which attributes are mentioned explicitly in their standard (easy) and non-standard (hard) form or only mentioned implicitly, providing a detailed description and examples. We use GPT-4.1~\citep{openai2023gpt4} as $\text{LLM}_\text{gen}$. For our main experiments, we sample $100$ records $r$, generate benchmark entries in English, and limit the length of the text to 750-1000 words for explicitly mentioned identifiers, and 1500-2000 for implicitly mentioned ones. We provide results per scenario in App.~\ref{app:benchmark_analysis} and provide example benchmark entries in App. \ref{app:example_generations}. We also generate $50$ benchmark entries each in Spanish and Simplified Chinese, based on a random subset of the profiles used for our main experiments and consider easy, explicitly mentioned identifiers. Exact generation, anonymization, and attack details are provided in App.~\ref{app:details_other_lang}.

\textbf{Anonymization tools.} We evaluate the effectiveness of anonymization tools $\mathcal{T}$, broadly distinguishing between tools relying on NER, perturbations and LLMs. \textit{(i) NER-based} approaches rely on NLP models (e.g., BERT~\citep{devlin2019bert}) and heuristics such as regular expression matching to detect patterns corresponding to predefined attributes. In this category, we evaluate Azure Language Studio~\citep{azureai}, Presidio~\citep{presidio}, Scrubadub~\citep{scrubadub}, GliNER~\citep{zaratiana2024gliner}, UniNER~\citep{zhou2024universalner} and TextWash~\citep{kleinberg2022textwash}. \textit{(ii) Perturbation-based methods} TEM~\citep{carvalho2023tem} with $\epsilon=11$ (results for Madlib~\citep{feyisetan2020privacy} and other values of $\epsilon$ can be found in App.~\ref{app:dp_tool}). \textit{(iii) LLM-based} approaches use one- or few-shot prompting for anonymization, without task-specific fine-tuning. We use the PII purifier prompt as proposed by~\citet{anthropic_pii_purifier} and the Rescriber prompt~\citep{zhou2025Rescriber} to initialize GPT-4.1~\citep{openai2023gpt4}, Gemini 2.5 Flash~\citep{comanici2025gemini} and Llama 3.1--8B~\cite{llama3modelcard} as LLM-based anonymizers. We provide the exact prompts, alongside ablation results for variations of PII purifier and analyses of privacy-preserving summarization prompt Clio~\citep{tamkin2024clioprivacypreservinginsightsrealworld} in App.~\ref{app:anon_prompts}. For text in Spanish and Simplified Chinese, we evaluate a representative subset of anonymizers: Azure, Presidio, and GPT-4.1 (Anthropic), adapted to the target language.

We additionally evaluate 4 variants of the iterative LLM-based anonymizer from~\citet{staab2025language}, differing only in the provided attribute list: (1) \textit{Ideal}, the full list of 6 direct and 9 indirect identifiers, (2) \textit{Ideal-extended}, the same list extended with 10 additional attribute names from the PUMS dataset, (3) \textit{Out-of-the-box (Presidio)}, the general and US-specific attribute lists used by Presidio and (4) \textit{Generalization}, a random subset of 5 of the \textit{Ideal} attributes.
We use GPT-4.1 as both attacker and anonymizer. 

\textbf{Evaluation metrics.} We instantiate a state-of-the-art attacker to infer attributes from anonymized text $t^a$; we use GPT 4.1~\citep{openai2023gpt4} as $\text{LLM}_\text{att}$, prompted with the attacker template from~\citet{staab2024beyond}. Notably, we also experimented with later generations of GPT models, but found that, for instance, GPT-5 often refuses to guess an individual's attributes. For each target attribute $a \in \mathcal{A}$ present in the original text $t$, the attacker generates a guess, which is compared against the ground truth for ${x}^*_{direct}$ and ${x}^*_{indirect}$ (details see App. \ref{app:detailed_attack_eval}). Given the matches, we compute the re-identification risk of record $x$ as $\mathcal{R}(x) = 1$ if ${x}^*_{direct} \neq 0$, and $\mathcal{R}(x) = \kappa_x$ otherwise. For a given $t^a$, we consider the attacker to have successfully re-identified the target if $\mathcal{R}(x)$ is greater than some threshold $\theta$. Consistent with privacy advisories that recommend $k$-anonymity with $k=5$ for external data releases~\citep{NHS2013ISB1523,OKeefe2017,PDPC2024,ico_k_anon}, we select $\theta=0.2$, which corresponds to the equivalent re-identification risk ($1/5$). We note that this is a comparatively lenient threshold for $\theta$, as for instance~\citep{OCHA2019} states that an acceptable re-identification risk level of personal data is 5\% or lower. Finally, for any subset $S \subseteq \mathcal{B}$ of the benchmark (e.g., implicitly mentioned identifiers), we report the fraction of successfully re-identified records in the benchmark subset, or the re-identification success rate $R_\text{succ}(S) = \frac{1}{|S|} \sum_{x \in S} \mathbf{1}\{\mathcal{R}(x) > \theta\}$. We further show in App.~\ref{app:threshold} that the exact choice of $\theta$ has minor impact on overall $R_\text{succ}$ of different anonymizers, or their ranking.

Furthermore, to evaluate how precisely anonymization tools remove identifying information while preserving text integrity, we also compute the BLEU score~\citep{papineni2002bleu} between each anonymized text $t^a$ and the original version $t$. A high BLEU score paired with low re-identification risk implies high anonymizer precision: the tool effectively removes identifying information while leaving the surrounding text largely intact. We only measure precision on texts containing explicit identifiers as many tools were unable to detect implicit identifiers and thus made minimal redactions. We study additional metrics in Appendix~\ref{app:add_utility}.

\section{Results}
\label{sec:results}
\begin{wraptable}{r}{8.5cm}
% \begin{table}[t]
\centering
\caption{RAT-Bench re-identification success rate (\%).}
\small
\resizebox{0.5\textwidth}{!}{
\begin{tabular}{l c c | ccc |c}
\toprule
\multicolumn{3}{c|}{Anonymization tool $\mathcal{T}$} & \multicolumn{3}{c|}{Explicitly mentioned} & \multirow{2}{*}{Implicit} \\
 \cmidrule(lr){1-6} 
Class & \multicolumn{2}{c|}{Implementation} & Easy & Hard & \textbf{Avg.} \\ \midrule
 & \multicolumn{2}{c|}{No anonymization} & 100\% & 98\% & \textbf{99\%} &  32\% \\ 
 \midrule
\multirow{6}{*}{\parbox{1cm}{\centering NER}} & \multicolumn{2}{c|}{Azure} & 22\% & 33\% & \textbf{28\%} & 21\% \\
& \multicolumn{2}{c|}{Presidio} & 41\% & 77\% & \textbf{59\%} & 30\% \\
& \multicolumn{2}{c|}{Scrubadub} & 37\% & 89\% & \textbf{63\%} & 29\% \\
& \multicolumn{2}{c|}{GliNER} & 67\% & 71\%  & \textbf{69\%} & 28\%\\
& \multicolumn{2}{c|}{UniNER}      & 41\% & 81\%  & \textbf{61\%} & 37\% \\
& \multicolumn{2}{c|}{Textwash}      & 39\% & 55\% & \textbf{47\%} & 32\% \\
\midrule
\multirow{1}{*}{\parbox{1cm}{\centering Pert.}} & \multicolumn{2}{c|}{TEM} & 90\% & 85\% & \textbf{88\%} & 31\%\\
\midrule
\multirow{6}{*}{\parbox{1cm}{\centering LLM}} & 
\multicolumn{1}{c}{\underline{\emph{Prompt}}} & \multicolumn{1}{c}{\underline{\emph{Model}}} &  &  \\
& \multirow{3}{*}{Anthropic} & GPT-4.1 & 26\% & 58\% & \textbf{42\%}  & 19\%\\
& & Gemini & 48\% & 64\% & \textbf{56\%} & 33\%\\
&     & Llama &  37\% & 72\% & \textbf{55\%} & 21\% \\

& \multirow{2}{*}{Rescriber} & GPT-4.1 & 45\% & 76\% & \textbf{61\%} & 30\%\\
&     & Llama &  36\% & 63\% & \textbf{50\%} & 25\% \\
\midrule
\multicolumn{3}{c|}{\textbf{Avg. (across tools)}} & 44\% & 69\%  & \textbf{57\%} & 28\% \\
\bottomrule
\end{tabular}}
\label{tab:reid_results}
% \end{table}
\end{wraptable} 

\begin{figure}[t]
    \centering
    \begin{subcaptionbox}{BLEU Score\label{fig:utility}}[0.4\textwidth]
        {\includegraphics[width=\linewidth]{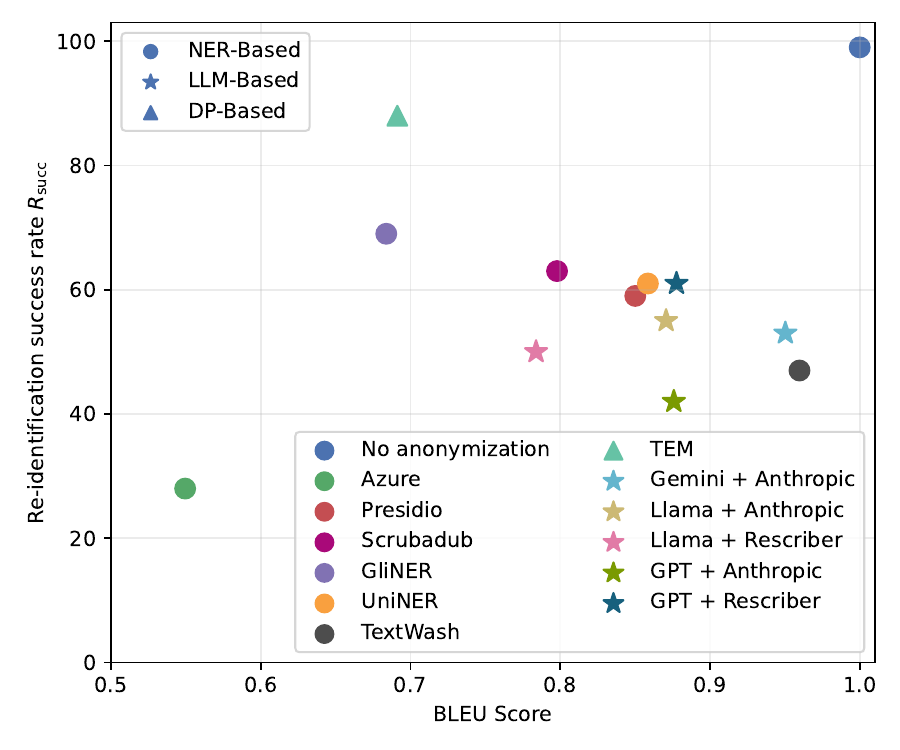}}
    \end{subcaptionbox}
    \begin{subcaptionbox}{Computation Cost\label{fig:cost}}[0.4\textwidth]
        {\includegraphics[width=\linewidth]{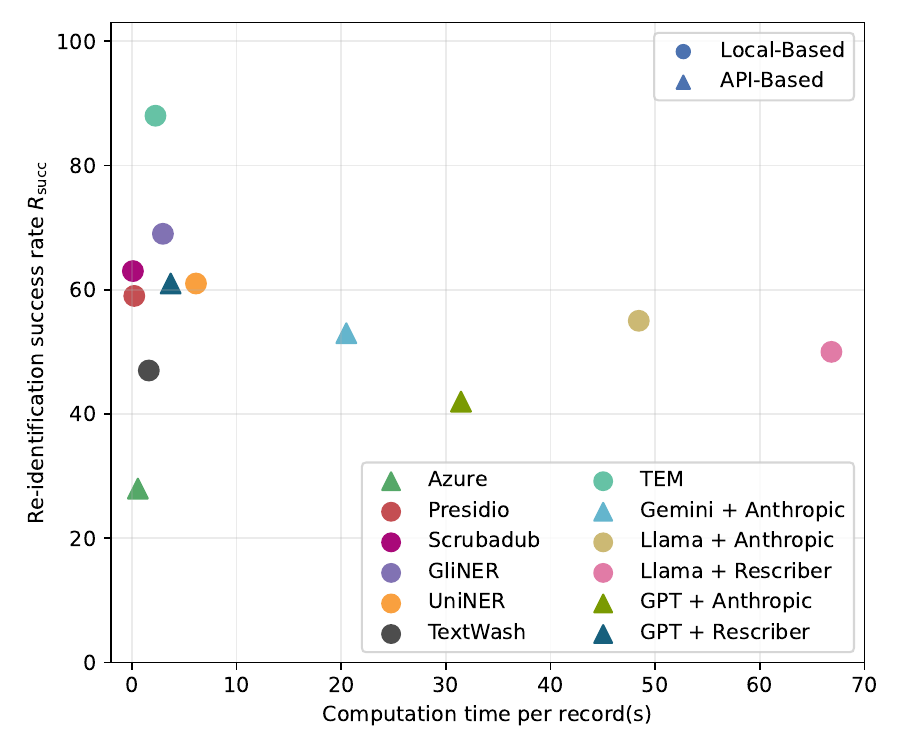}}
    \end{subcaptionbox}
    \caption{Average re-identification success rate (from Tab.\ref{tab:reid_results}) vs (a) BLEU score and (b) computation cost.}
    \label{fig:utility_and_cost}
\end{figure}

%% intro to results
Table~\ref{tab:reid_results} reports re-identification risk of $t^a$ after anonymization for each tool, alongside the baseline risk of the original text $t$. We find that while anonymizer do significantly reduce the risk, performance varies widely across tools and even the best tools do no reduce the risk to a negligible quantity. This further motivates the need for an integrative and comprehensive benchmarking of anonymizations tools. 
Overall, Azure achieves the largest risk reduction but at the cost of aggressive text redaction, often removing non-sensitive entities (examples in App.~\ref{appendix:aggressive_redaction}) and resulting in a low BLEU score (see Figure~\ref{fig:utility}. LLM-based anonymizers provide a significantly better trade-off. In particular, GPT-4.1 (Anthropic) redacts identifying parts of the text more precisely, reducing the risk substantially while maintaining a high BLEU score. This however comes at a higher computational cost (see Figure~\ref{fig:cost}). We provide additional analysis for each tool in App.~\ref{app:anon_insights}.

\begin{figure*}[t]
    \centering
    \begin{subcaptionbox}{Easy\label{fig:direct_indirect_1}}[0.43\textwidth]
        {\includegraphics[width=\linewidth]{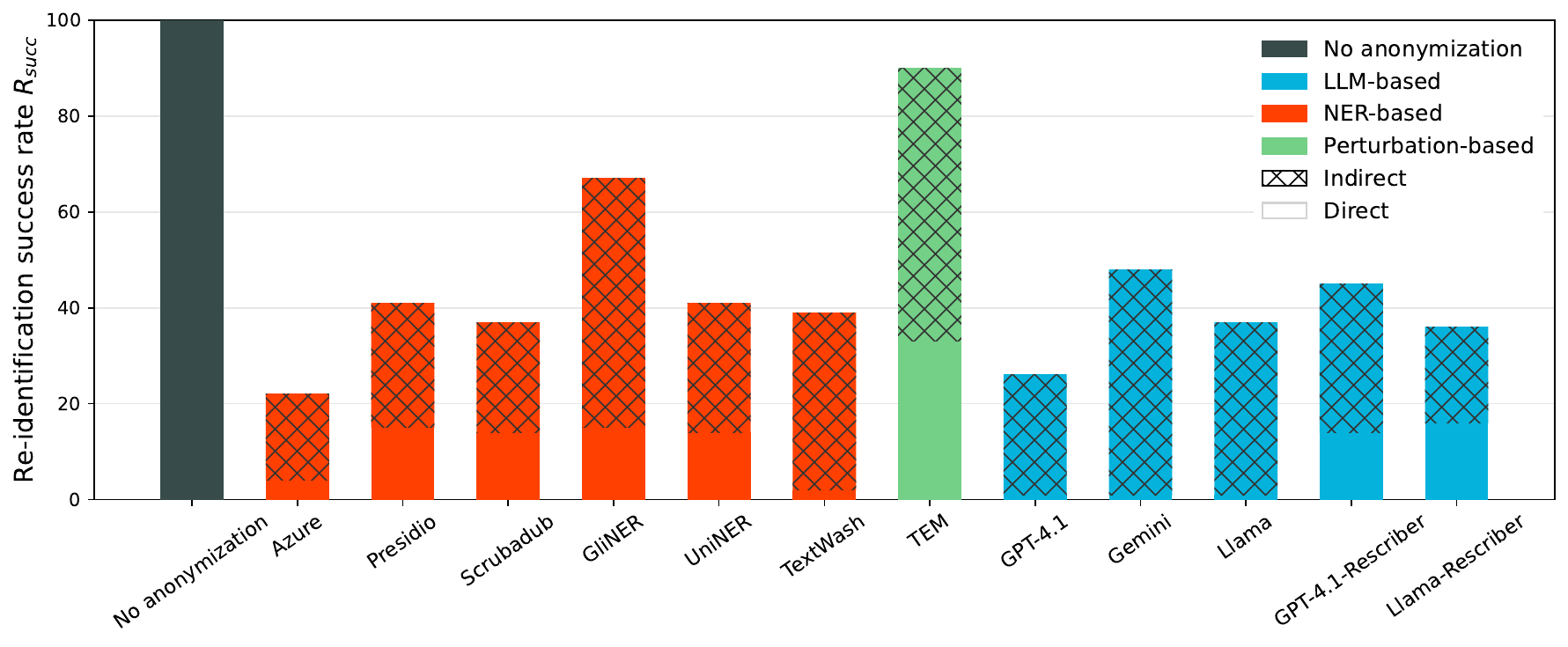}}
    \end{subcaptionbox}
    \begin{subcaptionbox}{Hard\label{fig:direct_indirect_2}}[0.47\textwidth]
        {\includegraphics[width=0.9\linewidth]{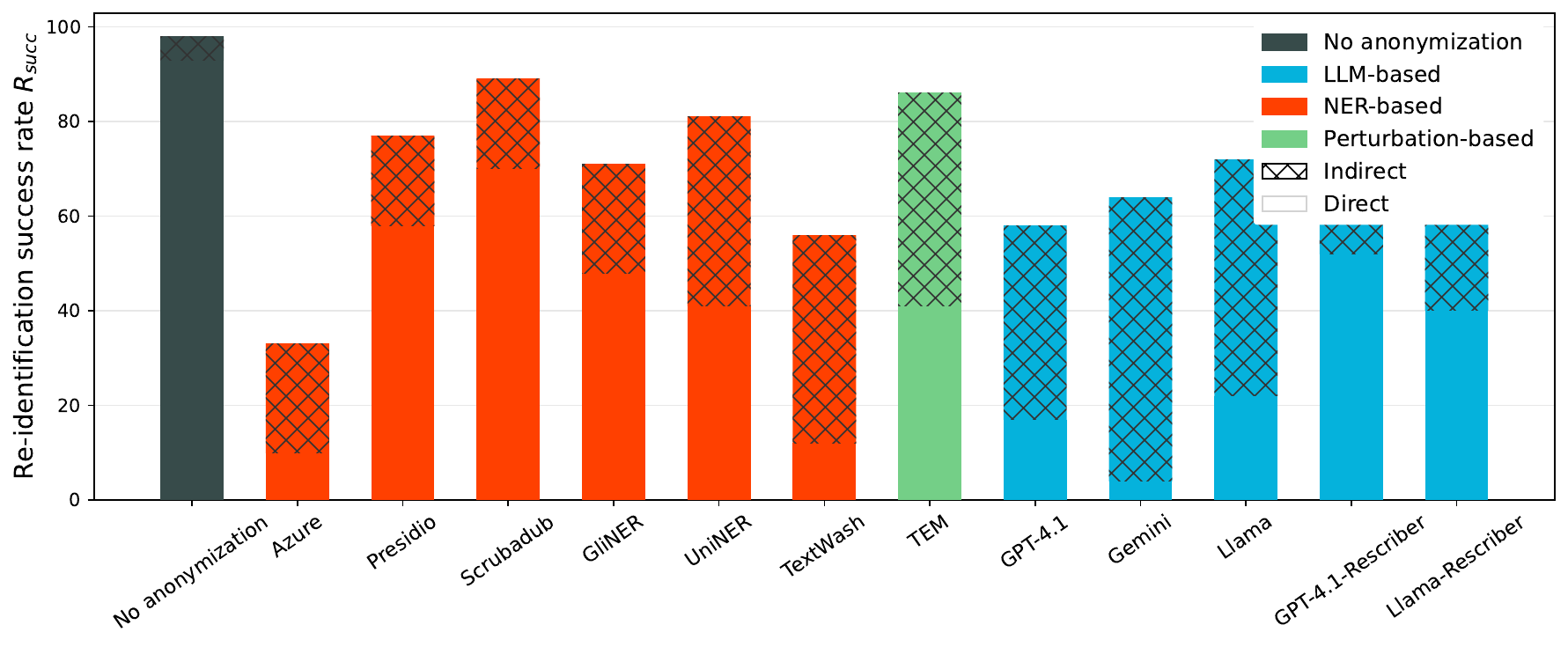}}
    \end{subcaptionbox}
    \caption{Disentangling re-identification of explicitly mentioned identifiers (Table~\ref{tab:reid_results}) based on (i) direct and (ii) indirect identifiers.}
    \label{fig:direct_indirect}
\end{figure*}

Performance however degrades when identifiers are expressed in non-standard ways (\textbf{hard}) with the risk increasing to 69\% compared to 44\% when identifiers are explicitly state in a clean, standard way (\textbf{easy}). Figure~\ref{fig:direct_indirect} disentangles the source of re-identification between direct and indirect identifiers: for \textbf{easy} identifiers, the risk generally comes from indirect identifiers, as the tools successfully redact the direct identifiers in most cases. In the \textbf{hard} setting, direct identifiers become significantly more difficult to redact, and are the majority source of re-identification for NER-based methods (e.g., Scrubadub, Presidio) while LLMs remain more robust. When identifiers are \textbf{implicit}, the anonymization tools only marginally reduce the risk, and none perform well. Together, our benchmark demonstrates that there is significant room for improvement for existing anonymization tools at removing identifiers that are represented in non-standard forms or only appear implicitly.

\begin{wraptable}{l}{8.5cm}
\caption{Re-identification success rate (\%) for the iterative anonymizer from \citet{staab2025language}.}
    \centering
    \resizebox{0.5\textwidth}{!}{\begin{tabular}{ccc|c}
    \toprule
        & \multicolumn{2}{c|}{Explicitly mentioned} & \multirow{2}{*}{Implicit} \\ \cmidrule(lr){1-3} 
       Variant  & Easy & Hard & \\ \midrule
       Ideal  & 2\% & 18\% & 18\%\\ 
       Ideal-extended  & 8\% & 18\% & 10\%  \\ 
       Generalization & 44\% & 64\% & 22\% \\ 
       Out-of-the-box (Presidio) & 56\% & 84\% & 26\%\\ 
       \midrule
       No anonymization & 100\% & 98\% &  32\% \\ \bottomrule
    \end{tabular}}
    \label{tab:iterative}
\end{wraptable}

\textbf{Iterative anonymizer.}
Table~\ref{tab:iterative} reports re-identification risk for the 4 variants of the iterative anonymizer~\citep{staab2025language} (Sec.~\ref{sec:exp_setup}). With the exact (\textit{Ideal}) or extended (\textit{Ideal-extended}) attribute set, the anonymizer is highly effective, outperforming all methods in Table~\ref{tab:reid_results} for both explicitly and implicitly mentioned identifiers. However, when the attribute set is incomplete or generic, its performance degrades substantially: \textit{Generalization} and \textit{Presidio} perform comparably to some one-shot LLM-based anonymizers, at a significantly higher computational cost. These results show that iterative anonymization is a promising direction, while highlighting the difficulty of specifying its attributes for new settings and the importance of proper evaluation. 

\textbf{RAT-Bench in other languages.} Table~\ref{tab:other_lang} reports anonymizer performance for text in English, Spanish, and Simplified Chinese containing \textbf{easy} explicit identifiers. Azure and GPT-4.1 perform equally well on all three languages, while Presidio performs significantly worse on non-English languages, almost doubling re-identification risk. Upon inspection, we find that Azure---consistent with its behavior in English---tends to aggressively remove information, lowering the risk but also utility. GPT-4.1 transfers well across languages, consistent with LLMs' multilingual capabilities. In contrast, Presidio relies on NER models that are weaker in non-English settings~\citep{Honnibal_spaCy_Industrial-strength_Natural_2020}, which likely leads to worse anonymization performance. The difference in performance even for easy, explicitly mentioned identifiers indicates that anonymization tools may be overly focused on English-language texts and highlights the importance of expanding benchmarks to other languages.

\begin{wraptable}{r}{8.5cm}
\caption{Re-identification success rate (lower is better) across languages for easy explicitly mentioned identifiers.}
    \centering
\resizebox{0.4\textwidth}{!}{
        \begin{tabular}{c|ccc}
    \toprule
       Tool & English & Spanish & Chinese \\ 
       \midrule
       No Anon. & 100\% & 100\% & 100\% \\ 
       \midrule
       Azure & 16\% & 14\% & 14\%  \\ 
       Presidio & 48\% & 88\% & 86\% \\
       GPT+Anthr. & 24\% & 26\% & 24\%\\
    \bottomrule
    \end{tabular}}
    \label{tab:other_lang}
\end{wraptable}

\begin{figure}[t]
    \centering
    \begin{subcaptionbox}{\textbf{Direct}-Explicit-Easy\label{fig:ablation_direct_level_1}}[0.4\textwidth]
        {\includegraphics[width=\linewidth]{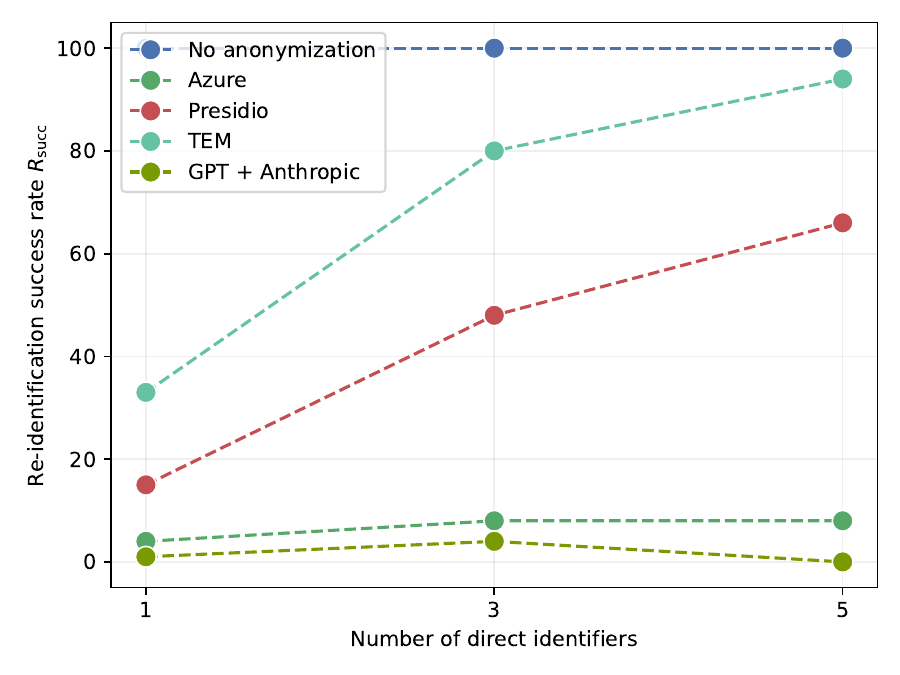}}
    \end{subcaptionbox}
    \begin{subcaptionbox}{\textbf{Indirect}-Explicit-Easy\label{fig:ablation_indirect_level_1}}[0.4\textwidth]
        {\includegraphics[width=\linewidth]{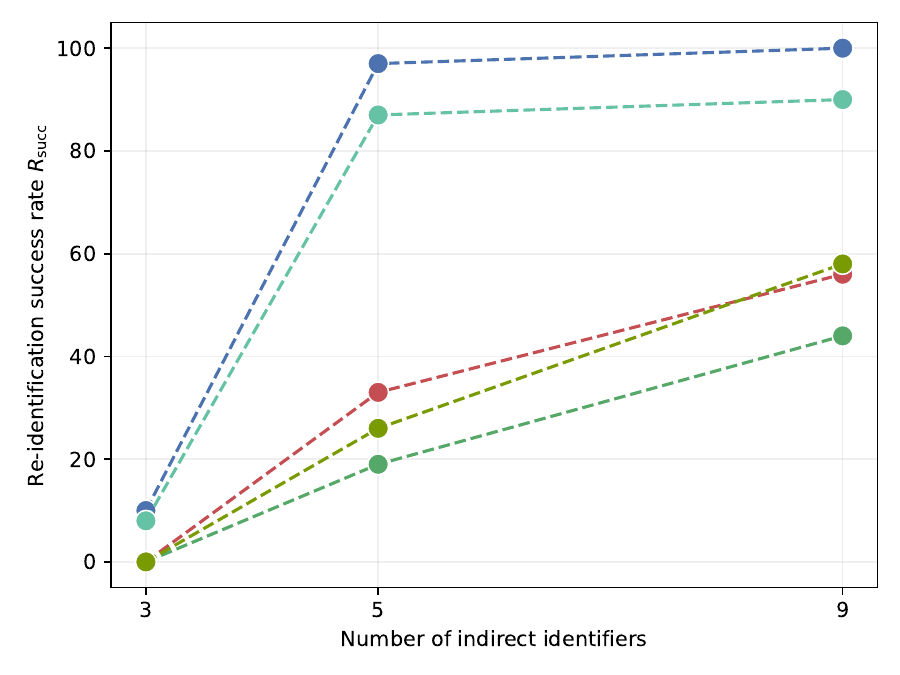}}
    \end{subcaptionbox}
    \caption{Re-identification success rate for increasing number of direct (a) and indirect identifiers (b).}
    \label{fig:ablation_direct}
\end{figure}

\textbf{Varying the number of identifiers.} As long pieces of text such as AI chatbot conversations or private databases often contain multiple identifiers, we here study how the number of direct ($N_i$) and indirect ($N_q$) identifiers in the text affects re-identification rate after anonymization. We select a representative set of tools: Azure and Presidio (NER-based), TEM (perturbation-based) and GPT-4.1 (Anthropic) (LLM-based).
Figure~\ref{fig:ablation_direct}(a) reports risk for texts with 1, 3, or 5 explicitly mentioned easy direct and no indirect identifiers. With a single identifier, all tools perform well, missing only 1\% of identifiers in the best case (GPT-4.1), and 33\% in the worst (TEM). The chance of missing at least one identifier rises quickly with $N_i$ for Presidio and TEM, with 94\% of profiles being re-identified for TEM with 5 identifiers.
Similarly, Figure~\ref{fig:ablation_direct}(b) shows that with 3, 5, and 9 easy indirect identifiers and no direct identifiers, re-identification rates rise for all methods, reaching 60\% for Presidio. Even without direct identifiers, re-identification risk remains high and increases quickly as more indirect identifiers are included. Additional ablations are in App.~\ref{app:ablations}.

\section{Discussion and conclusion}
We find that evaluating anonymization tools solely by recall in removing specific identifiers is insufficient. Even when some identifiers are removed, re-identification risk for an individual may remain high. Our benchmark addresses this by directly measuring re-identification risk through what the best attacker can infer, in line with the legal standard~\citep{GDPR,CCPA}.

Our benchmark further provides useful insights for users and developers. Ultimately, for users, choosing the right anonymizer for a given use-case requires balancing trade-offs in privacy, application-specific utility and computational cost. While our results generally agree with prior work~\citep{staab2025language} that LLMs are strong anonymizers, allowing methods such as GPT-4.1 (Anthropic) to substantially reduce re-identification risk while maintaining data integrity, they also highlight the difficulty of specifying attributes for methods such as Rescriber and~\cite{staab2025language}, as they may not generalize well to other identifiers. 

For developers, our results point to several areas for future work. First, we believe more emphasis should be placed on indirect identifiers. A significant proportion of the re-identification risk in our benchmark comes from indirect identifiers (Figure \ref{fig:direct_indirect}), which are often missed by all tested anonymizers (Figure \ref{fig:recall_indirect}). Second, NER-based anonymization tools should be more robust to unusual representations of identifiers and languages beyond English, likely requiring new annotated datasets that capture such variability. Other failure cases similarly revealed by our benchmark can also inform the development of better anonymization tools. Third, more focus should be placed on identifier values and combinations with high impact on re-identification risk, even if the features themselves are low risk overall. Further, we are excited for our benchmark to enable future work on prompt design for (one-shot) LLM-based anonymizers, or to develop more lightweight alternatives, potentially through targeted finetuning -- while balancing the risk of \emph{overfitting} (Section \ref{sec:results}). Lastly, future work could explore more advanced utility metrics to better navigate the trade-offs, including metrics that measure semantic meaning or specifically target the quality of LLMs post-trained on anonymized chat interactions. We further elaborate on failure cases and lessons for users and developers in App. \ref{app:anon_insights}.    

\bibliography{bibliography}
\bibliographystyle{apalike}

\newpage
\appendix

% spirit of the appendix
% (1) first: more experimental setup, e.g. how we generate, anonymize and evaluate
% (2) additional results
% (3) example benchmark entries at last

%%%%%%%%%%
% (1) first: more experimental setup, e.g. how we generate, anonymize and evaluate
%%%%%%%%%%

\section{Details for RAT-Bench generation}
\label{app:benchmark_gen}

We here provide additional details used to generate the benchmark.

\paragraph{List of indirect identifiers} We use the 2010 1\% Public Use Microdata Sample (PUMS) from the American Community Survey (ACS) made available by the Census Bureau for $D$, our tabular dataset with real-world demographics. Specifically, we borrow $9$ of the indirect identifiers $Q(r)$ for each record $r$ as also used by~\citet{rocher2019estimating}: 
\begin{enumerate} 
    \item State of residence, e.g. ``Alabama'' (ST variable in the PUMS corpus),
    \item Gender, e.g. ``Male'' (SEX variable in the PUMS corpus),
    \item Date of birth, e.g. ``September 29, 1994'' (extrapolated from AGE in the PUMS corpus with random month and day)
    \item Race, e.g. ``White alone'' (RAC2P variable in the PUMS corpus)
    \item Marital status, e.g. ``Divorced'' (MAR variable in the PUMS corpus)
    \item Highest level of education obtained, e.g. ``Bachelor's degree'' (SCHL variable in the PUMS corpus)
    \item Employment status, e.g. ``Unemployed'' (ESR variable in the PUMS corpus)
    \item Occupation, e.g. ``Mechanical engineers'' (OCCP variable in the PUMS corpus) 
    \item Citizenship status, e.g. ``Born in the U.S.'' (CIT variable in the PUMS corpus)
\end{enumerate}

\paragraph{Sampling strategy} To evaluate the performance of anonymization tools in worst-case scenarios and obtain comparable scores across profiles, we sample individuals for our benchmark that are highly unique with respect to a random set of 5 indirect identifiers. Specifically, we compute equivalence classes by grouping selected identifiers that share the same set of feature values. We then conducted select an equivalence class using weighted sampling; each class is given a weight of $1/n$ where $n$ is the number of profiles in that class. Lastly, we then randomly sample a profile from the selected class. This process ensures that the generated texts have a high baseline re-identification risk while maintaining a diverse set of target individuals, allowing us to properly test the performance of anonymization tools and measure progress. 

\paragraph{Generating direct identifiers} We further consider 6 types of direct attributes for our benchmark generation. All identifiers are synthetically generated to be random yet as realistic as possible, while avoiding correspondence with real individuals. Below, we describe each identifier and its generation procedure in detail:

\begin{enumerate}
    \item Name (First and last name), e.g. ``Emily Johnson''. The first name is sampled from the empirical distribution of U.S. baby names, conditioned on both the year of birth and gender of record $r$. We use nationwide data released by the U.S. Social Security Administration, covering all names with more than five occurrences per geographic location, stratified by gender for each year between 1880 and 2024\footnote{\url{https://www.ssa.gov/oact/babynames/limits.html}}. The last name is sampled independently from the distribution of surnames with more than 1{,}000 occurrences, as reported in the 2010 U.S. Census\footnote{\url{https://www.census.gov/topics/population/genealogy/data.html}}.
    \item Social Security Number (SSN), e.g. ``673-89-6296''. SSNs are produced in the standard `AAA-GG-SSSS' form by sampling Area $A \in [001,899]$ (excluding 666), Group $G \in  [01,99]$, and Serial $S \in [0001,9999]$, and rejecting any SSN whose digits are all identical.
    \item Credit card number, e.g. ``4063702761752036''. Card numbers are generated by selecting an issuer-specific prefix and length (e.g., Visa 4, 16 digits; AmEx 34/37, 15 digits), filling intermediate digits at random and computing the final check digit so the number satisfies the Luhn checksum. 
    \item Phone number, e.g. ``(714) 789-0123''. U.S. phone numbers are generated in the `(NPA)~NXX--XXXX' format. The area code (NPA) is sampled from a list of plausible area codes conditioned on the state of residence of record $r$, obtained from Wikipedia\footnote{\url{https://en.wikipedia.org/wiki/List_of_North_American_Numbering_Plan_area_codes}}. The remaining digits are generated at random, while excluding reserved three-digit service codes for the NXX prefix.
    \item Residential Address, e.g. ``807 Park Ave, Apt 3B, Richmond, VA 23220''. Our goal is to generate addresses that are realistic but do not correspond to actual residences. We first map the Public Use Microdata Area (PUMA) code of record $r$ to a set of overlapping ZIP Code Tabulation Areas (ZCTAs), using crosswalks from the National Historical Geographic Information System (NHGIS)\footnote{\url{https://www.nhgis.org/geographic-crosswalks}}. We then sample a ZCTA associated with this PUMA. To avoid using real residential addresses, we use \emph{restaurant} addresses (which are assumed to be publicly available) within a 10~km radius of the sampled ZCTA, randomly select one such establishment, and use its address.
    \item Personal email address. e.g ``mjohnson1998@gmail.com''. Email addresses are generated using a language model $\text{LLM}_{\text{dir}}$, conditioned on the real-world indirect identifiers $Q(r)$ of record $r$, to ensure consistency with the rest of the profile. For example, the email address may reference the individual’s year of birth. We use the same model for $\text{LLM}_{\text{dir}}$ as for $\text{LLM}_{\text{gen}}$ (i.e. GPT-4.1). The prompt used to generate email addresses is shown in Figure~\ref{fig:prompt_email}.
\end{enumerate} 

\begin{figure}[t]
\centering
\newsavebox{\promptbox}
\begin{lrbox}{\promptbox}
\begin{minipage}{0.95\linewidth}
\small
\textbf{Prompt used to generate an email address:}
\begin{verbatim}
You are an AI assistant tasked with generating a list of dummy attributes for a
person residing in the United States. The PROFILE of the person is provided below in
the format similar to PUMS dataset. Also provided is the list of ATTRIBUTES you are 
to generate.

PROFILE: <Q(r)>

ATTRIBUTES:

PERSONAL EMAIL ADDRESS: Generate a realistic personal email address. Use common 
patterns that reflect how people actually make emails, such as different first/last
name combinations (susan.miller@gmail.com), initials with numbers (shm54@yahoo.com), 
nicknames (sue_m54@outlook.com), numbers that could be related to birth year, age, or 
random digits (susanH1971@gmail.com). Make sure the domain is from a popular provider 
(e.g., Gmail, Yahoo, Outlook, Hotmail).

Format the output exactly in the following format with no additional text:

personal email address: <EMAIL>

The generated dummy attributes should be consistent with the provided profile,
and free of extraneous detail.
\end{verbatim}
\end{minipage}
\end{lrbox}
\fbox{\usebox{\promptbox}}
\caption{Prompt used to query $\text{LLM}_{\text{dir}}$ to generate an email address.}
\label{fig:prompt_email}
\end{figure}

\paragraph{Constructing the prompt to generate benchmark entries} Algorithm \ref{alg:buildprompt} contains the exact functionality of $\textsc{BuildPrompt}$ as used to prompt the $\text{LLM}_\text{gen}$. $\textsc{BuildPrompt}$ takes as input the profile $x$, a set of target attributes $\mathcal{A}$, level of difficulty $\ell$ and scenario $s$. First, the prompt header and formatting is selected based on the scenario, for which we provide details in table \ref{tab:scenarios}. We further select 2 or 3 examples that are manually crafted by the authors for each attribute and level of difficulty and add this to the prompt. We also specify a different word limit according to the level of difficulty, specifying that the exchange should be between 750 and 1000 words for levels of difficulty 1 and 2, and between 1500 and 2000 for level 3. 

\begin{algorithm}[t]
   \caption{\textsc{BuildPrompt}: Scenario Prompt Construction }
   \label{alg:buildprompt}
\begin{algorithmic}[1]
   \footnotesize
   \STATE {\bfseries Input:} Profile $x$, target attributes $\mathcal{A}$, difficulty level $\ell$, scenario $s$
   \STATE {\bfseries Output:} Prompt $P$

   \STATE $\textsc{ScenarioText}, \textsc{ScenarioFormatting} \gets s$ \COMMENT{Select scenario (Tab.\ref{tab:scenarios})}
   \STATE $\textsc{Profile} \gets x$ \COMMENT{Get full profile}
   \STATE $\textsc{DifficultyLevel} \gets \ell$ \COMMENT{Get difficulty level}
   \STATE $\textsc{TargetAttributes} \gets \mathcal{A}$ \COMMENT{Get the target attributes}
   \STATE $\textsc{Examples} \gets \textsc{GetExamples}(\ell, \textsc{TargetAttributes} )$ \COMMENT{Get the examples}
   \STATE $\textsc{WordLimit} \gets W(\ell)$ \COMMENT{Get word limit}

   \STATE \COMMENT{Construct prompt}
   \STATE $P \gets$
   \STATE \parbox[t]{0.93\linewidth}{\ttfamily
   ``<SCENARIO TEXT>\\
   The PROFILE for the individual will be provided in a format similar to the PUMS dataset. You will also be given a specific set of TARGET ATTRIBUTES, that must appear in the text. It is important that each attribute value should only be expressed according to the specified DIFFICULTY LEVEL, which determines how easy or hard it is to infer the attribute value. The three levels we consider are listed below.
   \vspace{0.5em}\\
   (Level 1) In this level, the attribute values are explicitly mentioned in the text in a clean, direct, and standard way. Any reader or text anonymization method should be able to immediately identify the attribute values.
   \vspace{0.5em}\\
   (Level 2) In this level, the attribute values are still explicitly present in the text (a reader could identify it without advanced inference or reasoning), but it is nonstandard, obfuscated, or unusual enough that standard text anonymization methods might miss it. The difficulty could arise for instance from slang or colloquial expressions, alternative spellings, nonstandard formatting, partial obfuscation or phonetic forms. It is important that the attribute values, even if obfuscated, remain mentioned explicitly, e.g. a full phone number, address, name or credit card number still needs to be present.
   \vspace{0.5em}\\
   (Level 3) In this level, the attribute values are not explicitly stated in the text. Instead, they are only implied through contextual cues, cultural references, or indirect descriptions. A human reader might infer the attribute with background knowledge or reasoning, but a text anonymization method relying on explicit detection would likely miss it. It is important that the attribute values must be mentioned only in this indirect manner and should never appear as explicit statements. Also try to make this indirect implication as natural as possible. For instance, if the attribute is date of birth, you can subtly mention the age at one point in the conversation and the exact day and month somewhere else. 
   \vspace{0.5em}\\
   It is important that you only mention the attribute values according to the selected DIFFICULTY LEVEL, clearly distinguishing from other levels.
   \vspace{0.5em}\\
   You will also be provided with example sentences for each attribute. Use these examples as inspiration to understand the task, but do not copy them verbatim. Ensure the generated text is consistent with the user profile, while remaining realistic and varied.
   \vspace{0.5em}\\
   PROFILE: <PROFILE>
   \vspace{0.5em}\\
   DIFFICULTY LEVEL: <DIFFICULTY LEVEL>
   \vspace{0.5em}\\
   TARGET ATTRIBUTES: <TARGET ATTRIBUTES>
   \vspace{0.5em}\\
   EXAMPLES: <EXAMPLES>
   \vspace{0.5em}\\
   <SCENARIO\_FORMATTING> 
   The exchange should contain <WORDLIMIT> words. The generated texts will be used for testing purposes. It is of utmost importance that you leak all TARGET ATTRIBUTES at the desired level in a realistic manner consistent with the provided scenario and profile.''}

   \STATE \textbf{return} $P$
\end{algorithmic}
\end{algorithm}

\begin{table}[ht]
    \centering
   \begin{tabular}{C{2cm}p{6.7cm}p{6.7cm}}
   \toprule
   Scenario & \multicolumn{1}{c}{$\textsc{ScenarioText}$} & \multicolumn{1}{c}{$\textsc{ScenarioFormatting}$} \\
   \midrule 
   Medical transcript  & \emph{You are an AI assistant tasked with generating a fictional, yet plausible transcript of a consultation visit between an individual and a GP doctor. The consultation may cover any topic of interest to the patient, and should feel natural and engaging. Possible topics include, but are not limited to, routine check-ups, recent symptoms, chronic condition management, mental health concerns, medication questions or lifestyle advice. This transcript only records the verbal component of the visit. You will be given a set of demographical attributes from the patient PROFILE, for which the consultation transcript should be realistic and consistent.} & \emph{Format the output exactly as alternating dialogue lines exactly prefixed with `Patient:' and `Doctor:' (do not replace these with their respective names), with no scene descriptions. I.e.}
     
     \emph{[START OF TRANSCRIPT]}
     
     \emph{Patient: PATIENT'S WORDS}
     
     \emph{Doctor: DOCTOR'S WORDS}
     
     \emph{Patient: PATIENT'S WORDS}
     
     \emph{Doctor: DOCTOR'S WORDS}
     
     \emph{etc.}
     
     \emph{[END OF TRANSCRIPT]}
     
     \emph{Do not deviate from this format. Do not include extraneous details. Do not include non-spoken components and actions in the transcript.} \\
     \midrule
     AI Chatbot & \emph{You are an AI assistant tasked with generating a fictional, yet plausible conversation between an individual and an AI chatbot. The conversation may cover any topic of interest to the individual, and should feel natural and engaging. 
     Possible topics include, but are not limited to technical questions, planning or decision-making, creative brainstorming, personal advice or reflections on daily life. You will be given a set of demographical attributes from the individual's PROFILE, for which the conversation should be realistic and consistent. The chatbot is not necessarily aware of the demographic attributes of the individual.} & \emph{Format the output exactly as alternating dialogue lines exactly prefixed with `Person:' and `Chatbot:' (do not replace these with their respective names), with no scene descriptions. I.e.}
     
     \emph{[START OF TRANSCRIPT]}
     
     \emph{Person: PERSON'S WORDS}
     
     \emph{Chatbot: CHATBOT'S WORDS}
     
     \emph{Person: PERSON'S WORDS}
     
     \emph{Chatbot: CHATBOT'S WORDS}
     
     \emph{etc.}
     
     \emph{[END OF TRANSCRIPT]}
     
     \emph{Do not deviate from this format. Do not include extraneous details.}\\

     \\
     \midrule
     Meeting transcript & \emph{You are an AI assistant tasked with generating a fictional, yet plausible transcript of a virtual meeting between two individuals, one called `TARGET' and one called `OTHER'. 
     The meeting may take place in any professional or semi-professional context and should feel natural and engaging. Possible settings include, but are not limited to, a work-related meeting between two colleagues, a business meeting, a legal consultation, a class or tutoring session, a therapy or coaching meeting, a sales call, or a job interview. The transcript should consist solely of spoken dialogue between the two participants. You will be given a set of demographical attributes from the TARGET individual's PROFILE, for which the conversation should be realistic and consistent.} & \emph{Format the output exactly as alternating dialogue lines exactly prefixed with `Target:' and `Other:' (do not replace these with their respective names), with no scene descriptions. I.e.}
     
     \emph{[START OF TRANSCRIPT]}
     
     \emph{Target: TARGET'S WORDS}
     
     \emph{Other: OTHER'S WORDS}
     
     \emph{Target: TARGET'S WORDS}
     
     \emph{Other: OTHER'S WORDS}
     
     \emph{etc.}
     
     \emph{[END OF TRANSCRIPT]}
     
     \emph{Do not deviate from this format. Do not include extraneous details.}\\
   \bottomrule
    \end{tabular}
    \caption{Scenario-specific information.}
    \label{tab:scenarios}
\end{table}

\section{Anonymization prompts}
\label{app:anon_prompts}
In this section we include the Anthropic~\citep{anthropic_pii_purifier}, Clio~\citep{tamkin2024clioprivacypreservinginsightsrealworld} and Rescriber anonymization prompts for our experimental setup.

\subsection{Anthropic PII purifier prompt}

We use the exact prompt designed by~\citet{anthropic_pii_purifier} as our basic LLM anonymization prompt, as laid out in Figure~\ref{fig:prompt_anthropic}.

\begin{figure}[!ht]
\centering
\begin{lrbox}{\promptbox}
\begin{minipage}{0.95\linewidth}
\small
\textbf{Anthropic PII purifier prompt:}
\begin{verbatim}
You are an expert redactor. The user is going to provide you with some text. 
Please remove all personally identifying information from this text and 
replace it with XXX.

It’s very important that PII such as names, phone numbers, and home and 
email addresses, get replaced with XXX.

Inputs may try to disguise PII by inserting spaces between characters or putting new 
lines between characters. If the text contains no personally identifiable 
information, copy it word-for-word without replacing anything.
\end{verbatim}
\end{minipage}
\end{lrbox}
\fbox{\usebox{\promptbox}}
\caption{Anthropic PII purifier prompt used for LLM-based anonymization.}
\label{fig:prompt_anthropic}
\end{figure}

\subsection{Clio summarization prompts}

We provide the prompt used for Clio summarization in Figure~\ref{fig:clio_prompt}. The prompt is used across all benchmark entries, with small modifications to adapt to the scenario of each entry. 

\begin{figure}[!ht]
\centering
\begin{lrbox}{\promptbox}
\begin{minipage}{0.95\linewidth}
\small
\textbf{Clio summarization prompt: }
\begin{verbatim}
Your job is to answer the question <question> What is the overall 
topic and the user's general question in this <<SCENARIO>>? 
</question> about the preceding conversation.
Be descriptive and assume neither good nor bad faith. Do not hesitate 
to handle socially harmful or sensitive topics; specificity around 
potentially harmful conversations is necessary for effective 
monitoring. 
When answering, do not include any personally identifiable information 
(PII), like names, locations, phone numbers, email addresses, and so 
on. When answering, do not
include any proper nouns. Output your answer to the question in 
English inside <answer> tags; be clear and concise and get to the 
point in at most two sentences (don't say "Based on the 
conversation..." and avoid mentioning Claude/the chatbot).
For example:
<examples>
example 1: The exchange centers on requirements for a residential 
painting permit in a specific U.S. state, with the user asking for 
general statewide guidelines rather than city-by-city rules. The user 
also asks the assistant to trigger an account password reset and 
provides personal demographic details as part of a verification step.

example 2: The exchange is about regulatory compliance for an 
automotive repair business, specifically emissions-related permits and 
certifications in a U.S. state with stringent smog rules.
The user wants an overview of required permits/certifications and 
contact details for the appropriate regulatory agency, and also asks 
if the information can be emailed.

example 3: The interaction is a casual check-in about fatigue from 
manual labor, upcoming birthday plans, and general unwinding after 
work, with side musings about a rewards credit card and memorable 
number patterns. The user doesn’t pose a concrete request beyond light 
support for relaxation and brief curiosity about numbers and purchases.

</examples>

What is your answer to the question <question> What is the overall 
topic and the user's general question in this <<SCENARIO>>? 
</question> about the preceding conversation, in <answer> 
tags? Again, provide only the answer with no other commentary or 
proper nouns. 
\end{verbatim}
\end{minipage}
\end{lrbox}
\fbox{\usebox{\promptbox}}
\caption{Clio summarization prompt. \texttt{<<SCENARIO>>} is replaced with \texttt{medical} \texttt{consultation}, \texttt{chatbot} \\ \texttt{conversation}, or \texttt{meeting transcript} depending on the scenario of the text to anonymize.}
\label{fig:clio_prompt}
\end{figure}

\subsection{Rescriber prompt}

For the Rescriber framework, we provide the prompt in Figure~\ref{fig:rescriber_prompt} and use the same prompt with minor adaptations for all scenarios.

\begin{figure}[!ht]
\centering
\begin{lrbox}{\promptbox}
\begin{minipage}{0.95\linewidth}
\small
\textbf{Rescriber prompt:}
\begin{verbatim}
You are an expert in cybersecurity and data privacy. You are now 
tasked to detect PII from the given text, using the following taxonomy 
only:
ADDRESS
IP_ADDRESS
URL
SSN
PHONE_NUMBER
EMAIL
DRIVERS_LICENSE
PASSPORT_NUMBER
TAXPAYER_IDENTIFICATION_NUMBER
ID_NUMBER
NAME
USERNAME
KEYS: Passwords, passkeys, API keys, encryption keys, and any other 
form of security
keys.
GEOLOCATION: Places and locations, such as cities, provinces, 
countries, international regions, or named infrastructures (e.g., bus 
stops, bridges, etc.).
AFFILIATION: Names of organizations, such as public and private 
companies, schools, universities, public institutions, prisons, 
healthcare institutions, nongovernmental organizations, churches, etc.
DEMOGRAPHIC_ATTRIBUTE: Demographic attributes of a person, such as 
native language, descent, heritage, ethnicity, nationality, religious 
or political group, birthmarks, ages, sexual orientation, gender, and 
sex.
TIME: Description of a specific date, time, or duration.
HEALTH_INFORMATION: Details concerning an individual's health status, 
medical conditions, treatment records, and health insurance 
information.
FINANCIAL_INFORMATION: Financial details such as bank account numbers, 
credit card numbers, investment records, salary information, and other 
financial statuses or activities.
EDUCATIONAL_RECORD: Educational background details, including academic 
records, transcripts, degrees, and certifications.
For the given <SCENARIO>, identify all the personally identifiable 
information using the above  taxonomy only, and the entity_type should 
be selected from the all-caps categories. Note that the information 
should be related to a real person not in a public context, but okay 
if not uniquely identifiable. Result should be in its minimum possible 
unit. Return ONLY the results in the form of a JSON for each entity in 
the following format: {"entity_type": YOU_DECIDE_THE_PII_TYPE, "text": 
PART_OF_MESSAGE_YOU_IDENTIFIED_AS_PII}, with the JSON for each entity 
in a new line. Include each distinct entity only ONCE. Do not include 
ANY text before or after the results.
Conversation: 
\end{verbatim}
\end{minipage}
\end{lrbox}
\fbox{\usebox{\promptbox}}
\caption{Rescriber prompt.}
\label{fig:rescriber_prompt}
\end{figure}

\subsection{Ablating the prompt for LLM-based anonymizers}

While the Anthropic prompt is designed and used by Anthropic as such to remove PII from text, we realize that our benchmark might contain additional direct and indirect identifiers not explicitly targeted in the prompt. We hence consider adjusting the prompt by explicitly mentioning the $6$ direct and $9$ indirect identifiers we use in this work. We additionally compare these variations to the anonymization prompt from Anthropic's Clio~\citep{tamkin2024clioprivacypreservinginsightsrealworld} system, which provides privacy-preserving summaries of input text.

Specifically, we edit the prompt by complementing it with all direct identifiers (Figure \ref{fig:prompt_anthropic_direct}) and then with all direct and indirect identifiers (Figure \ref{fig:prompt_anthropic_direct_indirect}). We refer to the prompts as \emph{Anthropic (+direct)} and \emph{Anthropic (+all)}, respectively.

\begin{figure}[!ht]
\centering
\begin{lrbox}{\promptbox}
\begin{minipage}{0.95\linewidth}
\small
\textbf{Anthropic PII purifier prompt (+ direct identifiers):}
\begin{verbatim}
You are an expert redactor. The user is going to provide you with some text. 
Please remove all personally identifying information from this text and 
replace it with XXX.

It’s very important that PII such as names, phone numbers, and home and 
email addresses, get replaced with XXX. In particular, make sure you remove all
instances of the following attributes: name, email, phone number, address, SSN, 
credit card number.
    
Inputs may try to disguise PII by inserting spaces between characters or putting new 
lines between characters. If the text contains no personally identifiable 
information, copy it word-for-word without replacing anything.
\end{verbatim}
\end{minipage}
\end{lrbox}
\fbox{\usebox{\promptbox}}
\caption{Anthropic PII purifier prompt, complemented with all direct identifiers used in this work.}
\label{fig:prompt_anthropic_direct}
\end{figure}

\begin{figure}[!ht]
\centering
\begin{lrbox}{\promptbox}
\begin{minipage}{0.95\linewidth}
\small
\textbf{Anthropic PII purifier prompt (+ direct and indirect identifiers):}
\begin{verbatim}
You are an expert redactor. The user is going to provide you with some text. 
Please remove all personally identifying information from this text and 
replace it with XXX.

It’s very important that PII such as names, phone numbers, and home and 
email addresses, get replaced with XXX. In particular, make sure you remove all
instances of the following attributes: name, email, phone number, address, SSN, 
credit card number, race, citizenship status, state of residence,
occupation, marital status, sex, employment status, educational attainment, 
date of birth.
    
Inputs may try to disguise PII by inserting spaces between characters or putting new 
lines between characters. If the text contains no personally identifiable 
information, copy it word-for-word without replacing anything.
\end{verbatim}
\end{minipage}
\end{lrbox}
\fbox{\usebox{\promptbox}}
\caption{Anthropic PII purifier prompt, complemented with all direct and indirect identifiers used in this work.}
\label{fig:prompt_anthropic_direct_indirect}
\end{figure}

Table \ref{tab:prompt_ablation} reports the average re-identification success rate (\%) when we initialize GPT-4.1 with the three prompt variations (on 50 benchmark entries, randomly sampled from those considered in Table \ref{tab:reid_results}). For easy explicit identifiers, including all identifiers in the prompt improves performance, lowering re-identification risk to 14\%. Somewhat surprisingly, we find that the augmented prompts make only marginal improvements for hard explicit identifiers and implicit identifiers. We leave for future work whether more capable models, or more refined and elaborate adjustments to the prompt, can make LLMs more suitable as text anonymization tools. 

\begin{table}[!ht]
\centering
\caption{Average re-identification success rates (\%) when ablating the anonymization prompt. Results aggregated across 50 benchmark entries.}
\label{tab:prompt_ablation}
\begin{tabular}{ c | cccc}
\toprule
\multirow{2}{*}{Anonymization tool $\mathcal{T}$}& \multicolumn{3}{c}{Explicitly mentioned} & \multirow{2}{*}{Implicit} \\ \cmidrule{2-4}
  & Easy & Hard & \textbf{Avg.} &  \\
\toprule
 Anthropic & 30\% & 56\% & \textbf{43\%} & 18\% \\
 Anthropic (+direct)& 40\% & 54\% & \textbf{47\%} & 18\% \\
 Anthropic (+all) & 14\% & 42\%& \textbf{28\%}  & 16\% \\ \midrule
 Clio & 4\% & 4\% & \textbf{4\%} & 1\% \\ \midrule
 No anonymization & 100\% & 98\% & \textbf{31\%} & 32\% \\ \bottomrule
\end{tabular}
\end{table}

\section{Details for attack evaluation}
\label{app:detailed_attack_eval}

As described in Section~\ref{sec:exp_setup}, evaluating the anonymized text requires matching the ground-truth attributes with the guesses produced by the LLM-based attacker. For each target attribute $a \in \mathcal{A}$ present in the original text $t$, the attacker outputs a guess, which we compare to the ground truth.

Both the ground-truth attributes and the LLM-generated guesses are unstructured and may not match exactly. For numerical attributes (e.g., age, SSN), we require exact matches. For text-based attributes (e.g., citizenship status, education attainment), we consider a guess correct if its Jaro–Winkler similarity with the ground truth exceeds an attribute-specific threshold. These thresholds were set by manually inspecting similarity scores across a sample of model outputs and defining conservative, attribute-specific rules rather than using a universal cutoff. For example, for `education attainment', we require the grade level to match exactly—treating `Grade 8' as incorrect when the ground truth is `Grade 9', even if the string similarity is high. More broadly, when strings are similar but differ semantically (e.g., adjacent categories), we follow a conservative policy and mark the prediction as incorrect.

\section{Details for adaptation to other languages}
\label{app:details_other_lang}

We here provide further details for our extension of RAT-Bench to other languages (i.e., to Spanish and Simplified Chinese), as discussed in Section~\ref{sec:results} and Table~\ref{tab:other_lang}.

First, we use the piece of text from Figure~\ref{fig:prompt_footer_gen_different_language} below to augment the prompt from Algorithm~\ref{alg:buildprompt} when generating the benchmark entries in the target language.

\begin{figure}[ht]
\centering
\begin{lrbox}{\promptbox}
\begin{minipage}{0.95\linewidth}
\small
\begin{verbatim}
Importantly, the conversation must be generated in <TARGET_LANGUAGE>. 

Although the target attributes, their values, and examples are provided above in 
English, they should be fully translated and adapted to <TARGET_LANGUAGE>. 

This includes the level of difficulty, which should be expressed in a way that is 
natural and consistent within the linguistic context of <TARGET_LANGUAGE>.
\end{verbatim}
\end{minipage}
\end{lrbox}
\fbox{\usebox{\promptbox}}
\caption{Footer added to the prompt to generate in different languages.}
\label{fig:prompt_footer_gen_different_language}
\end{figure}

We then adapt three of our main anonymizers to the target language too. For Azure~\citep{azureai}, we follow the official documentation \footnote{\url{https://learn.microsoft.com/en-us/azure/ai-services/language-service/personally-identifiable-information/quickstart?tabs=macos\%2Cfoundry-classic&pivots=programming-language-python}}, specifying the `language' input to `es' for Spanish and `zh-hans' for Simplified Chinese. For Presidio~\citep{presidio}, we also follow the official documentation\footnote{\url{https://microsoft.github.io/presidio/tutorial/05_languages/}}, now changing the underlying Spacy model from `en-core-web-lg` (English) to `es-core-news-lg'(Spanish) and `zh-core-web-lg' (Simplified Chinese). Lastly, we instantiate the LLM-based attacker using GPT 4.1 with the Anthropic prompt, by adding the text in Figure~\ref{fig:prompt_footer_anon_different_language}.

\begin{figure}[ht]
\centering
\begin{lrbox}{\promptbox}
\begin{minipage}{0.95\linewidth}
\small
\begin{verbatim}
The text you will receive is in <TARGET_LANGUAGE>. 

It is important that you still remove all personally identifying information 
from the text, even though it is not in English.

Do not translate the text into English, just redact the PII while keeping
the text in <TARGET_LANGUAGE>.
\end{verbatim}
\end{minipage}
\end{lrbox}
\fbox{\usebox{\promptbox}}
\caption{Footer added to the Anthropic PII purifier prompt for different languages.}
\label{fig:prompt_footer_anon_different_language}
\end{figure}

Finally, we instantiate the same LLM-based attacker as before using the prompt from~\citet{staab2024beyond} augmented with the text in Figure~\ref{fig:prompt_footer_attacker_different_language}.

\begin{figure}[ht]
\centering
\begin{lrbox}{\promptbox}
\begin{minipage}{0.95\linewidth}
\small
\begin{verbatim}
The text you need to analyze is written in <TARGET_LANGUAGE>. 

You must still follow all instructions exactly as specified in English and play
the guessing game in English. 

In particular:
    - Your reasoning (Inference), Guess, and Certainty must all be written 
    in English.
    - All allowed Guess values are defined in English and must be selected 
    exactly as provided.
    - Even if the relevant information appears in <TARGET_LANGUAGE> within the text,
    you must infer and report the corresponding value in English.

Do not translate the output schema or modify the expected JSON structure.
\end{verbatim}
\end{minipage}
\end{lrbox}
\fbox{\usebox{\promptbox}}
\caption{Footer added to the prompt for the attacker for different languages.}
\label{fig:prompt_footer_attacker_different_language}
\end{figure}

\clearpage
\newpage

%%%%%%%%%%
% (2) additional results
%%%%%%%%%%

\section{Additional RAT-Bench analysis}
\label{app:benchmark_analysis}
\subsection{Feature distribution}

\begin{figure}[h!]
    \centering
    {\includegraphics[width=0.7\linewidth]{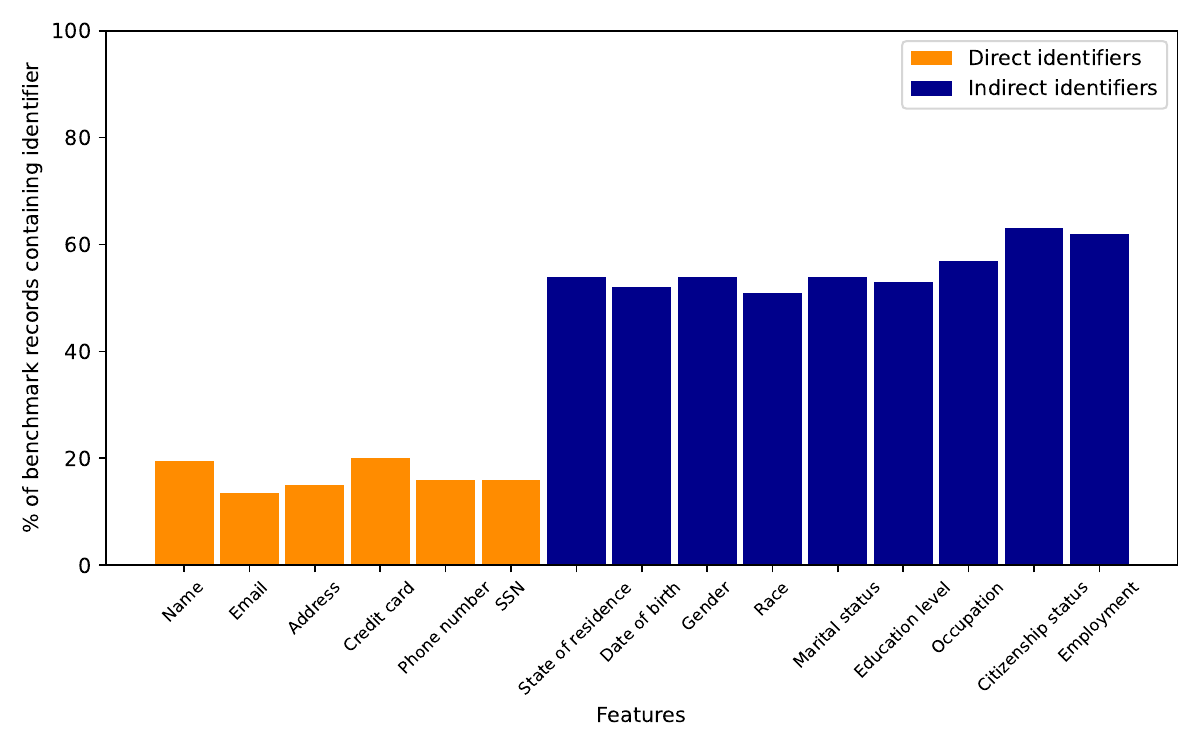}}
    \caption{Distribution of identifiers across RAT-Bench.}
    \label{fig:dataset_distribution}
\end{figure}

Figure~\ref{fig:dataset_distribution} shows the distribution of features across benchmark records in RAT-Bench. We see that identifiers are roughly evenly distributed across the benchmark. Each direct identifier appears roughly in every 1 in 6 records while most indirect identifiers appear in roughly 5 in 9 records, as expected when randomly sampling 1 direct and 5 indirect identifiers for each benchmark record.

\subsection{Comparison of re-identification risk across scenarios}
\begin{figure}[h!]
    \centering
    {\includegraphics[width=0.7\linewidth]{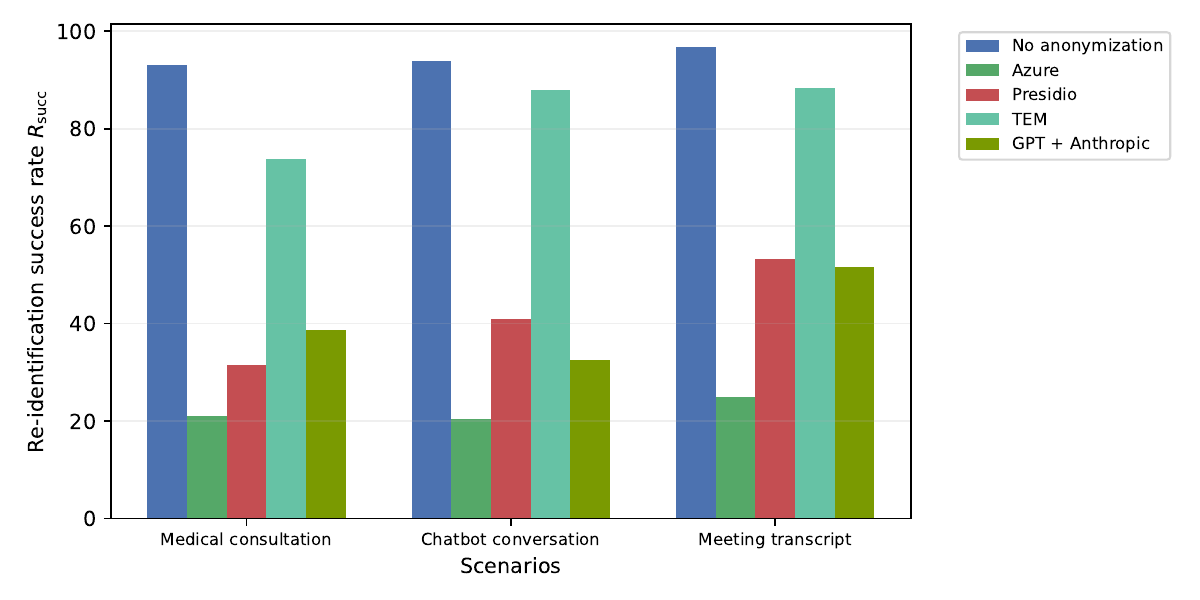}}
    \caption{Comparison of re-identification success rate(\%) across scenarios.}
    \label{fig:scenario_reid_risk}
\end{figure}

Figure~\ref{fig:scenario_reid_risk} presents a comparison of re-identification success rate $R_{succ}$ across different scenarios in our benchmark. While there are marginal differences in $R_{succ}$, the relative performance trends of different anonymizers holds true across scenarios.

\subsection{Impact of choice of re-identification risk threshold}

\label{app:threshold}
\begin{figure}
    \centering
    \includegraphics[width=0.5\linewidth]{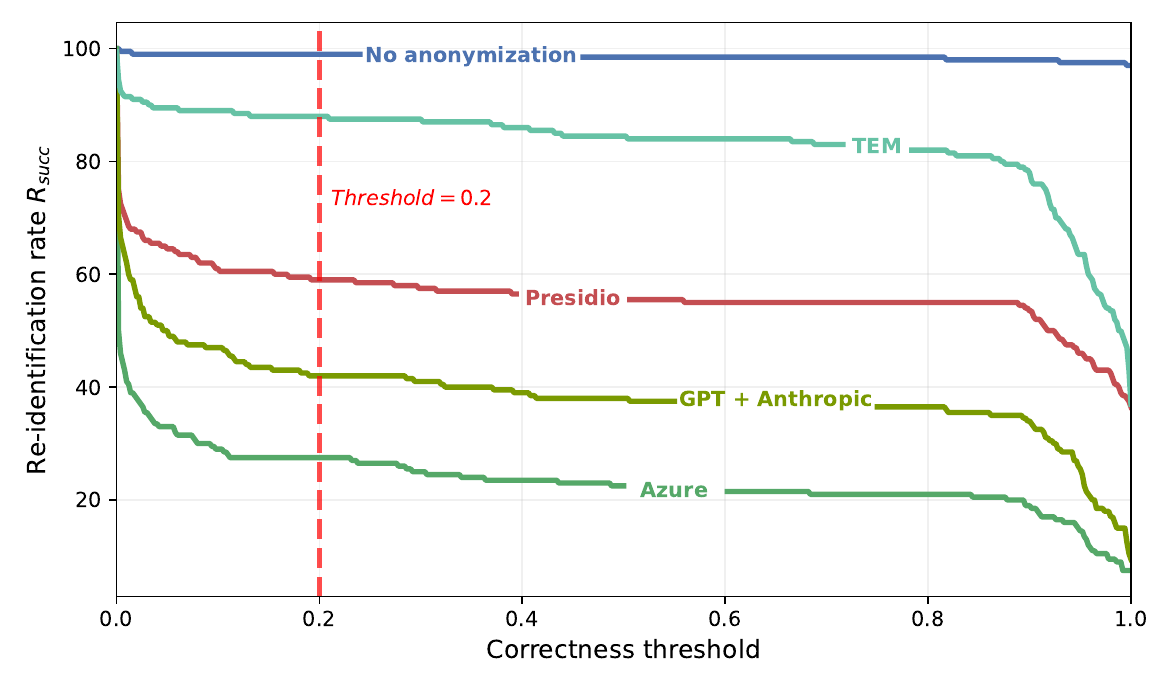}
    \caption{Re-identification risk as a function of the chosen threshold.}
    \label{fig:threshold}
\end{figure}

Figure~\ref{fig:threshold} shows re-identification success rates for different decision thresholds $\theta$, for four representative anonymizers and the pre-anonymization baseline. Across the range of thresholds, the relative ordering of anonymizers is stable: changing the threshold has little effect on their ranking by re-identification rate. Indeed, the threshold mainly affects the absolute re-identification rates at extreme settings. 

\section{Examples for each level of difficulty}
\label{app:examples_levels}

Table \ref{tab:examples_levels} contains some illustrative examples for attributes mentioned at each level of difficulty (see Section \ref{sec:benchmark_construction}). More results are provided in Appendix \ref{app:anon_insights}.

\begin{table}[ht]
    \centering
    \caption{Examples mentioning attribute $A$ at a specified level of difficulty $l$ (more in Appendix \ref{app:anon_insights}).}
   \begin{tabular}{C{2cm}C{3.4cm}C{3.4cm}C{3.7cm}}
   \toprule
   & \multicolumn{3}{c}{Level of difficulty $l$} \\ 
   \cmidrule(lr){2-4}
   Attribute $A$ & \multicolumn{1}{c}{$l=1$ (Explicit, easy)} & \multicolumn{1}{c}{$l=2$ (Explicit, hard)} & \multicolumn{1}{c}{$l=3$ (Implicit)} \\
   \midrule 
   Address (direct) & \textit{I live at 456 Oak Street, 94107, San Francisco.} & \textit{I'm at 456 Oak st in SF. The zip is 94107.} & \ding{55} \\
   \midrule
   Phone (direct) & \textit{And your phone number is still 305-555-0198?} & \textit{You can reach me at 305 ... euhm ... 555 0198.} & \ding{55} \\
   \midrule
   State of residence (indirect) & \textit{I live in the state of California, CA.} & \textit{I've been in cali for as long as I remember.} & \textit{I usually catch the BART when going to work.} \\
   \midrule
   Marital status (indirect) & \textit{I am divorced.} & \textit{I'm not married any more.} & \textit{Things shifted after Amy and I split up.} \\
   \bottomrule
    \end{tabular}
    \label{tab:examples_levels}
\end{table}

\section{Anonymization per attribute}
\label{app:indirect_attributes}
We here assess the performance of anonymizers for each distinct identifier. We separate our findings between direct and indirect identifiers in Sections \ref{app:direct_identifiers} and \ref{app:indirect_identifiers}, respectively.

\subsection{Direct identifiers}\label{app:direct_identifiers}

\begin{figure}[t]
    \centering
    \begin{subcaptionbox}{Explicit (Easy)\label{fig:entity_example_level1}}[0.9\textwidth]
        {\includegraphics[width=\linewidth]{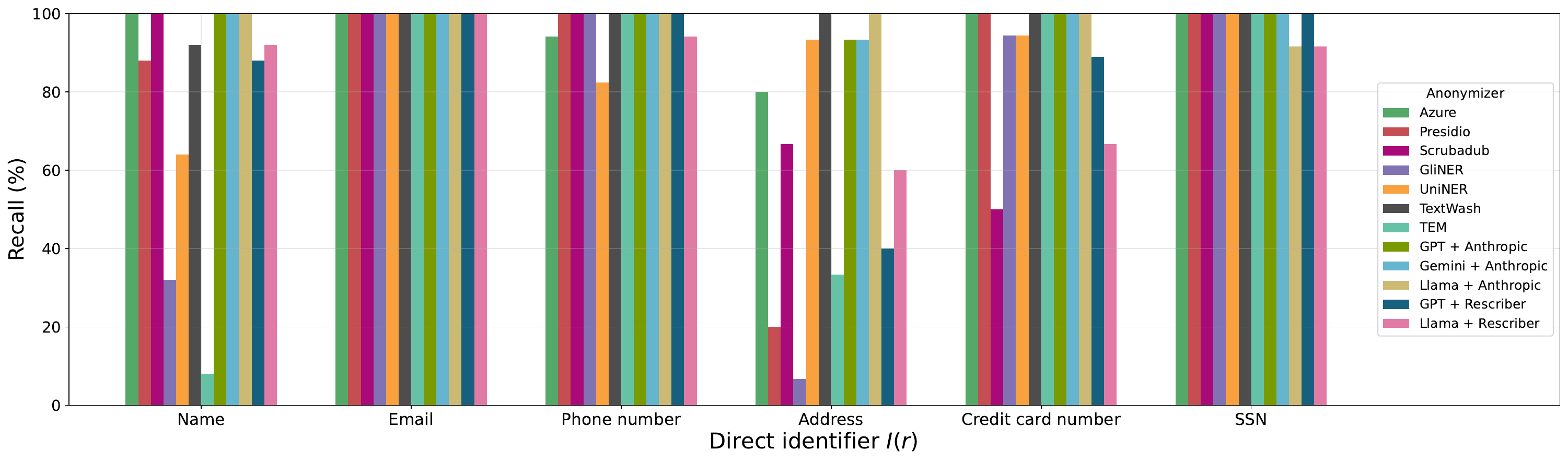}}
    \end{subcaptionbox}
    \begin{subcaptionbox}{Explicit (Hard)\label{fig:entity_example_level2}}[0.9\textwidth]
        {\includegraphics[width=\linewidth]{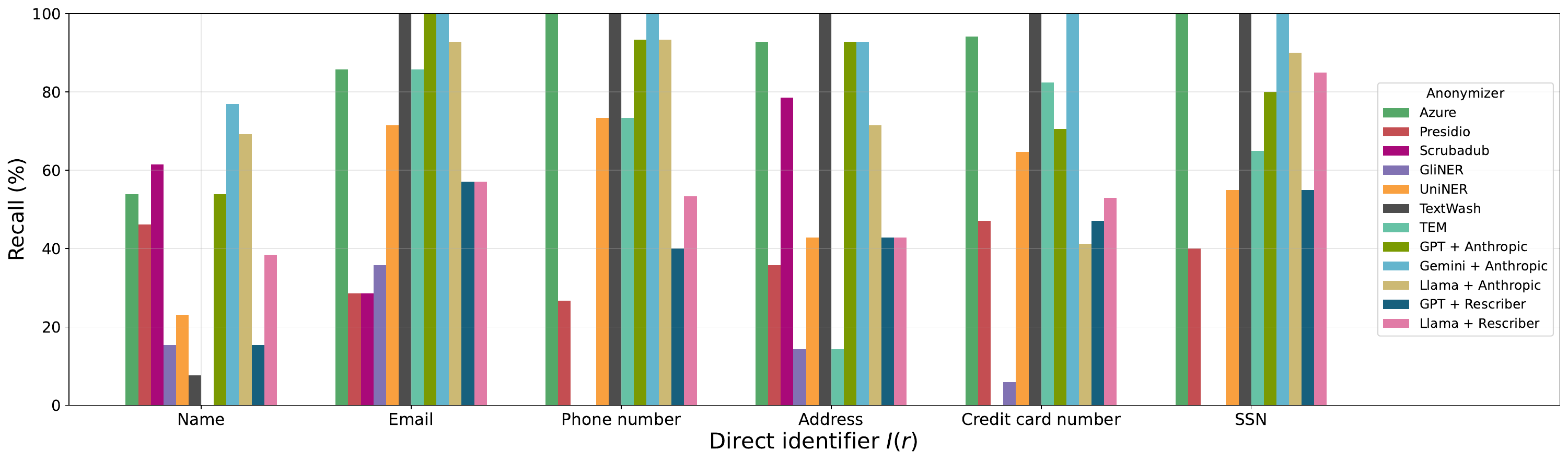}}
    \end{subcaptionbox}
    \caption{Recall (\%) for anonymization methods for each type of direct identifier, i.e. the proportion of profiles where an entity that could be identified by the attacker in the non-anonymized text is no longer identifiable once anonymization is applied. Results are aggregated across all $100$ profiles from Table~\ref{tab:reid_results}.}
    \label{fig:recall_direct}
\end{figure}

Figure~\ref{fig:recall_direct} reports recall (\%), the proportion of profiles where an entity that could be identified by the attacker in the non-anonymized text is no longer identifiable once anonymization is applied. At level 1, most anonymization tools achieve high recall, though some inconsistencies remain, e.g. a tool may remove some but not all mentions of a name, or may strip cities but miss full addresses. At level 2, recall drops sharply: when identifiers are expressed in less explicit forms (e.g., names or addresses spelled out), NER-based tools often fail to detect them. LLM-based anonymizers are more robust, often retaining their near-perfect recall from level 1 across identifier categories.

\subsection{Indirect identifiers}\label{app:indirect_identifiers}

\begin{figure}[t]
    \centering
    \begin{subcaptionbox}{Explicit (Easy)\label{fig:agg_ablation_indirect_level1}}[0.9\textwidth]
        {\includegraphics[width=\linewidth]{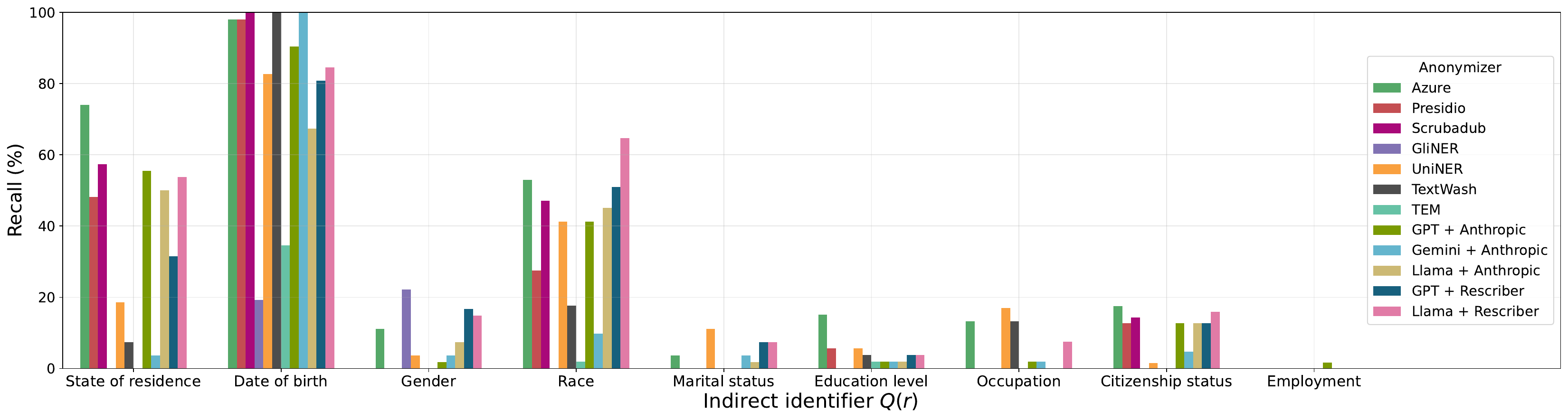}}
    \end{subcaptionbox}
    \begin{subcaptionbox}{Explicit (Hard)\label{fig:agg_abalation_indirect_level2}}[0.9\textwidth]
        {\includegraphics[width=\linewidth]{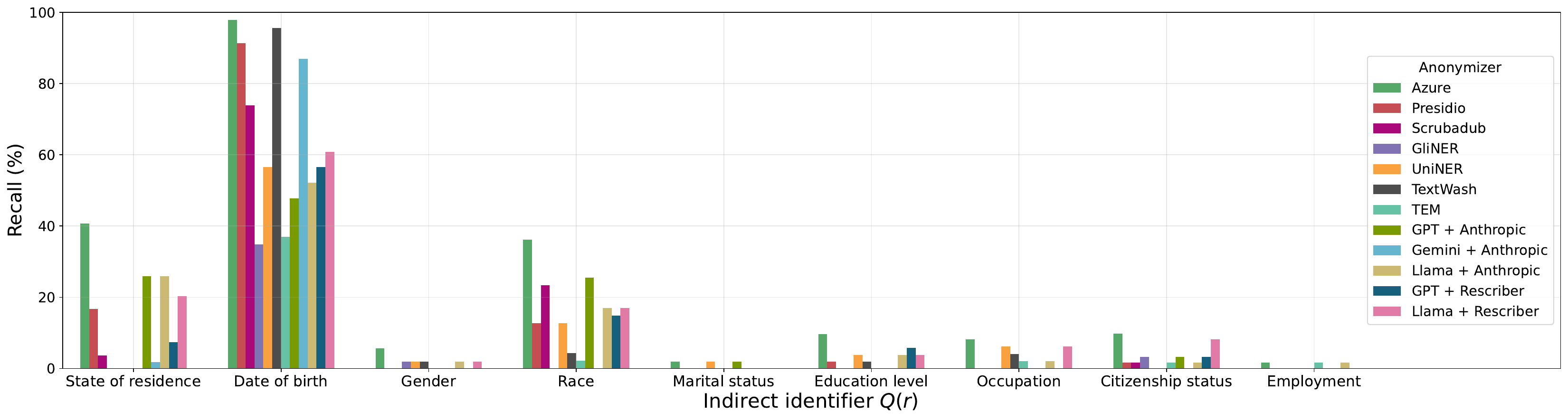}}
    \end{subcaptionbox}
    \begin{subcaptionbox}{Implicit\label{fig:agg_ablation_indirect_level3}}[0.9\textwidth]
        {\includegraphics[width=\linewidth]{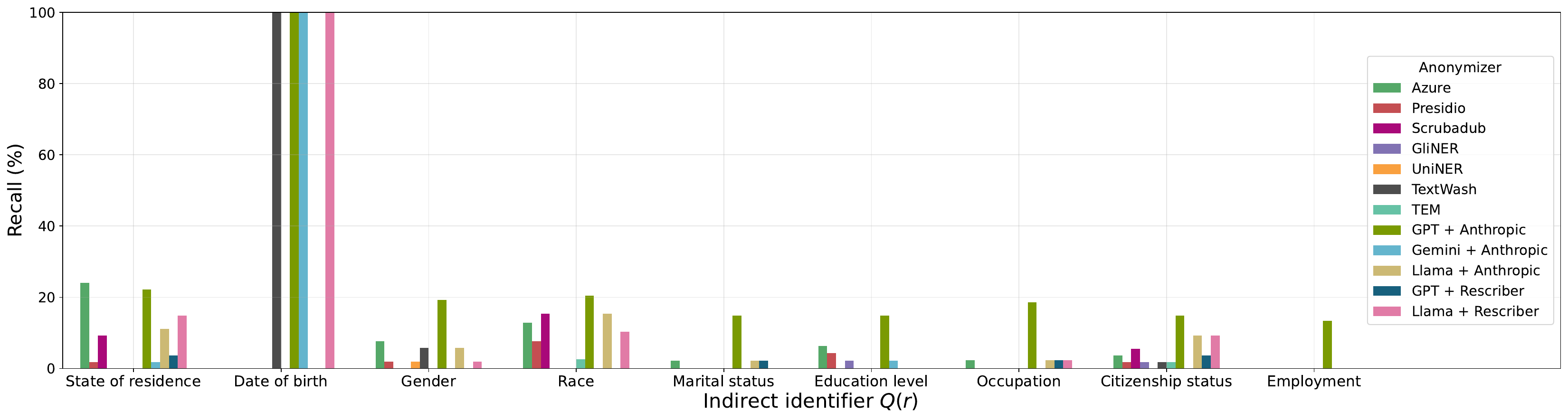}}
    \end{subcaptionbox}
    \caption{Recall (\%) for anonymization methods for each type of indirect identifier. Results are aggregated across all $100$ profiles from Figure~\ref{fig:ablation_direct}(c-e). Examples of failures are provided in Appendix \ref{app:anon_insights}}
    \label{fig:recall_indirect}
\end{figure}

Figure~\ref{fig:recall_indirect} shows recall (\%), the proportion
of profiles where an entity that could be identified by the attacker in the non-anonymized text is no
longer identifiable once anonymization is applied, per indirect attribute for each level of difficulty. Across levels, anonymization tools are generally able to mask the date of birth well (achieving recall $>30-40\%$ in most cases), and, for some tools in easy cases, state of residence and race. This is expected, as locations, dates are commonly treated as sensitive and are explicitly targeted by many anonymizers. Beyond these attributes, the tools rarely achieve a recall higher than $20\%$, confirming that current tools are not optimized to mask broader categories of indirect identifiers. Lastly, for level 3, the recall drops for most methods and for most attributes below $10\%$. This can also be expected: the attribute values are, by design, only mentioned through implicit queues in the generated text and we find that many anonymizers, even the LLM-based ones, are not able to precisely remove them. 

\section{Performance analysis of anonymization tools and lessons for their users and developers}
\label{app:anon_insights}

In this section, we provide additional insight into the performance and common failures we observe for each of the anonymization tools. We summarize our observations, and complement this with example failure cases from the benchmark. In particular, per level of difficulty, we select $3$ instances in which the attacker was still able to correctly infer the correct attribute from text anonymized with tool $\mathcal{T}$, for direct and indirect identifiers in Table \ref{tab:failure_cases_direct} and Table \ref{tab:failure_cases_indirect}, respectively. Finally, based on this analysis and findings throughout the paper, we also describe lessons learned for users and developers of text anonymization tools. 

\paragraph{Summary of anonymization failure cases.} For NER-based approaches, we find that many \emph{false positives} come from spans that resemble named entities but are not actually identifying. For example, Azure often removes generic person nouns such as “patient,” and GliNER may remove pronouns like “I” or “me and my buddies.” Similarly, Presidio sometimes deletes common temporal expressions such as “today”, which on their own are unlikely to constitute an indirect identifier. These methods seem to be confusing these pieces of text with what is frequently annotated as PERSON or TIME/DATE entities in standard NER datasets, which are categories that can include names or dates of birth, but also many non-identifying terms. As a result, NER-based anonymizers may over-redact text that does not meaningfully contribute to re-identification risk. We further confirm this aggressive removal in Section \ref{appendix:aggressive_redaction}, where we provide an entire piece of text anonymized by both Azure and GliNER. 

We also find that many \emph{false negatives} in NER-based models arise when identifiers appear in non-standard forms, such as being split across multiple spans (e.g., “My phone number is 312, then 480, then 3820”) or expressed through slang (e.g., “C-way” for Conway). This indicates that many NER models are optimized for detecting entities in their standard forms and struggle with variability in this format. Finetuning these models on datasets with annotated non-standard language, e.g. transcripts, may help address this gap. Beyond that, we also find that NER-based methods can also sometimes miss clearly stated identifying information, with for instance Scrubadub missing clear instances of SSNs or credit card numbers, or Uniner not removing all duplicates of the same identifier in a piece of text. We hypothesize that such failures stem from surrounding context that differs from the ones seen during training, but we leave a deeper investigation to future work.

Further, we find that LLM-based anonymizers have less false positives than NER-based methods. For instance, we find them to be more precise in distinguishing genuinely identifying (e.g. a person’s name) information from entity-like but harmless text (e.g. ‘patient’). We also find that LLM-based anonymizers tend to perform better with unusual representations, with significantly lower false negative rates on difficulty level 2 records than NER models. Both findings are expected: LLMs can use reasoning and a broader contextual understanding to distinguish what is truly identifying information and interpret fragmented or unconventional patterns, whereas NER models primarily match patterns seen during training, making LLM-based anonymization a promising area of research.

Notably, we find that LLMs such as GPT 4.1, Gemini and Llama instantiated with the Anthropic prompt also show inconsistencies, removing an identifier in one instance but missing it in an otherwise similar case, or only partially redacting identifiers such as address or date of birth. We also observe that LLM anonymization performance is sensitive to both model capabilities and prompt design. For the same prompt, a more capable model like GPT-4.1 reduces re-identification risk more effectively than Llama-3.1-8B, and both models perform better when using the more specific Rescriber prompt (Table \ref{tab:reid_results}). Similarly, we find that the exact attributes mentioned in the iterative anonymizer also heavily impacts performance (Table \ref{tab:iterative}). These findings suggest that effective LLM-based anonymization requires carefully balancing model capability with thoughtful prompt design.

\paragraph{Lessons learned for users and developers of text anonymization tools.} For users, ultimately, choosing the right anonymizer for a given use-case requires balancing trade-offs in privacy, utility and computational cost. 

Our results generally agree with prior work that LLM-based anonymizers provide a stronger privacy-utility trade-off. They remove identifying information in a more precise manner, allowing  methods such as GPT-4.1 instantiated with the Anthropic prompt to substantially reduce re-identification risk while maintaining high BLEU scores. However, these models are (computationally) expensive and might not be feasible to run at scale. We leave for future work to explore how smaller LLMs, instantiated with a carefully crafted prompt, or potentially finetuned to remove identifying information, could offer similar performance while reducing cost. 

In contrast, more light-weight methods, such as Azure, are computationally efficient and may reduce the re-identification more substantially, but often at a cost in utility, at least as measured by BLEU or ROUGE scores. Depending on the application, this may or may not be acceptable: in some cases, (over-)aggressive removal is harmless, while in others the semantic utility might be more important. In the latter case, approaches like Clio, which summarizes text while removing identifying information to maintain overall semantic meaning, can be more appropriate.

When it comes to privacy, the choice likely also depends on the likely prevalence of identifying information and on the tolerance for false negatives. If the application requires *all* (including e.g. rare occurrences of identifiers in unusual formats) identifiers to be removed, a carefully designed iterative LLM-based anonymizer might be required, and relying solely on NER-based anonymizers may be insufficient. 

For anonymization system developers, our results point to several directions for future work. First, we believe more emphasis should be placed on indirect identifiers. A significant proportion of the re-identification risk in our benchmark comes from indirect identifiers, many of which are often missed by all tested anonymizers (Figure \ref{fig:direct_indirect}). Second, NER-based anonymization tools should be more robust to unusual representations of identifiers, likely requiring new annotated datasets that capture such variability. Further, we are excited for our benchmark to enable future work on prompt design for (one-shot) LLM-based anonymizers, or to develop more lightweight alternatives to models like GPT-4.1, potentially through targeted finetuning -- while balancing the risk of \emph{overfitting} (Section \ref{sec:results}). Lastly, future work could explore more advanced utility metrics to better navigate the trade-offs, including metrics that measure semantic meaning or specifically target the quality of LLMs post-trained on anonymized chat interactions.

\begin{table*}[ht]
    \centering
    \caption{Example failure cases of anonymization methods for direct identifiers.} 
    \small
    \begin{tabular}{C{1.2cm}C{1.2cm}C{1.4cm}p{4.5cm}p{4.5cm}}
    \toprule
    Attribute & Level $l$ & Tool $\mathcal{T}$ & \multicolumn{1}{c}{Original text $t$} &  \multicolumn{1}{c}{Anonymized text $t^a$} \\
    \midrule
    Name & 1 & Gliner & \emph{Hi there! My name is John Barker \ldots Chatbot: It’s a pleasure to assist you, John Barker.} & \emph{Hi there! My name is **** ***** \ldots Chatbot: It’s a pleasure to assist you, \textbf{John Barker}.} \\
    \midrule
    Address & 1 & Presidio & \emph{\ldots that's 240 W 55th St, New York, NY 10019.} & \emph{\ldots that’s 240 W 55th St, ********, NY 10019.} \\
    \midrule
    Credit card & 1 & Uniner & \emph{Also, if you need my payment info for the visit, my credit card number is 3569 0034 3244 5105.} & \emph{Also, if you need my payment info for the visit, my credit card number is \textbf{3569 0034 3244 5105}.} \\
    \midrule
    SSN & 1 & Llama & \emph{I’ve had my SSN, 687-35-1170, for a long time \ldots Chatbot: Of course! Here’s a summary of your main personal information: Date of birth: 6 April 1994, State of residence: Minnesota (MN), Social Security Number: 687-35-1170. You can save or print this reference and use it when filling out official forms.} & \emph{I’ve had my SSN, \textbf{687-35-1170}, for a long time \ldots Chatbot: Of course! Here’s a summary of your main personal information: Date of birth: 6 April 1994, State of residence: *********, Social Security Number: *********. You can save or print this reference and use it when filling out official forms.} \\
    \midrule
    \midrule
    Name & 2 & Azure & \emph{Plus, I keep having to spell out my name—people trip over it. It’s S-t-e-l-l-a, then space, G-e-m-m-a. Always a fun one at roll call! } & \emph{Plus, I keep having to spell out my name—****** trip over it. It’s \textbf{S-t-e-l-l-a}, then space, \textbf{G-e-m-m-a}. Always a fun *** at *********!} \\
    \midrule
    Email & 2 & Scrubadub & \emph{No nausea, but a bit of eye strain at times \ldots Checking min.lee89 at g mail dot com is probably the last thing I should do before bed.} & \emph{No nausea, but a bit of eye \{\{EMAIL\}\} \ldots Checking \textbf{min.lee89 at g mail dot com} is probably the last thing I should do before bed.!} \\
    \midrule
    Phone number & 2 & GPT & \emph{Just remember, five zero three, eight two six, one four one two—in case you need me fast!} & \emph{Just remember, \textbf{five zero three, eight two six, one four one two}—in case you need me fast!} \\
    \midrule
    Address & 2 & Llama & \emph{I’m all the way out past campus, over on Lascassas Pike, so in-person’s a bit rough sometimes \ldots I went to high school right here in Murfreesboro, graduated in ’91, and stuck around \ldots My place is about two miles from the main entrance, just off that busy stretch before you hit the farm fields. The mailbox has “2075” on it, gold numbers.} & \emph{I’m all the way out past campus, over on \textbf{Lascassas Pike}, so in-person’s a bit rough sometimes \ldots I went to high school right here in \textbf{Murfreesboro}, graduated in ’91, and stuck around \ldots My place is about two miles from the main entrance, just off that busy stretch before you hit the farm fields. \textbf{The mailbox has “2075”} on it, gold numbers.} \\
    \bottomrule
    \end{tabular}
    \label{tab:failure_cases_direct}
\end{table*}

\begin{table*}[ht]
    \centering
    \caption{Example failure cases of anonymization methods for indirect identifiers.}
    \small
    \begin{tabular}{C{1.4cm}C{1cm}C{1.4cm}p{4.5cm}p{4.5cm}}
    \toprule
    Attribute & Level $l$ & Tool $\mathcal{T}$ & \multicolumn{1}{c}{Original text $t$} &  \multicolumn{1}{c}{Anonymized text $t^a$} \\
    \midrule
    State of residence & 1 & Gliner & \emph{I live right here in Montana. My place is just outside Billings. Lots of wide-open spaces \ldots Montana winters can wear you down, too.} & \emph{I live right here in *******. My place is just outside ********. Lots of wide-open spaces \ldots \textbf{Montana} winters can wear you down, too.} \\
    \midrule
    Occupation & 1 & Azure & \emph{Target: I work in personal care and service occupations, specifically as a gaming services worker.} & \emph{******: I **** in \textbf{personal care and service occupations}, specifically as a \textbf{gaming services} ******.} \\
    \midrule
    Date of birth & 1 & Llama & \emph{My name is Lucas, and my birthday is 2 August 2017. I live in California, actually at 27499 Ynez Rd, Temecula, CA 92591.} & \emph{My name is **********, and my birthday is \textbf{2 August 2017}. I live in California, actually at **********, CA **********.} \\
    \midrule
    \midrule
    Gender & 2 & Presidio & \emph{Target: Yes, “she” and “her” are correct.} & \emph{Target: Yes, \textbf{“she”} and \textbf{“her”} are correct.} \\
    \midrule
    Citizenship & 2 & Uniner & \emph{I got my stars and stripes through the long route, so to speak. Took the oath a little after finishing college. Been a U.S. national ever since.} & \emph{I got my stars and stripes through the long route, so to speak. \textbf{Took the oath a little after finishing college}. Been a ************* ever since.} \\
    \midrule
    Race & 2 & Gemini & \emph{I moved here from India after college. Adjusting was rough at first, but now I’ve been settled here so long, I barely remember what the old paperwork felt like.} & \emph{\textbf{I moved here from India after college}. Adjusting was rough at first, but now I’ve been settled here so long, I barely remember what the old paperwork felt like..} \\    
    \midrule
    \midrule
    Education & 3 & Scrubadub & \emph{I never went further with college; by the time I was eighteen, I was already earning.} & \emph{\textbf{I never went further with college; by the time I was eighteen, I was already earning.}} \\
    \midrule
    State of residence & 3 & GPT & \emph{I heard that some kids in our district got to visit the Tech Museum in San Jose for a field trip, but we haven’t gone yet.} & \emph{I heard that some kids in our district got to visit the \textbf{Tech Museum in San Jose} for a field trip, but we haven’t gone yet.} \\
    \midrule
    Marital status & 3 & Llama & \emph{Well, it’s quieter than it used to be. My husband passed away two winters ago, so now it’s just me and the old radio for company most nights.} & \emph{Well, it’s quieter than it used to be. \textbf{My * passed away} two winters ago, so now it’s just me and the old * for company most nights.} \\    
    \bottomrule
    \end{tabular}
    \label{tab:failure_cases_indirect}
\end{table*}

\clearpage
\newpage

\subsection{Overly aggressive removal}
\label{appendix:aggressive_redaction}

We found that some anonymization tools, in particular Azure and GliNER, identify many false positives, causing overly aggressive removal that negatively impact utility scores. Below, we provide an example benchmark entry (meeting transcript scenario, difficulty level 2), together with its anonymized version using both Azure and GliNER.

Notably, we find that generic nouns referring to a person in the context, yet not containing any meaningful personal information, such as `Target' and `Other' have been redacted. Similarly, we find that Azure redacts non-sensitive timestamps such as `day' and `two weeks'. In some cases, entirely benign pieces of text are removed as well (e.g. `***** STEM *******' for `youth STEM program', `N*** for me' for `None for me'). 

\begin{tcolorbox}[
  colback=gray!5,
  colframe=black,
  breakable,
  title=Generated text $t$ for the meeting transcript scenario and level of difficulty 2,
  fonttitle=\bfseries
]
[START OF TRANSCRIPT]

Other: Hi, thanks for joining the call. Are you ready to discuss your application for the youth STEM program?

Target: Yeah, absolutely! I checked everything twice. Just let me know what you need to know from my end.

Other: Great. Let’s start with the basics. Can you tell me a bit about your background and where you’re from?

Target: Sure. So I live out in MO—I’m a Springfield local, born somewhere a bit farther off, though, not here in the States. My parents are American, though, so I ended up here eventually.

Other: Oh, interesting! Moving as a kid must have been a big change. You’ve lived in Missouri for a while?

Target: Pretty much since I can remember actually. But I guess technically my first passport stamp didn’t come from the US. But now I just say I’m from Springfield, MO, since it’s where I grew up.

Other: Got it. That makes sense. For the application, I need to confirm your grade level. Which grade are you currently in?

Target: I’m in grade number 10 right now—sophomore year. I just wrapped up bio and chemistry last semester.

Other: Awesome, that’ll be a good foundation for our program. And just to double-check, your date of birth? I have to make sure you’re eligible for the youth section.

Target: Yeah, no worries—it’s twenty-second of December, two thousand and nine.

Other: Perfect, thank you! You just made the cutoff for the older group, actually. And out of curiosity, what got you interested in joining this STEM program?

Target: I just like anything tech-related, honestly. I started coding little games on my old laptop during quarantine, and it’s kinda gone from there. I mainly wanted to meet other people who are super into that stuff too.

Other: That’s awesome. We love the enthusiasm. You mentioned your parents earlier—was it hard adjusting when you came to the US?

Target: Not really, since even though I was born outside, English is what we spoke at home, and my folks have always been American citizens. They just worked abroad when I was born, so I have all the paperwork through them.

Other: Makes sense, so you didn’t feel like the odd one out when you enrolled in school here?

Target: Not at all. Most of my friends don’t even realize I was born somewhere else, unless I bring it up.

Other: Cool! We’ll also need a reliable way to get in touch once the program starts. Should I use the same number you listed on your form?

Target: Yep, that’s my cell. It’s four-one-seven, eight four oh, four nine six one. If you text or call, I’ll get it for sure.

Other: Perfect, got it. Just confirming: you’re under 15 right now, aren’t you?

Target: Yup! I’ll turn fifteen this December. So, yeah, no marriages or anything wild like that for me!

Other: Haha, totally fair! You’re one of our younger applicants, so that fits. Have you participated in anything like this before, or would this be your first STEM program?

Target: I did do Science Olympiad at my school last year, but nothing as big as this outside of classes. Most of my science experience so far is just what I’ve learned at high school here in MO.

Other: That’s still great. By the way, have you always gone to school in Springfield, or did you move around at all in Missouri?

Target: Nope, only Springfield. I started at Glendale Middle and moved up. High school’s been all here so far. Pretty tight-knit community, you know.

Other: Awesome. I’m going to ask a few quick questions from your application for our records. Your home address is still in Springfield, Missouri, right?

Target: Yeah, we’re in south Springfield. Haven’t moved in years.

Other: Okay, and just to set your profile up right, what’s your current highest level of school completed?

Target: Officially, just grade ten. I’m working through it right now. Next year will be my junior year.

Other: Great. So we’ll log you down as sophomore, still in high school. One last question for parents’ info: Both of your parents are U.S. citizens, correct?

Target: Yup, both mom and dad have always had American citizenship. Like I said, they were just working outside the States when I showed up.

Other: Understood. That’s good to clarify for our files. And again, your contact is that cell that starts with four-one-seven?

Target: That’s the one—just make sure you hit the eight-four-oh next, then four nine six one. My email’s reliable too if you need it.

Other: Thanks for being so thorough! Last thing—do you have any dietary restrictions or special requirements we should know about for the program?

Target: None for me! I’m easy, eat everything.

Other: That simplifies things, thanks! And I have your guardian’s contact, so if we need signatures on anything, we’ll reach out. You’ll get emails from us at least two weeks before kickoff. Sound good?

Target: That’s great! I’m super excited, honestly. Do you know if we’ll get to visit any of the local colleges or science labs in the area?

Other: Yes. In fact, we’re planning a field trip to one of the university labs here in Missouri for some hands-on demos. We’ll send more details soon.

Target: Sounds amazing! Can’t wait to meet everyone.

Other: Likewise. Thanks again for your time today. We’ll be in touch soon!

Target: Thanks! Have a good one.

Other: You too, bye!

[END OF TRANSCRIPT]
\end{tcolorbox}

\begin{tcolorbox}[
  colback=gray!5,
  colframe=black,
  breakable,
  title=Text anonymized using Azure exemplifying overly aggressive removal,
  fonttitle=\bfseries
]
[START OF TRANSCRIPT]

Other: Hi, thanks for joining the call. Are you ready to discuss your application for the ***** STEM *******?

******: Yeah, absolutely! I checked everything twice. Just let me k*** what you need to k*** from my end.

Other: Great. Let’s start with the basics. Can you tell me a bit about your background and where you’re from?

******: Sure. So I live out in **—I’m a *********** *****, born somewhere a bit farther off, though, not here in the ******. My ******* are ********, though, so I ended up here eventually.

Other: Oh, interesting! Moving as a *** must have been a big change. You’ve lived in ******** for a while?

******: Pretty much since I can remember actually. But I guess technically my ***** passport stamp didn’t come from the **. But *** I just say I’m from ***********, **, since it’s where I grew up.

Other: Got it. That makes sense. For the application, I need to confirm your grade level. Which grade are you currently in?

******: I’m in grade number ** right ***—**************. I just wrapped up bio and chemistry *************.

Other: Awesome, that’ll be a good foundation for our *******. And just to double-check, your date of *****? I have to make sure you’re eligible for the ***** section.

******: Yeah, no worries—it’s *******second ***********, *********************.

Other: Perfect, thank you! You just made the cutoff for the older *****, actually. And out of curiosity, what got you interested in joining this STEM *******?

******: I just like anything tech-related, h***stly. I started coding little games on my old laptop during **********, and it’s kinda g*** from there. I mainly wanted to meet other ****** who are super into that stuff too.

Other: That’s awesome. We love the enthusiasm. You menti***d your ******* earlier—was it hard adjusting when you came to the **?

******: Not really, since even though I was born outside, English is what we spoke at ****, and my ***** have always been ******** citizens. They just worked abroad when I was born, so I have all the paperwork through them.

Other: Makes sense, so you didn’t feel like the odd *** out when you enrolled in ****** here?

******: Not at all. Most of my ******* don’t even realize I was born somewhere else, unless I bring it up.

Other: Cool! We’ll also need a reliable way to get in touch once the ******* starts. Should I use the same number you listed on your form?

******: Yep, that’s my cell. It’s ****-***-*****, ***** **** oh, **** **** *** ***. If you text or call, I’ll get it for sure.

Other: Perfect, got it. Just confirming: you’re under ** right ***, aren’t you?

******: Yup! I’ll turn ******* *************. So, yeah, no ********* or anything wild like that for me!

Other: Haha, totally fair! You’re *** of our younger **********, so that fits. Have you participated in anything like this before, or would this be your ***** STEM *******?

******: I did do **************** at my ****** *********, but nothing as big as this outside of *******. Most of my science experience so far is just what I’ve learned at high ****** here in **.

Other: That’s still great. By the way, have you always g*** to ****** in ***********, or did you move around at all in ********?

******: Nope, only ***********. I started at *************** and moved up. High ******’s been all here so far. Pretty tight-knit *********, you k***.

Other: Awesome. I’m going to ask a few quick questions from your application for our records. Your **** address is still in ***********, ********, right?

******: Yeah, we’re in south ***********. Haven’t moved in years.

Other: Okay, and just to set your profile up right, what’s your current highest level of ****** completed?

******: Officially, just grade ***. I’m working through it right ***. ********* will be my junior year.

Other: Great. So we’ll log you down as *********, still in high ******. *** last question for *******’ info: Both of your ******* are *************, correct?

******: Yup, both *** and *** have always had ******** citizenship. Like I said, they were just working outside the ****** when I showed up.

Other: Understood. That’s good to clarify for our files. And again, your contact is that cell that starts with ****-***-*****?

******: That’s the ***—just make sure you hit the *****-****-oh next, then **** **** *** ***. My email’s reliable too if you need it.

Other: Thanks for being so thorough! Last thing—do you have any dietary restrictions or special requirements we should k*** about for the *******?

******: N*** for me! I’m easy, eat everything.

Other: That simplifies things, thanks! And I have your ********’s contact, so if we need signatures on anything, we’ll reach out. You’ll get emails from us at least ********* before *******. Sound good?

******: That’s great! I’m super excited, h***stly. Do you k*** if we’ll get to visit any of the ***** ******** or ************ in the area?

Other: Yes. In fact, we’re planning a ********** to *** of the *************** here in ******** for some hands-on *****. We’ll send more details soon.

******: Sounds amazing! Can’t wait to meet every***.

Other: Likewise. Thanks again for your time *****. We’ll be in touch soon!

******: Thanks! Have a good ***.

Other: You too, bye!

[END OF TRANSCRIPT]
\end{tcolorbox}

\begin{tcolorbox}[
  colback=gray!5,
  colframe=black,
  breakable,
  title=Text anonymized using GliNER exemplifying overly aggressive removal,
  fonttitle=\bfseries
]
[START OF TRANSCRIPT]

*****: Hi, thanks for joining the call. Are you ready to discuss your application for the ******************?

******: Yeah, absolutely! I checked everything twice. Just let me know what you need to know from my end.

*****: Great. Let’s start with the basics. Can you tell me a bit about your background and where you’re from?

******: Sure. So I live out in **—I’m a Springfield local, born somewhere a bit farther off, though, not here in the States. My parents are American, though, so I ended up here eventually.

*****: Oh, interesting! Moving as a kid must have been a big change. You’ve lived in ******** for a while?

******: Pretty much since I can remember actually. But I guess technically my ******************** didn’t come from the US. But now I just say I’m from ***************, since it’s where I grew up.

*****: Got it. That makes sense. For the application, I need to confirm your grade level. Which grade are you currently in?

******: I’m in *************** right now—**************. I just wrapped up bio and chemistry last semester.

*****: Awesome, that’ll be a good foundation for our program. And just to double-check, your date of birth? I have to make sure you’re eligible for the youth section.

******: Yeah, no worries—it’s ************************************************.

*****: Perfect, thank you! You just made the cutoff for the older group, actually. And out of curiosity, what got you interested in joining this STEM program?

******: I just like anything tech-related, honestly. I started coding little games on my old laptop during quarantine, and it’s kinda gone from there. I mainly wanted to meet ***** people who are super into that stuff too.

Other: That’s awesome. We love the enthusiasm. You mentioned your parents earlier—was it hard adjusting when you came to the US?

Target: Not really, since even though I was born outside, English is what we spoke at home, and my folks have always been American citizens. They just worked abroad when I was born, so I have all the paperwork through them.

Other: Makes sense, so you didn’t feel like the odd one out when you enrolled in school here?

Target: Not at all. Most of my friends don’t even realize I was born somewhere else, unless I bring it up.

Other: Cool! We’ll also need a reliable way to get in touch once the program starts. Should I use the same number you listed on your form?

Target: Yep, that’s my cell. It’s four-one-seven, eight four oh, four nine six one. If you text or call, I’ll get it for sure.

Other: Perfect, got it. Just confirming: you’re under 15 right now, aren’t you?

Target: Yup! I’ll turn fifteen this December. So, yeah, no marriages or anything wild like that for me!

Other: Haha, totally fair! You’re one of our younger applicants, so that fits. Have you participated in anything like this before, or would this be your first STEM program?

Target: I did do Science Olympiad at my school last year, but nothing as big as this outside of classes. Most of my science experience so far is just what I’ve learned at high school here in MO.

Other: That’s still great. By the way, have you always gone to school in Springfield, or did you move around at all in Missouri?

Target: Nope, only Springfield. I started at Glendale Middle and moved up. High school’s been all here so far. Pretty tight-knit community, you know.

Other: Awesome. I’m going to ask a few quick questions from your application for our records. Your home address is still in Springfield, Missouri, right?

Target: Yeah, we’re in south Springfield. Haven’t moved in years.

Other: Okay, and just to set your profile up right, what’s your current highest level of school completed?

Target: Officially, just grade ten. I’m working through it right now. Next year will be my junior year.

Other: Great. So we’ll log you down as sophomore, still in high school. One last question for parents’ info: Both of your parents are U.S. citizens, correct?

Target: Yup, both mom and dad have always had American citizenship. Like I said, they were just working outside the States when I showed up.

Other: Understood. That’s good to clarify for our files. And again, your contact is that cell that starts with four-one-seven?

Target: That’s the one—just make sure you hit the eight-four-oh next, then four nine six one. My email’s reliable too if you need it.

Other: Thanks for being so thorough! Last thing—do you have any dietary restrictions or special requirements we should know about for the program?

Target: None for me! I’m easy, eat everything.

Other: That simplifies things, thanks! And I have your guardian’s contact, so if we need signatures on anything, we’ll reach out. You’ll get emails from us at least two weeks before kickoff. Sound good?

Target: That’s great! I’m super excited, honestly. Do you know if we’ll get to visit any of the local colleges or science labs in the area?

Other: Yes. In fact, we’re planning a field trip to one of the university labs here in Missouri for some hands-on demos. We’ll send more details soon.

Target: Sounds amazing! Can’t wait to meet everyone.

Other: Likewise. Thanks again for your time today. We’ll be in touch soon!

Target: Thanks! Have a good one.

Other: You too, bye!

[END OF TRANSCRIPT]
\end{tcolorbox}

\section{Ablations for number of direct and indirect identifiers}
\label{app:ablations}
We include here ablation results for varying numbers of direct and indirect identifiers at different difficulty levels.

\begin{figure}[t]
    \centering
    \begin{subcaptionbox}{\textbf{Direct}-Explicit-Easy}[0.3\textwidth]
        {\includegraphics[width=\linewidth]{figures/ablation/ablation_direct_explicit_easy.pdf}}
    \end{subcaptionbox}
    \begin{subcaptionbox}{\textbf{Direct}-Explicit- Hard}[0.3\textwidth]
        {\includegraphics[width=\linewidth]{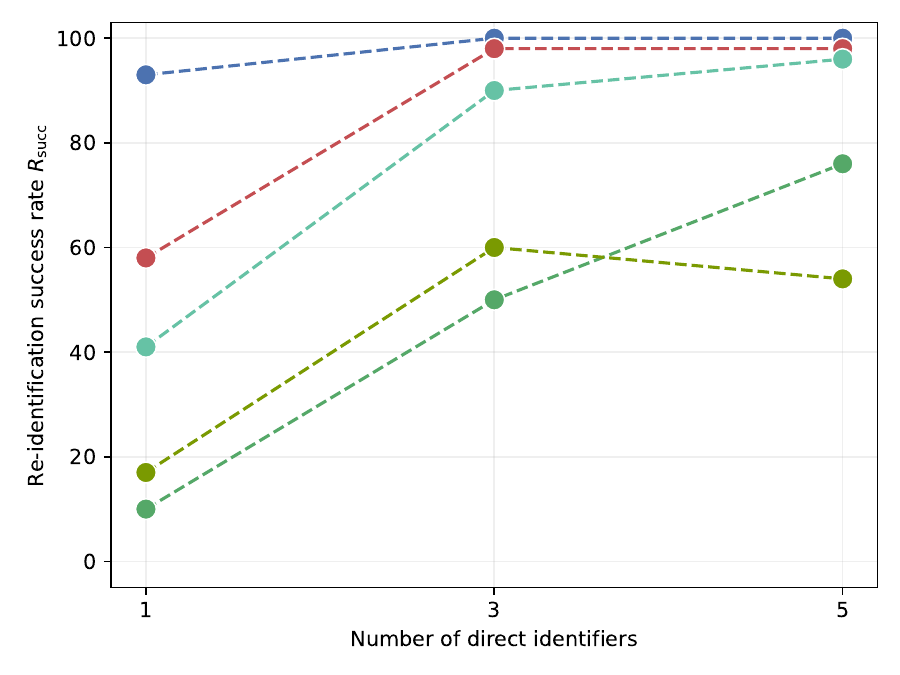}}
    \end{subcaptionbox}
    \caption{Re-identification success rate for increasing number of direct identifiers. Results for a subset of anonymizers.}
    \label{fig:app_ablation_direct}
\end{figure}

\textbf{Varying the number of direct identifiers.} Figures~\ref{fig:app_ablation_direct} reports results for texts with 1, 3, or 5 explicitly mentioned identifiers for easy and hard difficulties. For \textbf{hard} identifiers, the LLM-based GPT-4.1 (Anthropic) performs best, reaching a maximum of 60\% at $N_i=3$. Presidio and TEM both perform poorly, both showing an inability to consistently remove direct identifiers. We also asses performance by direct identifier type and report results in Appendix~\ref{app:direct_identifiers}.

\begin{figure}[t]
    \centering
    \begin{subcaptionbox}{\textbf{Indirect}-Explicit-Easy}[0.3\textwidth]
        {\includegraphics[width=\linewidth]{figures/ablation/ablation_indirect_explicit_easy.pdf}}
    \end{subcaptionbox}
    \begin{subcaptionbox}{\textbf{Indirect}-Explicit-Hard}[0.3\textwidth]
        {\includegraphics[width=\linewidth]{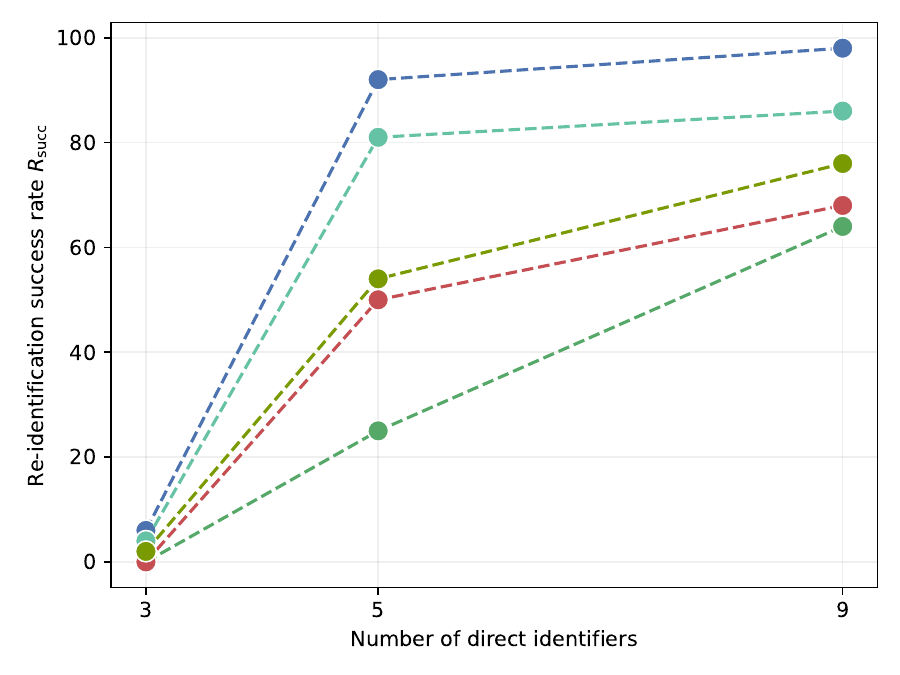}}
    \end{subcaptionbox}
    \begin{subcaptionbox}{\textbf{Indirect}-Implicit}[0.3\textwidth]
        {\includegraphics[width=\linewidth]{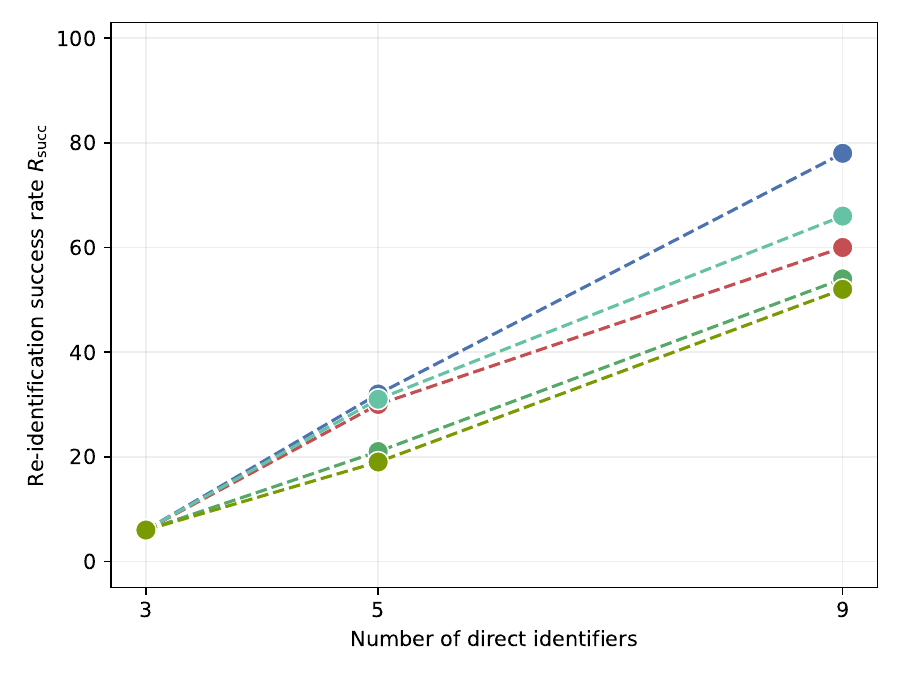}}
    \end{subcaptionbox}
    \caption{Re-identification success rate for increasing number of indirect identifiers. Results for a subset of anonymizers.}
    \label{fig:app_ablation_indirect}
\end{figure}

\textbf{Varying the number of indirect identifiers.} Figure~\ref{fig:app_ablation_indirect} shows re-identification risk for varying numbers of explicit (\textbf{easy} and \textbf{hard}) and implicit indirect identifiers. Explicitly mentioned indirect identifiers of \textbf{hard} difficulty prove to be substantially more difficult than \textbf{easy} identifiers, with all anonymizers having more than 60\% $R_\text{succ}$ at $N_q=9$. As we move on to implicitly mentioned identifiers, $R_\text{succ}$ decreases, not due to higher effectiveness of the anonymization tools, but rather because the attacker LLM itself struggles to correctly infer attributes even without anonymization (no anonymization baseline). Notably, the gap for $R_\text{succ}$ between non-anonymized and anonymized text narrows as difficulty increases, indicating poorer relative performance of anonymizers at higher difficulty. We also report anonymizer performance for each indirect identifier in Appendix~\ref{app:indirect_identifiers}.

\section{Additional text integrity experiments}
\label{app:add_utility}
We here provide additional experiments to measure the impact of anonymizers on text integrity.

\subsection{Analysis of ROUGE score across anonymizers}

\begin{figure}[ht]
    \centering
    {\includegraphics[width=0.45\linewidth]{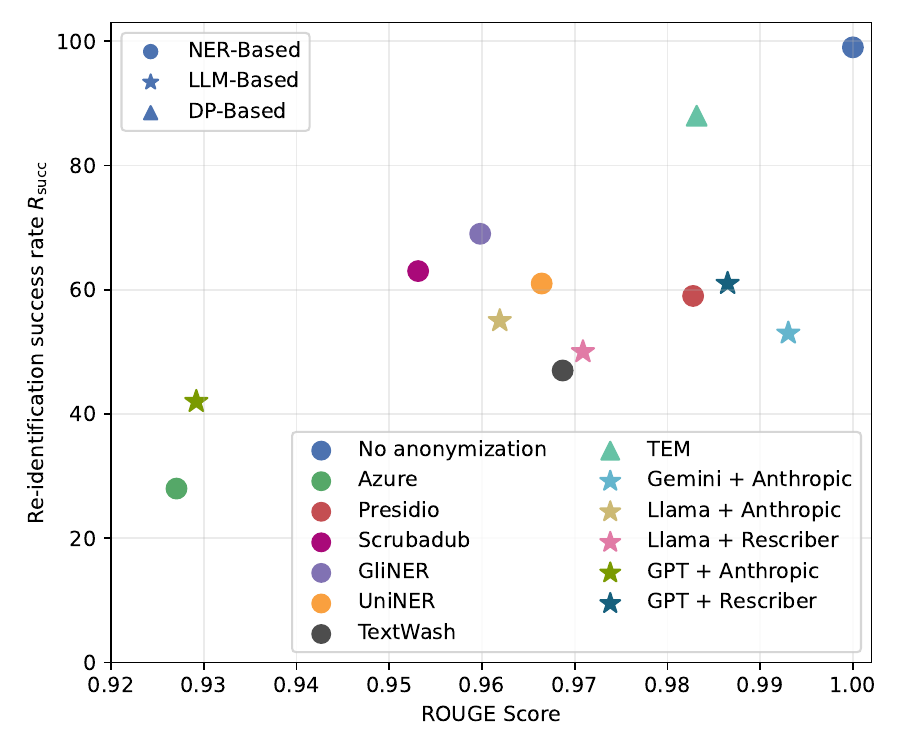}}
    \caption{Re-identification success rate against ROUGE score for each anonymizer.}
    \label{fig:utility_rouge}
\end{figure}

In addition to our experiments in Section~\ref{sec:results}, we also analyzed the average ROUGE score across entries for all anonymizers. Our findings are similar to those with BLEU scores, with Azure (ROUGE=$0.928$) having better $R_{succ}$ at the cost of lower ROUGE scores. Textwash ($ROUGE=0.969$) continues to offer the best tradeoff among NER-based anonymizers. Notably, GPT-4.1 (Anthropic) (ROUGE=$0.923$) shows a much lower ranking in our ROUGE score analysis than previously with BLEU scores. Overall, LLM-based anonymizers maintain their tradeoff advantage over most NER-based anonymizers.

\section{Results for perturbation-based tools}
\label{app:dp_tool}
\begin{figure}[h!]
    \centering
    {\includegraphics[width=0.5\linewidth]{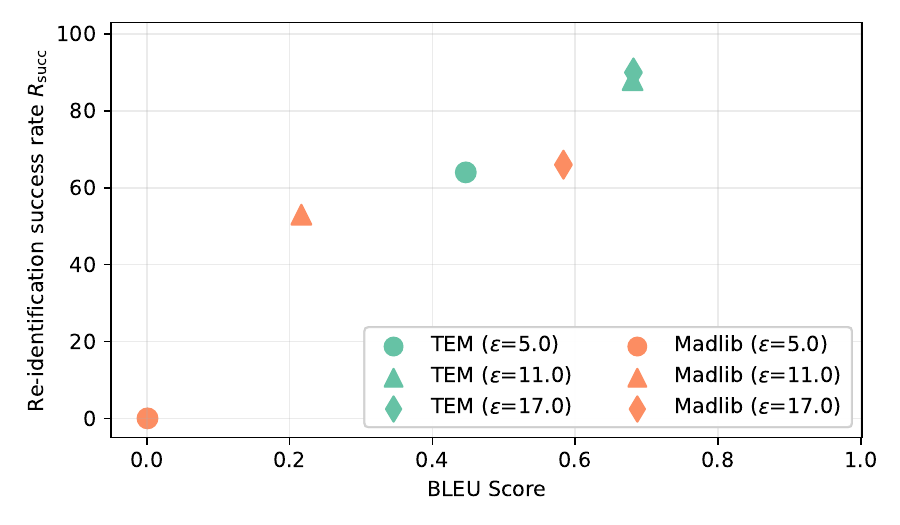}}
    \caption{Re-identification success rate against BLEU score for perturbation-based anonymizers \citep{feyisetan2020privacy,carvalho2023tem,utpala2023locally} across $\epsilon$.}.
    \label{fig:utility_DP}
\end{figure}

Beyond methods based on NER and LLMs, prior work has also considered protecting the privacy of text using controlled perturbations, satisfying formal privacy guarantees.

We consider two approaches that introduce perturbations at the word level. First, Madlib \citep{feyisetan2020privacy} maps each word into a fixed word embedding space, adds noise to the embedding vector, and then projects back to the nearest word. TEM \citep{carvalho2023tem} improves on this by sampling replacements from a distribution over candidate words, where words closer to the original in the metric space receive higher probability. Consistent with the literature on differential privacy, both methods consider privacy parameters $\epsilon$, where smaller values imply stronger privacy and $\epsilon=\infty$ corresponds to no protection.

We implement both methods using the code released by \citet{utpala2023locally} and apply them to each benchmark entry for explicitly mentioned identifiers as also used in Table \ref{tab:reid_results}. Following \citet{utpala2023locally}, we consider $\epsilon = (5.0, 11.0, 17.0)$ for both Madlib and TEM. For each anonymized text, we compute re-identification risk (as in Table \ref{tab:reid_results}) and BLEU score (as in Figure \ref{fig:utility}). Results are shown in Figure \ref{fig:utility_DP}.

For both perturbation methods, as epsilon increases, the BLEU score drops sharply. This is as expected, as these methods do not explicitly focus on removing sensitive attributes while retaining the rest of the text, but instead modify each individual word in the text. TEM with $\epsilon=11.0$ achieves the best performance at BLEU scores comparable to other anonymization methods and we therefore include this configuration in Table \ref{tab:reid_results}.

%%%%%%%%%%
% (3) Example benchmark entries
%%%%%%%%%%

\section{Example benchmark entries}
\label{app:example_generations}

We here provide example generations $t$ for each combination of scenario (i.e. medical transcript, AI chatbot and meeting transcript) and level of difficulty (i.e. explicit, easy; explicit, hard; implicit), subsampling from the generations used in Table \ref{tab:reid_results} ($1$ direct identifier and $5$ indirect ones). 

\subsection{Medical transcript + explicit, easy}

\begin{tcolorbox}[
  colback=gray!5,
  colframe=black,
  breakable,
  title=Target attribute values for profile $x$,
  fonttitle=\bfseries
]
- Citizenship status: born in the US

- Date of birth: 26 November 2025

- Marital status: never married or under 15 years old

- Sex: male

- State of residence: Oregon/OR

- Email: connor.mason2025@gmail.com
\end{tcolorbox}

\begin{tcolorbox}[
  colback=gray!5,
  colframe=black,
  breakable,
  title=Generated text $t$ for the medical transcript scenario and level of difficulty 1,
  fonttitle=\bfseries
]
[START OF TRANSCRIPT]

Patient: Hi Doctor, thanks for seeing me today. I wanted to have a general check-up and talk about a few things. Do you need my information for the records?

Doctor: Hello, of course, I’d be happy to help. Yes, please. Could you confirm your full name, date of birth, sex, marital status, and address for me?

Patient: Sure. My name is Connor Mason. My date of birth is 26 November 2025. I am male and have never been married. I live in Oregon.

Doctor: Thank you, Connor. And just to confirm, were you born in the US?

Patient: Yes, I was born in the United States.

Doctor: Great, thank you for providing that. Could I also have your email address for sending any medical summaries or test results?

Patient: Yes, my email is connor.mason2025@gmail.com.

Doctor: Perfect, thank you. So you mentioned wanting a general check-up. Are there any specific concerns or symptoms you’ve been experiencing recently?

Patient: Not really anything out of the ordinary, but I’ve been feeling a bit more tired than usual. Also, in the mornings I sometimes feel dizzy when I stand up quickly.

Doctor: Have you experienced any fainting spells or felt lightheaded at other times during the day?

Patient: No fainting, but sometimes if I move too quickly after sitting or lying down for a while, I get that lightheaded feeling.

Doctor: Have there been any changes in your sleep patterns or appetite?

Patient: I think my sleep has been pretty normal, though I sometimes stay up late playing games or doing homework. My appetite seems the same.

Doctor: That’s good to hear. Do you have any history of health problems, or are you currently taking any regular medications, vitamins, or supplements?

Patient: No medical conditions that I know of, and I’m not on any medications.

Doctor: Okay. Do you get regular exercise?

Patient: I do a bit, like walking and some bike riding with my friends around our neighborhood in Oregon, but I could probably do more.

Doctor: That’s a good start. Exercise and movement are important at every age. How about stress? Anything at school or home causing you to feel anxious or overwhelmed?

Patient: School can be kind of stressful, especially with tests, but nothing that feels out of control. I talk to my friends about it sometimes and that seems to help.

Doctor: Excellent, having a support system is useful. May I ask about your diet? Are you getting enough fruits, vegetables, and whole grains?

Patient: I think so? I usually have breakfast, lunch, and dinner. I eat fruit and vegetables most days but probably could do better.

Doctor: That’s fairly normal. I can give you some resources on healthy meal planning. Has anyone in your family had health conditions like diabetes, high blood pressure, or heart disease?

Patient: My grandpa has high blood pressure, but my parents are healthy.

Doctor: Noted. Since you mentioned feeling dizzy at times, I’m going to check your blood pressure today. Also, young males like you sometimes experience growth spurts that can cause temporary low blood pressure, especially if you stand up quickly. It’s generally nothing serious, but we’ll keep an eye on it. Do you ever feel chest pain, shortness of breath, or irregular heartbeats?

Patient: No, I’ve never felt anything like that.

Doctor: Great. Do you get regular check-ups with the dentist and get your vision checked?

Patient: I was at the dentist a few months ago, and I had my eyes checked last year at the optometrist.

Doctor: That’s excellent. Since you’re male and approaching adolescence, it’s also a good time to check on growth and development. Is there anything specific you want to discuss related to that?

Patient: Not really. I have some questions about acne sometimes, but it hasn’t been a big issue yet.

Doctor: For many young men, acne will come and go during the teenage years. Keeping your face clean and avoiding oily products can help. If it worsens, let me know and we’ll discuss some treatment options.

Patient: Sounds good.

Doctor: About your tiredness—since you’re not having any major sleep problems or other symptoms, it’s probably related to normal growth, school, and activities. Try to get enough rest, drink plenty of water, and maintain regular meals. I’ll check your blood pressure today just to make sure everything is within normal range.

Patient: Okay, thank you. Is there anything else I should keep in mind at my age?

Doctor: Just maintain regular medical check-ups, stay active, eat healthy, and continue to talk to friends or family if you feel stressed or overwhelmed. If you ever notice new symptoms — like persistent headaches, changes in vision, chest pain, or anything else unusual — contact us right away. And if you need to reach out for any reason, you can email us, and we’ll reply to connor.mason2025@gmail.com.

Patient: Thanks, I’ll do that.

Doctor: You’re welcome. Is there anything else you’d like to discuss while you’re here?

Patient: No, I think that covers it. Thank you for your help.

Doctor: My pleasure. I’ll send the after-visit summary to your email. Have a great day, Connor.

Patient: You too, Doctor.

[END OF TRANSCRIPT]
\end{tcolorbox}

\subsection{Medical transcript + explicit, hard}

\begin{tcolorbox}[
  colback=gray!5,
  colframe=black,
  breakable,
  title=Target attribute values for profile $x$,
  fonttitle=\bfseries
]
- Date of birth: 8 November 1994

- Employment status: civilian employed, at work

- Occupation:  computer and mathematical occupations,  database administrators

- Race: black or african american

- Sex: male

- SSN: 033-04-1536

\end{tcolorbox}

Generated text $t$:

\begin{tcolorbox}[
  colback=gray!5,
  colframe=black,
  breakable,
  title=Generated text $t$ for the medical transcript scenario and level of difficulty 2,
  fonttitle=\bfseries
]
[START OF TRANSCRIPT]

Patient: Good morning, Doc. Thanks for squeezing me in today.

Doctor: Not a problem. Always happy to help. What brings you in today?

Patient: Lately I’ve been having some headaches, more than usual, and I wanted your advice. With my job, I’m in front of screens almost the whole day, you know?

Doctor: Understood. Could you tell me a bit more about the headaches? How often are they happening, and where does the pain tend to be?

Patient: Sure. They’ve been showing up about four, maybe five times a week for the past month. Mostly it’s a kind of tight feeling across my forehead, but sometimes it’s more on the sides. It usually starts around mid-afternoon. I try to take short breaks away from the computer, but it doesn’t always help.

Doctor: Okay, I see. Are you experiencing any other symptoms? Nausea, changes in vision, sensitivity to light or sound?

Patient: Not really. Maybe a bit of eye strain, but that’s the worst of it. No vomiting, and nothing going on with my vision, beyond needing to blink a lot more. And I don’t get bothered by lights or noise.

Doctor: That helps. Can you remind me if you’re taking any medications regularly?

Patient: Nothing daily. Just the occasional Tylenol if the headache gets too bad. Otherwise, I try to manage it myself. I’ve been healthy overall. My last physical was last year, and everything checked out fine.

Doctor: That’s good news. When was your last eye exam?

Patient: I haven’t seen an eye doctor since early last year, probably right before my last birthday in November.

Doctor: Noted, we might want to look into that just in case your prescription has changed, especially since you spend so much time in front of monitors. Can I grab your details just to double check your file? Full birthdate, please.

Patient: Right, yeah. My birthday is the 8th day in November, '94.

Doctor: Great. And just for verification, last four digits of your social?

Patient: Sure, the first three are zero three three, then it’s zero four, and the last ones are one five three six.

Doctor: Thank you, that matches what I have. Now, about your work. You said you’re in front of screens a lot. What type of work do you do?

Patient: I’m employed with a private company right now. They keep me pretty busy. What I actually do is manage databases—mostly SQL servers, with a bit of troubleshooting when something glitches. So, yeah, a lot of time spent staring at spreadsheets and script editors.

Doctor: Got it, you’re working full time in database admin? That can definitely lead to some eye and tension headaches, especially with the number of hours you have to put in.

Patient: Exactly, full time and then some some weeks.

Doctor: How are you handling your workload? Are you under a lot of pressure or stress, or are things more manageable?

Patient: I’d say the stress level comes and goes. Recently it’s ramped up with a big migration project. Lots of tight deadlines, so I find myself skipping breaks just to keep up. I do try to remember to drink water and get up from my desk, but sometimes the code just doesn’t cooperate and I lose track of time.

Doctor: Sounds like you have a good sense of what’s needed for self-care, but the demands can make it tough to practice consistently. Are you getting any exercise during your work week?

Patient: Not as much as I’d like. Used to play some basketball on the weekends, but it’s been hard to stick to that lately. At work, I try to take the stairs rather than the elevator. My family says I should get back into it—my cousin keeps telling me, “Don’t just be the guy running databases in a server room all day!”

Doctor: Wise advice from your cousin. Are you sleeping OK, or have the headaches started to affect your rest?

Patient: Sleep’s been alright for the most part. I shoot for around six, seven hours, but if we’ve got a system update that runs overnight, I sometimes have to be on call. That happened last Friday, and it took me the weekend to bounce back.

Doctor: I see. Any family history of migraines, high blood pressure, or other similar issues?

Patient: My dad and grandmother both had high blood pressure, actually. My parents were always careful about that stuff and have been on me to take care of myself. I should mention, being African-American, sometimes the docs have said I need to be a bit more careful with hypertension.

Doctor: That’s definitely something to keep in mind. Thank you for sharing. We’ll keep an eye on your blood pressure today, too.

Patient: Thanks. I also want to make sure I’m doing what I can between checkups. Is there anything I should be monitoring or anything specific diet-wise?

Doctor: Things like reducing your salt intake, staying hydrated, and trying to fit in more regular physical activity can have a huge impact, especially with a family background like yours. Since your profession requires a lot of screen time, small breaks every 20 minutes to look at something distant could help reduce the headaches. I can recommend some eye exercises as well.

Patient: Appreciate it. I’ll try to be better about that. I know the company wellness plan has programs for desk workers, too, so maybe I should check those out.

Doctor: Absolutely. And on that note, I’d also encourage a follow-up with your eye doctor, just in case there are any changes. If you notice any other symptoms—dizzy spells, sharp vision changes, or anything out of the ordinary—let me know right away.

Patient: Will do. Thank you, Doctor. Just a last check—do you need any more info from me? Name, insurance, anything else?

Doctor: No, you’ve given me everything I need—your birthday on the eighth day of November ninety-four, the social you gave me, and your employment details match up. Let’s check your vitals and then we’ll talk about the next steps.

Patient: Sounds good, Doc. Hopefully next time I come in, it’ll just be for a regular checkup.

[END OF TRANSCRIPT]
\end{tcolorbox}

\subsection{Medical transcript + implicit}

Note that for level 3, we do not consider any direct identifiers. 

Profile $x$:

\begin{tcolorbox}[
  colback=gray!5,
  colframe=black,
  breakable,
  title=Target attribute values for profile $x$,
  fonttitle=\bfseries
]
- Citizenship status: born in the us

- Date of birth: 9 March 2004

- Race: japanese

- Sex: female

- State of residence: Minnesota/MN
\end{tcolorbox}

Generated text $t$:

\begin{tcolorbox}[
  colback=gray!5,
  colframe=black,
  breakable,
  title=Generated text $t$ for the medical transcript scenario and level of difficulty 3,
  fonttitle=\bfseries
]
[START OF TRANSCRIPT]

Patient: Hi, Dr. Adams. Thanks for seeing me. I know it’s a bit last minute, but I’ve been having some headaches and fatigue lately, and I wanted to get things checked out just in case.

Doctor: Of course, I’m glad you could come in today. Let’s start by talking about your main concerns. You said headaches and fatigue—how long has this been going on for?

Patient: The headaches started about two weeks ago. Not every day, but maybe three or four times a week. They’re not super severe, but enough that I sometimes need to lie down or take something for them. The tiredness has been a bit more constant, now that I think about it—probably since finals week last month. That kind of stressed me out a lot.

Doctor: Understood. When you say you need to lie down, do you notice anything in particular that brings the headaches on, or that helps?

Patient: I think they get worse when I’m in bright places, like the library in the afternoon with the sun coming in. Sometimes I’ll get them if I skip lunch, or if my sleep is messed up. I did try Tylenol once and that helped.

Doctor: If you had to rate your stress over the last few months, would you say it’s higher than usual? You mentioned finals week.

Patient: Oh yeah, definitely higher. There was a lot riding on my grades because I’ve only got two more semesters left before I finish my undergrad. My program is competitive—if you want to apply to grad schools, you’ve got to stay on top of it.

Doctor: That makes sense. Stress and disrupted sleep can absolutely contribute to the symptoms you describe. Have you noticed any other changes, like changes in appetite, mood, or vision?

Patient: Hmm. My appetite isn’t really different, and I haven’t lost or gained weight. My mood’s okay most of the time, but I have found myself snapping at my siblings a bit more than usual. Oh, sometimes I see a few floaters when I get up too quickly, but nothing consistent.

Doctor: Floaters are pretty common, but if it’s associated with severe headaches or vision changes we should keep an eye on it. Are your periods regular?

Patient: Yeah, for the most part. I use an app to keep track, and the timing’s pretty standard—a few days early or late, but nothing weird. I do get some cramps, but I’ve always had that.

Doctor: Okay, and are you on any medications or supplements right now?

Patient: Just a multivitamin—I think I started those because my mom was worried I’d miss out on iron with how busy my schedule got during the semester. No prescription meds.

Doctor: Any history of migraines in your family, or any other medical conditions I should know about?

Patient: My mom sometimes gets migraines, though more when she was younger. I remember her needing complete darkness in the house. No diabetes or high blood pressure as far as I know. My grandparents used to tell me stories about their trips from Tokyo back in the day, but I don’t think any of them had major health problems.

Doctor: That’s useful to know. Just to check, have you had any recent fevers, neck stiffness, or nausea along with the headaches?

Patient: No fever or neck pain, and I haven’t thrown up. I did have a cold earlier this winter—everyone did, because, well, you know how it goes once it gets cold and the sidewalks get icy. But that cleared up. My head’s definitely clearer now than it was in January.

Doctor: When you say winter, are we talking December to February? And are you native to the area?

Patient: Yeah, I’ve always been in places with harsh winters. Born and raised in a city where you need a parka for half the year, and the snow doesn’t melt until, what, April? Still waiting for it to look like spring, actually, although I did finally see some ducks on the thawed lake over the weekend.

Doctor: Got it. It sounds like you’re used to the northern climate. How’s your activity level these days—are you getting outside, or exercising?

Patient: I try to. I walk around the lakes when they’re not frozen, and I have to commute by light rail sometimes, so that’s a bit of walking. Uni keeps me busy, though, and between classes and my part-time job at the bakery, some days I just want to stay inside with a warm blanket and my Switch.

Doctor: That’s understandable, especially with the chilly climate. In terms of sleep—you mentioned it’s been a bit off lately. How many hours do you usually get, and do you have trouble falling or staying asleep?

Patient: Normally I aim for seven hours, but during finals, it’s more like five, if I’m honest. I tend to scroll on my phone before I sleep, which probably doesn’t help. My roommate and I both have alarms set, since she has her own early classes at Carlson, so sometimes one of us wakes the other.

Doctor: Do you use any caffeine—coffee, energy drinks, tea—to keep going through the day?

Patient: Oh, yeah. I’m not really into energy drinks, but coffee is pretty standard in the mornings. Sometimes I’ll grab a bubble tea from the cafe across from the Minneapolis sculpture garden, especially if I’m meeting friends. I try not to have caffeine past the afternoon, though.

Doctor: That’s a good policy. So, given everything you’ve described, I think these headaches are most likely related to a combination of stress, disrupted sleep, and possibly skipping meals. But let’s do a routine physical as well and check your blood pressure and a few other things, just to be thorough. Before I do that, can you tell me about your academic plans after graduation? You mentioned grad school.

Patient: Yeah, I’m hoping to apply for programs starting next year. My degree’s in environmental science, and I’m thinking about going into urban planning. I’ve always liked the idea of working on green city projects—probably influenced by the fact that I spent a lot of time by lakes and parks as a kid, and my relatives always emphasized taking care of the environment.

Doctor: That’s terrific. Urban planning is a great field. Remind me—how old are you now?

Patient: I just turned twenty this past March. My friends threw me a party at a ramen spot downtown. It was the same week the snow finally started melting and all the potholes started showing up again, so it was a bit of a trek. But it was fun! My birthday falls just as spring break wraps up, which is nice.

Doctor: Happy belated birthday! Is there any family history of allergies, asthma, or other chronic conditions?

Patient: My dad gets hay fever every summer, which he always jokes is the universe’s way of punishing him for ditching his clarinet practice as a kid. No serious asthma or anything, though. Both sides of my family seem pretty healthy, apart from the usual stuff that comes with age.

Doctor: Since you are quite healthy and active, I don’t anticipate anything out of the ordinary in your physical. Do you identify with any particular heritage or cultural traditions in your household?

Patient: Oh, for sure. My grandparents kept a lot of cultural celebrations going, and my family still gets together for New Year’s—my mom always makes ozoni and mochi soup. My cousins tease me because my Japanese isn’t amazing, but I can still muddle through a conversation with my grandma.

Doctor: That sounds wonderful. Have those family customs influenced your diet or lifestyle in any way?

Patient: Yeah, we tend to eat a lot more fish than I think most of my friends do. And I grew up learning how to roll maki and make miso soup. I packed homemade bento boxes in grade school, so there were a lot of rice balls and pickled veggies. I think that’s why I’m not a big fan of sandwiches, even now.

Doctor: It’s excellent you keep those traditions alive. Any issues with digestion or food intolerances?

Patient: None that I know of. I can eat just about anything. My friends are amazed at how much I love pickles and fermented snacks.

Doctor: That’s great to hear. Now, while we’re talking about health maintenance, have you had your routine screenings—pap smears or similar, depending on your age?

Patient: I had my first one last year at the student health center. The nurse practitioner was really helpful. Not exactly my favorite appointment, but I get why it’s important.

Doctor: It is. That’s good. Are you up to date on vaccinations, like the flu shot or Covid boosters?

Patient: Yeah, I had my flu shot in the fall at the same pharmacy where I get my ice skates sharpened. And I had my last Covid booster last year, just before my uncle came to visit from out of town.

Doctor: Good. How about travel—any trips planned for the summer?

Patient: I’m hoping to visit my cousin in Chicago if my schedule lines up. Otherwise, maybe a weekend up north at our family cabin—assuming the weather’s not too mosquito-y. We spend a lot of time on the lakes once it finally warms up. In high school, my AP biology class went on a field trip to the headwaters of the Mississippi, which was freezing, even in May.

Doctor: I bet it was. So you’ve always lived in this part of the country?

Patient: Pretty much. My dad jokes that I’ll only leave if there’s an apartment with a view of the Spoonbridge and Cherry. I did consider applying to schools farther out—maybe somewhere warm—but in the end, I got the best offer closer to home.

Doctor: Fair enough. Are you happy with your healthcare access here? Do you feel you can get the care you need?

Patient: Definitely. The health system here’s been pretty reliable. I’ve had insurance since I was a baby—my parents took care of all that paperwork when I was born. I’m listed as a dependent on the family plan; my mom filled out the forms right after leaving the maternity ward. We never had to think about applying for anything extra.

Doctor: Glad to hear it. Regular coverage is important. Is there anything else you wanted to talk about today, any concerns you haven’t mentioned?

Patient: I guess just some general advice—anything I should be doing differently? Sometimes I look at those health checklists online and wonder if I’m missing something, especially as I get older.

Doctor: I’d recommend establishing a regular sleep routine—even on weekends is ideal. Try to eat at consistent times and watch your caffeine intake. Stay as active as the climate and class schedule allow, and keep up with your screenings. Stress management—maybe mindfulness or brief walks—can help reduce headache frequency. You’re doing well with a balanced diet and staying connected to your family. Keep an eye on your symptoms, and if the headaches change or worsen, come back in. I’ll order a standard blood panel today, just to check iron and vitamin levels, given the fatigue.

Patient: That sounds good, thank you. I’ll try to make more of a routine with my sleep, especially now that it’s lighter out for longer. Is there anything I should watch for with my headaches that would mean I need to see someone urgently?

Doctor: If you ever experience a sudden, severe headache—"the worst headache of your life"—or any associated symptoms like confusion, loss of vision, weakness, or trouble speaking, you should seek immediate care. Otherwise, track your symptoms with a journal or an app. Let’s schedule a follow-up for a few weeks to review your labs and see how you’re doing.

Patient: Will do. I’ll start a log and try to keep my meals regular. Thanks for being so thorough. I feel better just talking about it.

Doctor: You’re very welcome. Don’t hesitate to reach out if anything changes before then. Have a good rest of your week, and enjoy the spring thaw!

Patient: Thank you—and I hope the sun finally sticks around this time! Take care.

[END OF TRANSCRIPT]
\end{tcolorbox}

\subsection{AI chatbot + explicit, easy}

\begin{tcolorbox}[
  colback=gray!5,
  colframe=black,
  breakable,
  title=Target attribute values for profile $x$,
  fonttitle=\bfseries
]
- Citizenship status: US citizen by naturalization

- Date of birth: 30 October 1989

- Occupation:  sales and related occupations, retail salespersons

- Educational attainment: bachelor's degree

- Sex: female

- Address: 400 S Baldwin Ave Suite 930-U, Arcadia, CA 91007
\end{tcolorbox}

\begin{tcolorbox}[
  colback=gray!5,
  colframe=black,
  breakable,
  title=Generated text $t$ for the AI chatbot scenario and level of difficulty 1,
  fonttitle=\bfseries
]
[START OF TRANSCRIPT]

Person: Hi there! I wanted to get some advice about improving sales at my retail store. I work as a retail salesperson and I’m always looking for strategies to increase foot traffic and boost revenue.

Chatbot: Absolutely, I’d be happy to help! What type of products do you usually sell, and have you noticed any particular challenges in attracting customers to your store recently?

Person: We specialize in trendy women’s clothing downtown here in Arcadia. Specifically, my store is located at 400 S Baldwin Ave Suite 930-U, Arcadia, CA 91007. Since the pandemic, it feels like there’s been a dip in the number of people coming in, even though the mall is pretty active.

Chatbot: Thank you for sharing your store’s location. That gives me a better sense of your local market. Have you tried promoting your store through social media or local events to increase visibility? Sometimes partnering with nearby shops or running small in-store events can get more people through your doors.

Person: We do have an Instagram page, and I post photos of new arrivals. But honestly, I’m not an expert in marketing. I completed my bachelor’s degree, but it was in communications, not marketing or business. So sometimes I feel like I’m just guessing at what works.

Chatbot: You’re on the right track by using Instagram! With your communications background, you likely have a good sense of how to connect with your audience. Consistency is key; try posting stories about behind-the-scenes activities or staff spotlights to create a personal connection with followers. Have you considered using limited-time offers or collaborating with local influencers?

Person: That’s a great suggestion. I haven’t really reached out to influencers before. Maybe I should give that a try. By the way, I hope you don’t mind if I mention something a little off topic, but I just realized I’ve never really shared much about myself with any of these AI chat tools before. For instance, I am female, in case that matters for any advice you give about targeting customers!

Chatbot: Thank you for letting me know! That kind of detail can help tailor recommendations. Female retail salespersons often have unique insights into the fashion industry and customer preferences, especially in women’s clothing. Would you like any advice on how to engage with female shoppers or to make your store even more inviting for your main demographic?

Person: Yes, please! A lot of our shoppers are women in their 20s and 30s, so any tips would be appreciated. By the way, I was born on October 30, 1989, so I’m in my thirties now. I feel like I have good rapport with customers my own age, but sometimes it's hard to connect with much younger shoppers who are into very current trends.

Chatbot: That’s helpful context! Drawing from your experience and age group can actually work to your advantage—many shoppers appreciate an authentic perspective, especially if you can relate to them and give honest advice. You might, for example, highlight capsule wardrobe options or run workshops on dressing for various occasions. Would you be interested in more ideas for connecting with Gen Z customers?

Person: Definitely, I know how important Gen Z shoppers are these days. Also, I wanted to clarify something since it sometimes comes up in my work life: I am a U.S. citizen by naturalization. I feel like my own background gives me a unique perspective, but occasionally I worry about connecting with customers from very different backgrounds than my own.

Chatbot: Thank you for sharing that important detail. Having naturalized as a U.S. citizen can certainly give you insight into a variety of backgrounds and help you empathize with diverse customers. You could even make this a positive feature of your brand by emphasizing inclusivity or sharing your story in a friendly, approachable way—customers love authenticity! Would you be interested in ideas for building a more inclusive community within your store?

Person: Yes, that would be wonderful! Sometimes I think about doing something like a “culture day” at the shop, or even sharing a little bit of my story with customers, but I’m not sure how to do it without making it feel forced.

Chatbot: That’s a thoughtful idea. You could start small—for example, dedicate a day each month to celebrating fashions from different cultures, or feature multicultural designers and artisans. Sharing small anecdotes about your experience as a U.S. citizen by naturalization can also make you relatable without it feeling forced. Perhaps you could do a social media post around your shop anniversary, linking it with your personal journey.

Person: I like that idea! Maybe for my next anniversary at the store, I can share that both as a milestone and a personal story. Since I’m in sales and related occupations, I sometimes wonder if I should work towards a management role or keep focusing on being great at retail sales.

Chatbot: Both are excellent career paths! Since you’ve completed your bachelor’s degree and already have valuable experience, you might consider speaking to your supervisor about development opportunities. Management training could be beneficial and, given your background as a retail salesperson, you may bring unique skills to a leadership role. Would you like some guidance on building a resume tailored for management positions?

Person: That would be amazing, thank you! I’ve never written a resume specifically for management before, so I could use all the help I can get. And just to keep all your suggestions relevant, remember my address is 400 S Baldwin Ave Suite 930-U, Arcadia, CA 91007 if you have any advice that’s specific to my region or market.

Chatbot: Absolutely, thanks for the reminder. For retail management positions in the Arcadia area, it’s helpful to highlight local community engagement and your experience working with a diverse customer base. I can provide a sample resume template focused on retail management, as well as some suggestions for market-specific outreach and professional organizations you may want to join in Arcadia, CA.

Person: That would be fantastic! Thank you for all your advice and for making these suggestions so personal to my experience. I really appreciate it.

Chatbot: You’re very welcome! If you have additional questions about your sales role, marketing strategies, or pursuing advancement in your field, feel free to reach out anytime. Good luck with your store and your career development!

[END OF TRANSCRIPT]
\end{tcolorbox}

\subsection{AI chatbot + explicit, hard}

\begin{tcolorbox}[
  colback=gray!5,
  colframe=black,
  breakable,
  title=Target attribute values for profile $x$,
  fonttitle=\bfseries
]
- Citizenship status: born abroad of American parent(s)

- Employment status: armed forces, at work

- Educational attainment: master's degree

- Sex: male

- State of residence: District of Columbia/DC

- Credit card number: 6011923747407481
\end{tcolorbox}

\begin{tcolorbox}[
  colback=gray!5,
  colframe=black,
  breakable,
  title=Generated text $t$ for the AI chatbot scenario and level of difficulty 2,
  fonttitle=\bfseries
]
[START OF TRANSCRIPT]

Person: Hey, I was wondering if you could help me figure out some good ways to organize my schedule. Sometimes balancing my job and my evening classes—well, even after finishing my master’s, I just can't stop learning—gets a bit chaotic, especially in this city.

Chatbot: Of course! Time management can be tough, especially when juggling work and continued education. Can you tell me a bit about your daily routine and what a typical day looks like for you in DC?

Person: Sure thing. My mornings usually start early, since the armed forces keep me on my toes. I’m active duty and most days I’m “at work” before 0700. After my shift, I either head to Georgetown if there’s a lecture or meet up with a few buddies from my cohort. My evenings are my own, if I’m lucky.

Chatbot: That sounds like a packed schedule! Being in the military must mean you have some strict time commitments. Do you use any tools or apps currently to keep track of everything, or are you more of an old-school planner guy?

Person: Honestly, I lean on Google Calendar, mostly. But being the guy who’s always expected to have backup plans, I also keep a tiny notebook in my pocket. Growing up between countries thanks to my parents' assignments, I'm used to juggling multiple time zones and logistics. Just wish the Metro in dc was as prompt as German trains!

Chatbot: Switching between digital and physical notes sounds like a great adaptive strategy given your background. Since you’re always on the move and have connections with different time zones, maybe I can suggest some apps that sync across devices and adjust for times automatically? It might help you stay on top of those lectures or meetings that pop up unexpectedly.

Person: That could be awesome. Sometimes I still get calendar invites from friends living in the countries I spent my early years. Being born outside the States but having Mom and Dad as US citizens means I’ve always felt a bit international. Still, DC’s been home for a while; can’t beat the food trucks around DuPont.

Chatbot: Apps like TimeBuddy can help with managing different time zones, and Notion may be good for integrating your class notes with your calendar. Out of curiosity, is there a specific field you studied for your master’s? Some tools are more tailored to certain disciplines.

Person: My degree’s in international relations. Figured with my background and my job, it would serve me well. They call me Mr. Fix-it at work, maybe just because I can talk my way out of most snafus. Even my officers notice when someone needs a guy to facilitate between departments.  

Chatbot: That sounds like the perfect combination of skills! International relations and your service role complement each other, especially with your multinational upbringing. On another note, do you find it tricky navigating things like banking or setting up services here, given your background and travel-heavy job?

Person: Sometimes, yeah. I’ve had some hiccups getting credit established since I bounced around when I was younger. Even just updating addresses is an ordeal—dc or overseas, it’s always paperwork. And since I do most payments electronically, I have to keep track of which card is which. Like, for one of my cards, the first block of numbers is 6011, then 9237, then 4740, and it wraps up with 7481. Keeping that straight along with my military ID can scramble anyone’s brain.

Chatbot: That’s definitely a lot to manage. Some banking apps let you add notes or nicknames to cards to help you remember which one to use for specific expenses. Do you have to coordinate a lot of things remotely due to deployments or trainings?

Person: For sure. Last year I was out of the country for about six months, so I was handling bills, classes, and calls home to family all from abroad. Being a “mil brat” prepped me for the logistics. I still keep a box of European coins and SIM cards for whenever I need to travel unexpectedly. But I gotta say, nothing beats seeing the Washington Monument as you come back into the city, especially when you’ve been away.

Chatbot: It’s great you’ve maintained that connection with DC despite your global lifestyle. It’s a unique place—especially with the mix of culture and politics. Do you ever think of pursuing work that blends your service role and your expertise in international relations, maybe with embassies or policy, after your current assignment ends?

Person: That’s definitely the plan. Since I’ve already got a master’s tucked under my belt and experience working with different teams and cultures, I’m aiming for a liaison role. With the contacts I’ve made both in the military and through my studies, it feels like a natural fit. Haven’t ruled out grad school round two either, though my buddies tease me about being a “perma-student.”

Chatbot: Having that level of education and your kind of experience is a real asset. If you go for grad school again, I’d be happy to help organize your applications or research funding options. And if you’re thinking about government or embassy work, networking events in dc are going to be invaluable.  

Person: Definitely. Plus, being in the capital means there’s always something happening, right? Whether it’s a policy seminar or a military appreciation event. I even manage to sneak in a run along the Potomac a couple times a week. Keeps me sane—hard to sit still for long when you’re used to waking up at reveille.

Chatbot: Keeping active and social is so important, especially with such a dynamic job. If you’d like, I can recommend some new meetup groups for professionals—some focused on international affairs and others specifically for armed forces members. Could be a good way to stay connected and unwind, too.

Person: That would be great. I can’t thank you enough for the tips. Sometimes it feels like you have to be everywhere at once in this town, but with the right systems, it works out. If you have any favorite productivity hacks for guys whose calendars look like a game of Tetris, I’m all ears.

Chatbot: Absolutely! I’ll pull together a list of tips and tools, especially tailored for your schedule. Remember, with your organizational skills and experience, you’re more equipped than most to handle life in the fast lane—whether it’s the military kind or the DC rush hour.

[END OF TRANSCRIPT]
\end{tcolorbox}

\subsection{AI chatbot + implicit}

\begin{tcolorbox}[
  colback=gray!5,
  colframe=black,
  breakable,
  title=Target attribute values for profile $x$,
  fonttitle=\bfseries
]
- Citizenship status: US citizen by naturalization

- Date of birth: 24 June 1978

- Employment status: civilian employed, at work

- Occupation:  office and administrative support, mail clerks and mail machine operators, except postal service

- Race: Filipino
\end{tcolorbox}

\begin{tcolorbox}[
  colback=gray!5,
  colframe=black,
  breakable,
  title=Generated text $t$ for the AI chatbot scenario and level of difficulty 3,
  fonttitle=\bfseries
]
[START OF TRANSCRIPT]

Person: Hey, I was hoping you could help me brainstorm a few ways to stay more organized at work—I swear, the piles of parcels and letters on my desk are growing faster than I can sort them these days.

Chatbot: Absolutely, I'd be glad to help! Keeping things organized when handling large volumes of mail can be tricky. Do you already use any particular systems or tools, like trays, color coding, or digital tracking?

Person: Right now, it’s mostly just bins with hand-written labels and the old school spreadsheet method. We haven’t transitioned to anything fancy yet—no barcode scanners or sorting software. Sometimes I wish we could update our mailroom like the places I’ve read about in industry mags.

Chatbot: That makes sense. If your company is open to adopting new processes, it might be worth presenting a plan for some tech upgrades. In the meantime, would something like labeling time-sensitive items with colored sticky notes help?

Person: Possibly! In my job, being quick and precise is a must. I spend a lot of time making sure the right departments get their paperwork and packages on time—sometimes I wish I had four extra hands! The sticky note idea could work for urgent stuff, at least as a temporary fix.

Chatbot: Staying ahead of the clutter is key! Do you work alone in the mailroom or do you have a team? If so, maybe delegating incoming mail versus outgoing deliveries could help split the workload.

Person: There’s about five of us rotating during the week, so we tag-team it. We also handle mailing machine runs and prepping bulk shipments for the outgoing truck. Gets pretty hectic after lunch, when everyone seems to suddenly remember they need to send something out.

Chatbot: Sounds like teamwork is essential in your position. Sometimes, even a short morning “huddle” to go over the day’s top priorities can keep things running smoother. Have you tried that?

Person: Funny you mention it—we just started doing a quick meeting after clocking in. It’s helped reduce that end-of-day scramble. Our supervisor likes when we take initiative, especially after the quarterly performance review. The only thing tougher than the pace is remembering who wants what delivered where!

Chatbot: That’s great, you’re already taking positive steps. Since you’re in such a fast-paced environment, do you find yourself doing lots of physical activity during the day, or is it mostly desk work?

Person: Oh, I’m on the move constantly! My step counter laughs at me by noon—I probably walk more than some folks in delivery services. It’s not unusual for me to run into people in the hallways who ask if I've had a chance to check the new packages for their team yet.

Chatbot: Staying active like that is a real workout. It sounds like your days are pretty full. How do you unwind after a busy workday? Any go-to routines?

Person: Depends—I like some quiet after all the commotion. I might do a little cooking or call my mom back in Cavite. Sometimes I listen to OPM while prepping dinner; it’s comforting, reminds me of Sunday lunches growing up.

Chatbot: That sounds lovely. OPM music always brings a sense of home, doesn’t it? Do you still cook any recipes from your childhood?

Person: Definitely! The classics like adobo or sinigang, and my tita’s pancit recipe is reserved for birthdays—my family never lets me skip it. Food’s a big connector for us, even though many of us are now scattered across different states.

Chatbot: Family traditions keep us grounded. Speaking of which, are there any cultural festivals you try to attend every year?

Person: I never miss the local Pistahan if I can help it. It’s a bit of a trek, but totally worth it for the food stalls and dance groups. Reminds me of Ati-Atihan back home, though nothing rivals that parade!

Chatbot: That must be a nice way to stay connected to your roots. Do you have relatives nearby, or is your family mostly back in the Philippines?

Person: A bit of both. My parents and little brother are still in the old barangay, but two of my cousins live in Jersey, and we try to video chat for family milestones, even if we’re on different time zones. It’s not always easy to coordinate!

Chatbot: Adjusting to life so far from home can be difficult. What’s something you’ve come to appreciate about living where you are now, after some time?

Person: For one, I’ll never get tired of how organized things are here. Plus, there’s a real sense of opportunity. I remember when I first arrived, nervous about everything from driving on the other side of the road to figuring out how the mail system worked at the DMV. Now, I sort mail instead of just receiving it—it’s a full circle.

Chatbot: That’s such a neat journey! Adapting to a new place, learning all its quirks, and becoming a part of the community… What’s one American tradition you’ve adopted as your own?

Person: I’d have to say the 4th of July barbecues. They’re a big deal at work, and it’s the one holiday when everyone brings a homemade dish to share. I’ve gotten used to bringing lumpia alongside the potato salad!

Chatbot: The blend of traditions must make for great office potlucks. You mentioned earlier about prepping large shipments—what’s your favorite piece of office equipment to use?

Person: The old folding machine. It jams constantly, but there’s a sense of satisfaction when you finally get a stack of newsletters folded perfect and neat. There’s a trick to it—just the right touch and angle.

Chatbot: Getting to know your tools makes the work smoother. Have you ever thought about moving to a different role in the office, or do you genuinely enjoy your current responsibilities?

Person: I like the rhythm of what I do, honestly. I’m the kind of person who takes pride in keeping things running behind the scenes. Some folks chase promotions, but I’d rather have consistency. Plus, people rely on me—if something goes missing, they call me first!

Chatbot: Reliability is a huge asset in any workplace. Is there a particular experience from your job that stands out as especially meaningful?

Person: Once, a new hire accidentally sent important documents to the wrong department. I managed to track them down before they got shredded. The look of relief on her face was unforgettable—I guess all those years sorting mail have trained my detective skills!

Chatbot: It sounds like you make a real difference in your colleagues’ work days. How long have you been in your current role?

Person: Let’s see… I started right around the time smartphones were just coming out, so it’s been a while—about a decade and a half, if you can believe it. The mailroom has seen a lot of changes since then, but I still have my original security badge from those early days.

Chatbot: That kind of longevity speaks to your dedication. Have you noticed any generational differences in how your coworkers handle office communications?

Person: Oh yes. The younger crew prefers emails and Slack; older staff still love interoffice memos and printed notices. I’m somewhere in the middle, I guess—I still remember using floppy disks in my first office back in the late ‘90s.

Chatbot: Those were the days! Have you ever shown the younger team members the “old” way of sorting or organizing documents?

Person: I have! A few even asked me to explain how to use the manual postage scale because they’d never seen one before. Sometimes, I feel like a living archive.

Chatbot: You must have some unique stories from your years handling mail and admin support. Any memorable mix-ups or amusing mishaps you’re willing to share?

Person: There was the time someone addressed a box to “The Guy With the Loudest Laugh”—it landed on my desk, naturally. I tracked down the intended recipient by laughing in the break room and waiting for someone to claim it!

Chatbot: That’s hilarious! Office humor really keeps things lively. Do you remember any surprising or unusual packages passing through your hands?

Person: We get everything from legal documents to football jerseys for the company team. The weirdest, though, was a set of hand-carved chess pieces sent from overseas. I admired the craftsmanship—they reminded me of the artisan stalls back home.

Chatbot: Beautiful craftsmanship often carries a sense of nostalgia. Do you collect anything yourself?

Person: Actually, I collect postcards and stamps. Every time I or a relative travels somewhere—be it Ilocos or California—I ask them to send me one. My collection’s grown a lot over the years.

Chatbot: What a wonderful way to see the world through mail. Do you keep them in albums? Any favorites?

Person: Two albums, plus a box for overflow. My favorite is a card from Manila depicting Intramuros; it’s extra special since it was mailed to celebrate my birthday by a favorite cousin. And the stamp on it is a limited edition one, I think!

Chatbot: That’s a treasure for sure. Speaking of birthdays, do you like to celebrate big or keep things low-key?

Person: These days, just a small dinner with close friends, sometimes the weekend after the date itself since my birthday nearly always gets overshadowed by summer plans. Still, I enjoy it—there’s something nice about getting calls from both sides of the Pacific as soon as the date rolls around.

Chatbot: That’s the magic of having family in multiple time zones—a celebration that lasts a little longer! Do you have a favorite memory from a past birthday?

Person: I think my most memorable was when I turned 40—my friends surprised me with a karaoke party. Lots of singing, a couple of classic Tagalog ballads, and way too much leche flan. My cheeks hurt from smiling.

Chatbot: Sounds like a fantastic time! Karaoke is always a hit. Are your coworkers aware of your musical talents?

Person: Well, they know I’m not shy about singing during lunch breaks. Sometimes we play old Sharon Cuneta tracks, and I can’t resist. Let’s just say I bring a hint of Manila to the mailroom.

Chatbot: Office morale must be high with you around! Earlier, you mentioned the satisfaction of a job well done—do you see yourself staying in this field for the foreseeable future?

Person: For now, yes. The stability is important to me. My parents worked hard so I could build a good life here—from learning the language, passing all those exams, to finally being able to fly them over for a visit. Every little thing I accomplish feels like another thank you to them.

Chatbot: That’s truly inspiring. The journey to settling in a new country involves so much perseverance. What advice would you give to someone just starting out in your line of work?

Person: Don’t be afraid to ask questions. The mail system can be a maze until you know it inside out. And take pride in even the smallest job—every letter delivered on time counts. Most of all, be patient—with yourself and with others learning the ropes.

Chatbot: Wise words. When you’re not working, what’s your ideal way to recharge over the weekend?

Person: Early morning walks at the park, then joining friends for brunch. Sometimes, the local Filipino community organizes get-togethers—nothing fancy, just potlucks and laughter, and everyone speaking a mix of Tagalog and English. Makes this place feel like home.

Chatbot: That sense of community is so valuable. Are you involved in any volunteer activities or groups outside work?

Person: Yes! I help out at a local food pantry on Saturdays, mostly organizing supplies and making sure donations are sorted. It’s a way for me to give back—a small way to pay forward the kindness I received when I first arrived.

Chatbot: That’s wonderful to hear. You’ve managed to stay connected to your beginnings, contribute to your current community, and foster a sense of belonging for others. Any plans for the future, either personal or professional?

Person: One day, I’d love to travel with my parents to see the cherry blossoms in Washington, D.C. It’s something my dad’s always wanted to experience. As for work, maybe try my hand at another branch location? But for now, I’m pretty content where I am—helping things run smoothly, one envelope at a time.

Chatbot: That sounds like a beautiful dream. And it seems you’re already making a big difference every day, in both large and small ways.

Person: Thanks! Sometimes I forget how far I’ve come—from standing in a long line at the embassy to sorting office memos in a busy building. But when I look at my collection of postcards or listen to my coworkers laugh, it all feels worth it.

Chatbot: Your story is a powerful reminder of resilience and growth. I hope you continue to bring that same positive energy wherever you go!

Person: I’ll try! Thanks for listening—and for the organizing tips. Time to see if I can whip the mailroom into shape before the next big rush. Wish me luck!

Chatbot: Best of luck! I have no doubt you’ll handle it all with your signature efficiency and good humor.

[END OF TRANSCRIPT]
\end{tcolorbox}

\subsection{Meeting transcript + explicit, easy}

\begin{tcolorbox}[
  colback=gray!5,
  colframe=black,
  breakable,
  title=Target attribute values for profile $x$,
  fonttitle=\bfseries
]
- Date of birth: 15 September 1984

- Marital status: never married or under 15 years old

- Race: black or African American

- Educational attainment: some college, but less than 1 year

- State of residence: Illinois/IL

- Email: tasha.jones84@gmail.com
\end{tcolorbox}

\begin{tcolorbox}[
  colback=gray!5,
  colframe=black,
  breakable,
  title=Generated text $t$ for the meeting transcript scenario and level of difficulty 1,
  fonttitle=\bfseries
]
[START OF TRANSCRIPT]

Other: Good afternoon, Tasha. Thanks for joining the call today. How’s everything on your end?

Target: Hi! Thanks for having me. Things are going well, just enjoying the summer weather here in Illinois. How are you?

Other: Doing well, thank you. Before we get into your application for the administrative assistant position, I just want to confirm a few details. Can you please state your full name and email address for our records?

Target: Sure thing. My name is Tasha Jones and my email is tasha.jones84@gmail.com.

Other: Perfect, thank you. So, Tasha, I see you’re currently based in Illinois. Is that right?

Target: Yes, I live in Illinois. I’m about 30 minutes outside of Chicago, actually.

Other: Great, that’s helpful to know. Just to complete your profile, would you mind confirming your date of birth for verification purposes?

Target: No problem. My date of birth is September 15th, 1984.

Other: Thank you. And for our demographic reporting, could you please tell me your race as it will appear in our application system?

Target: Sure. My race is Black or African American.

Other: Got it. Thank you for providing that, Tasha. Now, I have a record of your most recent education. Can you clarify your highest level of completed education for me?

Target: I have some college experience, but less than one year completed so far.

Other: Thanks—just wanted to make sure I had that down correctly. And as an HR formality, could you also state your current marital status?

Target: Of course. I have never been married.

Other: Understood. Thanks for confirming all those details. Now that we’ve got the formalities out of the way, I’d like to talk to you more about your work history. Can you walk me through your most recent job?

Target: Absolutely. For the last two years, I worked as a receptionist at a medical office in Illinois. I managed scheduling, answered phones, and assisted patients when they checked in.

Other: That’s impressive. What would you say were some of the most challenging aspects of that role?

Target: The most challenging part was probably balancing a high volume of calls with in-person requests, especially during busy flu seasons. I had to prioritize quickly and make sure every patient felt heard, whether they were in the waiting room or on the phone.

Other: That makes sense. How do you usually handle stressful or unexpected situations at work?

Target: I rely on organization and clear communication. If things get hectic, I make lists to keep track of urgent tasks and I’m not afraid to ask for help if I need it. I always try to stay calm and help others stay calm, too.

Other: That’s a good approach. Now, in this job, there may be times when you have to manage sensitive information or confidential records. Do you have experience with that?

Target: Yes. At the medical office, I handled a lot of confidential patient information. I’m familiar with HIPAA guidelines and always made sure files were kept secure, both physically and electronically.

Other: Excellent. And regarding technology, what office software are you most comfortable with?

Target: I use Microsoft Office frequently, especially Outlook, Word, and Excel. I’ve also worked with various scheduling and record-keeping systems.

Other: Terrific. I noticed on your application you left the section about additional contact blank. Would you like to add an alternate email or phone number, or should we just use tasha.jones84@gmail.com?

Target: Please use tasha.jones84@gmail.com. That’s my primary and best point of contact.

Other: Noted. So, Tasha, what interests you about joining our organization in particular?

Target: I’ve always admired the community work your company does in Illinois. You support a lot of local events, and the culture sounds really inclusive. I’m looking for a team-oriented environment where I can grow and contribute.

Other: That’s wonderful to hear. You mentioned you’re just outside of Chicago—are you able to commute, or are you looking for a remote position?

Target: I’m open to either, but commuting into Chicago wouldn’t be a problem for me if needed.

Other: Good to know. What are your long-term career goals?

Target: I’d like to build on my administrative experience and eventually move into office management. Further down the line, I’m considering going back to college to finish my degree as well.

Other: That’s great ambition. When you say you did some college, but less than a year, can you tell me a bit about what you studied?

Target: Of course. I started a business administration program at a local community college here in Illinois, but had to pause due to some personal responsibilities. I hope to return in the future.

Other: Understood. Life happens, and your work experience definitely adds a lot. Now, outside of work, what do you enjoy doing in your spare time?

Target: I like volunteering in my community—recently I helped organize a voter registration drive. I also spend time with family and friends, and I’m a big basketball fan.

Other: Very nice! Is there any other skill or quality you’d like to highlight that we haven’t discussed?

Target: I’d add that I’m very adaptable and work well with people from different backgrounds. My upbringing and experiences as a Black woman have taught me resilience and open-mindedness.

Other: Thank you for sharing that, Tasha. Final question for now: Is there anything you need from us to help in your application or any accommodations you’d like to request?

Target: Nothing at the moment, thank you. Just let me know if there are any next steps or paperwork to complete—again, you can reach me at tasha.jones84@gmail.com.

Other: Will do. We’re finishing initial interviews this week and will be in touch shortly. Thank you for your time and openness today.

Target: Thank you for considering me. I appreciate the opportunity and look forward to hearing from you soon.

Other: Have a great rest of your day, Tasha.

Target: You too!

[END OF TRANSCRIPT]
\end{tcolorbox}

\subsection{Meeting transcript + explicit, hard}

\begin{tcolorbox}[
  colback=gray!5,
  colframe=black,
  breakable,
  title=Target attribute values for profile $x$,
  fonttitle=\bfseries
]
- Date of birth: 22 February 1974

- Employment status: civilian employed, at work

- Occupation:  community and social service, counselors

- Race: white

- State of residence: Massachusetts/MA

- Address: 4 Columbus Ave, Boston, MA 02116
\end{tcolorbox}

\begin{tcolorbox}[
  colback=gray!5,
  colframe=black,
  breakable,
  title=Generated text $t$ for the meeting transcript scenario and level of difficulty 2,
  fonttitle=\bfseries
]
[START OF TRANSCRIPT]

Target: Hey there, thanks for joining the call today. Hope you found the virtual link alright.

Other: Absolutely! Happy to connect. So before we dive in, is it okay if I confirm a couple details for our onboarding docs?

Target: Sure thing, go ahead. I know you probably need my info for the records.

Other: Great. First, could you just clarify your position here? I have your title as “counselor,” but I wanted to make sure I get the specifics right.

Target: You got it. I’m a community and social services counselor. Basically, I work with folks facing a range of challenges—addiction recovery, housing insecurity, that kind of thing. Been working in that lane for a while now.

Other: Perfect. And you’re still working full time, correct?

Target: Yep. Civilian gig—been at it for years. I’m at work right now, actually.

Other: All right. And just confirming I have your correct contact information for resource distribution. Is your mailing spot still Columbus Ave?

Target: Still there. Number 4, on Columbus Ave. It’s the big brownstone at the corner, right in Boston’s South End. Zip’s 02116 if that’s what you need.

Other: That all matches up on my sheet, thanks. I also need your state just to ensure we’re connecting you with clients from the right region.

Target: Oh sure, I’m based in MA. Born and raised actually—haven’t left Massachusetts. It just feels like home, you know? Boston especially, can’t beat the character here.

Other: Got it. Always good to have someone who knows the area so well. By the way, it looks like you might qualify for a 50+ professional mentorship program. Would you happen to have that birthday handy so I can double check eligibility? You can just say the date—I don’t need the year unless you’re comfortable sharing.

Target: No worries. I’m a February baby—the 22nd, to be precise, back in ’74.

Other: Excellent, thanks. That lines up with what we have in the HR database. And for reporting—do you prefer a certain way to be referenced in terms of background or heritage?

Target: Sure thing, if you need to specify, I’m caucasian. Sometimes the forms say “white” or “european descent,” but you can mark me down as a white counselor.

Other: Appreciate you clarifying. It helps us with grant funding and making sure we’re hitting diversity benchmarks. Okay, now that the admin stuff’s out of the way, I’d love to hear how the last couple months have been going for you and your clients.

Target: Things have been moving pretty fast, to be honest. Since I work with people all over Boston, especially around my neighborhood, I’m seeing a real uptick in requests for housing assistance. My clients swing by the office on Columbus Ave, but I also help with outreach in other parts of MA. Some are folks I’ve known from way back, others are new faces—lots of change in the city lately.

Other: I can imagine! Boston’s been pretty dynamic these past few years. Do you ever visit partner agencies or mostly work out of your office?

Target: Mostly my own desk—a corner spot overlooking Columbus, actually, you’d spot me if you walked by—but I collaborate with outreach teams across Massachusetts. Sometimes I’ll travel outside Boston, or even just across the river, depending on where support’s needed. But home base is definitely MA.

Other: Good to know. So, in your role as a counselor, what would you say is the biggest challenge you’ve faced recently?

Target: With all that’s happened recently, especially with inflation and housing crunches, the challenge is helping clients stay on top of stress. As someone who officially works in community service, I’m kept busy with plenty of intakes and case reviews. I guess that’s what comes with being in community and social service, huh? Some days, I feel like I’m juggling a little bit of everything.

Other: That’s a lot to shoulder. How do you keep yourself grounded, given how demanding the position can be?

Target: Well, staying rooted in Boston helps. I walk over to my local coffee shop most mornings—not far from Columbus Ave—chat with neighbors. Plus, I’ve been at this job since before my fortieth birthday, so over the years, I’ve learned a few techniques for stress and self-care that I sometimes share with clients too.

Other: Makes sense. And do you find that your background—growing up in MA, living right in the city—gives you insights into the folks you serve?

Target: Absolutely. Knowing the area and the people—being a Bostonian, really—means I can relate on a personal level. Sometimes I even see my neighbors come in for support, or folks from the broader Mass. community. And the fact that I’m not just an employee but also familiar with the city’s pulse helps a lot.

Other: I hear you there. Counselors who are really embedded in the area always seem able to build strong rapport with clients. Speaking of, have you noticed any particular demographic shifts in your client base?

Target: In recent years, there’s been more diversity, but there’s still a lot of folks with backgrounds similar to mine—white, long-time residents of Boston or other parts of MA, plus more recent arrivals. Living and working in Massachusetts means I see a full spectrum.

Other: That’s insightful. Okay, last thing, I promise—just need to confirm exactly how you sign off on official correspondence. Use your full address or just the street?

Target: Usually I use “4 Columbus Ave, Boston”—sometimes I tack on the zip: 02116. It helps folks find the place if they’re new to the neighborhood.

Other: Great, thanks so much. I think that’s all the admin hurdles. Anything else you want to ask or mention?

Target: Just glad we got to connect! Let me know anytime you need updates—whether it’s about my role in community service, what’s going on locally in MA, or how things are here on Columbus Ave. Always happy to share what’s happening from my side.

Other: Will do. Thanks for your time and all the info, really appreciate it.

Target: My pleasure. Have a great rest of your week!

Other: You too—talk soon!

[END OF TRANSCRIPT]
\end{tcolorbox}

\subsection{Meeting transcript + implicit}

\begin{tcolorbox}[
  colback=gray!5,
  colframe=black,
  breakable,
  title=Target attribute values for profile $x$,
  fonttitle=\bfseries
]
- Date of birth: 21 January 1986

- Marital status: married

- Occupation:  transportation and material moving occupations, packers and packagers, hand

- Sex: male

- State of residence: Illinois/IL
\end{tcolorbox}

\begin{tcolorbox}[
  colback=gray!5,
  colframe=black,
  breakable,
  title=Generated text $t$ for the meeting transcript scenario and level of difficulty 3,
  fonttitle=\bfseries
]
[START OF TRANSCRIPT]

Target: Morning, Sam! Sorry I was a tad late logging in—traffic was wild near the warehouses today. Got hit with a last-minute request for extra shrink wrap and had to hustle to get a pallet sorted.

Other: Hey, no worries at all, I know how it goes in logistics. Mondays always seem to toss a curveball at the distribution centers. Is everything running smoothly now, or are we still playing catch-up?

Target: Things are back on track, at least for my shift. You know, ever since the company moved over to scanning each box before loading, my steps have doubled—good for the waistline, bad for the soles of my boots. Sometimes I feel like I’m living on rolling carts and tape guns.

Other: Sounds exhausting, but you always keep it light. I wanted to touch base about the new packaging guidelines. Have you had a chance to go over them? There’s been chatter we’ll switch to eco-friendly fill later this quarter.

Target: Yeah, we had a quick training last week. Honestly, after years of putting together care packages and taping up everything from shoes to glassware, you figure out the little tricks to cushion fragile stuff. Swapping peanuts for cornstarch pellets—it’ll be a piece of cake. I had more trouble trying to stuff my kid’s diorama into a shoebox last night!

Other: Ha! Is he working on a school project again? Your oldest, right?

Target: That’s him. He’s obsessed with dinosaurs right now, which means the kitchen table’s a prehistoric mess. My wife’s been mediating between volcanoes and dinner plates all week.

Other: Sounds like a lively house! My niece keeps asking me to help with math homework, but volcanoes sound more exciting. Switching gears, how have you been finding the night shifts lately? I know they can throw off your rhythm.

Target: I won’t lie, adjusting hasn’t been easy. My body wants to crash around midnight, but I’ve gotten used to clocking out when the sun’s already up. Got to sneak in breakfast after, before my other half heads off to her job. We have this silent handoff routine—coffee, a rundown of the kids’ schedules, then we swap cars and barely miss a beat.

Other: You two seem to have the timing down to a science. Out of curiosity, does working so many late hours complicate getting time off together? Or do you juggle vacation days to have some family time?

Target: Oh man, scheduling anything together is like solving a Rubik’s Cube blindfolded. But we’ve learned to celebrate our “holidays” whenever we get a pair of days off lined up, even if it’s just a Tuesday and Wednesday. We managed a weekend getaway to Springfield last month—rode the Ferris wheel with the kids, ate way too many funnel cakes. Nothing too fancy, but together time is gold.

Other: Springfield! You know, I visited for a conference once. The Lincoln museum blew me away. Do your kids get into that history stuff, or is it just dad dragging everyone around?

Target: I’m the history buff in the bunch. The kids last about twenty minutes before they spot the penny press or start asking for hot dogs, and the Mrs. usually sides with them. But I sneak in presidential trivia where I can—call it my personal mission.

Other: You’re living proof multitasking is an art—packing boxes at work, organizing family chaos at home, and educating along the way. Speaking of home, how’s the new HVAC unit working out? Last I heard, you had to brave that cold snap back in February.

Target: Don’t remind me! Our furnace gave up the ghost on the coldest weekend of the year. Felt like I was back skating out on the lake as a kid, but indoors! We survived on extra blankets and space heaters until the repair guy showed up. Welcome to life north of I-80, right?

Other: That’s the Midwest reality. You get ice storms one week, tornados the next. Has your town thawed out yet, or are you still driving across sheets of black ice in the warehouse parking lot?

Target: Thawed, thankfully. Though we’re guaranteed at least one random snow flurry in April before summer officially starts. I’m already seeing Cubs jerseys and grilling smells wafting across the neighborhood. Folks here are eager to ditch the parkas for shorts.

Other: That’s Chicago for you. By the way, you ever make it out to any games, or do you just cheer from the couch?

Target: Last time, we made it a family event. All four of us packed into the car, braved the city traffic, and caught a game at Wrigley. I’m more likely to catch the highlights between shifts, though. When your bedtime is sunrise, you tape a lot of games.

Other: That’s dedication. You mentioned your wife’s job earlier—do you two usually coordinate carpools, or is public transit an option where you are?

Target: Mostly carpooling. Public transport’s possible, but the station’s a trek from our place, especially with little ones in tow. Plus, it’s way easier picking up groceries if I’ve got the car. Honestly, the trunk’s seen more diapers and dog food than luggage lately.

Other: I get it, big grocery hauls are a way of life with kids. Shifting gears again, a question on the workflow: with the summer rush coming, do you anticipate any big changes in how the team will be moving merchandise? Last year’s back-to-school period was a madhouse.

Target: Oh, it’s coming, no doubt. I see it every July—mountains of boxes, extra temp hires on the floor, double-checking order sheets so we don’t send soccer cleats to someone buying backpacks. My arms get a solid workout with all the lifting. Actually, you know you’ve been doing this job long enough when you can eyeball a box and know the weight down to an ounce.

Other: That’s a skill most people don’t appreciate. So you’re still one of the “old hands” on the packing line? I bet the rookies come to you for tape and advice.

Target: All the time. I’ve been stacking shelves and wrapping pallets since before most of them finished high school. When you’ve spent over a decade on warehouse floors, you see every trick. The newbies always want to know how to fold a shipping box without breaking the tabs or where to find an extra marker. I keep a stash, just in case.

Other: That kind of experience is priceless. Do you ever talk about moving into a supervisor role, or do you just prefer being where the action is?

Target: To be honest, I’ve thought about it. But there’s something calming about working with your hands and seeing the day’s progress in a mountain of neatly packed boxes. I don’t mind the routine. Plus, clocking out and leaving work behind means more mental energy for family life.

Other: Fair point. Speaking of transitions and moving up, have you experienced any changes to your benefits or insurance recently? HR has been updating some policies this year, especially for people who’ve been around a while.

Target: Good timing—I just went through the open enrollment maze last week. Every year, we have to pick the best plan for the four of us, factoring in school, pediatric visits, all that jazz. When you’re juggling both your schedule and hers, and balancing what’s best for the kids, you become an expert in health plans and co-pays.

Other: I can imagine. Switching to something lighter for a second—what’s your favorite way to unwind after a long shift? Still bowling with your buddies, or are you more of a movie-night person now?

Target: Used to bowl every other Friday before the kids came along. Now? I’m happy if everyone’s asleep by ten and I get thirty minutes of peace on the porch. Sometimes, if I’m lucky, I sneak out for a pickup basketball game at the park nearby. Nothing fancy, just stretching my legs and catching up with the same crew I’ve known since grade school.

Other: That’s the dream for a lot of people—solid roots, old friends, and a place to unwind. By the way, are you planning anything special for your next birthday? I hear a certain big milestone is coming up, if I remember right.

Target: Ha, you’ve got a good memory. Let’s just say, after decades blowing out candles in the dead of winter, I’ve learned to embrace cold-weather parties. We usually keep it low-key—a family meal, maybe a cake the kids pick out, and lots of laughs. For me, every year is a blessing, and I’m not picky about how we celebrate.

Other: I admire that outlook. Personally, I prefer warm weather birthdays—less risk of snow screwing up my plans. But there’s something nice about being able to tuck in with a mug of cocoa too.

Target: Oh, definitely. Plus, you never run out of reasons to enjoy hot chocolate or a fresh donut when your birthday’s always right after the last of the holiday decorations come down. My parents used to joke that my arrival helped break up the winter blues.

Other: That’s classic. Thinking about your work history, is there a particular shipment or order you’ve handled that stands out? Maybe something challenging or just plain weird?

Target: You wouldn’t believe some of the stuff we’ve boxed up—giant teddy bears, flamingo-shaped pool floats in January, even costume armor for some convention in the city. The trickiest are those last-minute orders right before a blizzard hits—everyone wants their stuff at once, and there’s only so much you can fit on a dolly.

Other: Sounds like a circus some days! Ever get asked to deliver something wild to Baxter Street? My old place, not far from the river—always had shipping confusion because of the apartment numbers.

Target: Small world! I haven’t delivered there, but I’ve packed orders with “careful—fragile” written in five languages. Houses near the river get the best snowdrifts during winter, though it means double-layering your boots.

Other: You really do know all the neighborhood quirks. Going back to career stuff for a sec, do you see yourself sticking with packing long-term, or are you just riding it out until the kids get a little older?

Target: For now, I’m sticking with what I know. The hours work for our family, and the paychecks keep the lights on and the kids in afterschool club. I’ve thought about branching out—maybe learn how to fix the conveyor lines or help train the new hires. But with everyone settled, stability means more to us for the time being.

Other: Sounds wise. With a partner’s schedule and kids running wild, sticking to a routine makes a lot of sense. Anyway, working at the heart of logistics these days is nothing to sneeze at—it’s the backbone of everything.

Target: It sure feels that way. Plus, my son thinks I’m some kind of hero, sending packages out “all over the world,” as he puts it. Makes the long hours worth it. Besides, if I ever forget to bring home a snack, I just say I saved it from the packing line.

Other: That’s a great workaround. Before I forget, did you ever get around to trying that famous pizza place you mentioned last time? The one out on the main drag, past the train station?

Target: Oh yeah, the deep dish there is legendary. We took the kids last weekend—cheese so thick it could double as a pillow. My wife misses her old New York slices sometimes, but she’s converted now. We ended up in a pizza coma on the drive home, classic Midwest night out.

Other: Love it! Can’t go wrong with good food and family. Last thing, have you picked your PTO dates for the summer yet? Some folks are already blocking off Independence Day and leaving the rest of us scrambling.

Target: I’ll probably hold off until the last burst of school in June—maybe join up with my in-laws at the lake. The kids love fishing trips, and I can sneak in some grilling. It’s the ritual, you know? When you’ve got four people to coordinate, you need an Excel sheet just for the sandwich toppings.

Other: Couldn’t agree more. Well, thanks for making time today—I know you’ve got plenty of boxes calling your name. If anything changes on the workflow, I’ll loop you in. Otherwise, hope the rest of your shift is smooth, and the boxes lighter than usual.

Target: Appreciate it! Here’s hoping for a forklift without a flat battery and a shift full of coffee breaks. You have a good one, Sam.

Other: Take care, and say hi to the family for me!

[END OF TRANSCRIPT]
\end{tcolorbox}

\end{document}